\documentclass{article}
\usepackage{ijcai19}

\usepackage{times}
\usepackage{soul}
\usepackage{url}
\usepackage[hidelinks]{hyperref}
\usepackage[utf8]{inputenc}
\usepackage[small]{caption}
\usepackage{graphicx}
\usepackage{amsmath}
\usepackage{booktabs}
\usepackage{float}
\usepackage{subcaption}
\usepackage{enumerate}
\usepackage{xcolor, soul}
\usepackage[british]{babel}
\usepackage{comment}
\sethlcolor{yellow}
\usepackage{amsmath,amssymb,amsfonts}
\usepackage{algorithmic}
\usepackage{graphicx}
\usepackage{textcomp}
\usepackage{caption}
\usepackage{multirow}
\usepackage{hyperref}
\usepackage{makecell}
\usepackage{tabularx}
\usepackage{breqn}
\usepackage{longtable}
\usepackage{multicol}
\title{Automatic Scene Generation: State-of-the-Art Techniques, Models, Datasets, Challenges, and Future Prospects}

\author{
Awal Ahmed Fime$^1$\and
Saifuddin Mahmud$^2$ \footnote{Contact Author: Saifuddin Mahmud \\ Email: smahmud@bradley.edu}\and
Arpita Das$^3$\and
Md. Sunzidul Islam$^3$ \And
Hong-Hoon Kim$^1$  \\
\affiliations
$^1$Advanced Telerobotics Research Lab, Computer Science, Kent State University, Kent, Ohio, 44240, USA\\
$^2$Computer Science and Information Systems, Bradley University,
Peoria, Illinois, 61625, USA.\\
$^3$Department of Computer Science and Engineering, Khulna University of Engineering \& Technology, Khulna - 9203, Bangladesh.\\
\emails
afime@kent.edu,
smahmud@bradley.edu,
arpitadas93332@gmail.com,
sunzidulislam12@gmail.com,
jkim72@kent.edu 
}

\begin{document}

\maketitle

\begin{abstract}
Automatic scene generation is an essential area of research with applications in robotics, recreation, visual representation, training and simulation, education, and more. This survey provides a comprehensive review of the current state-of-the-arts in automatic scene generation, focusing on techniques that leverage machine learning, deep learning, embedded systems, and natural language processing (NLP). We categorize the models into four main types: Variational Autoencoders (VAEs), Generative Adversarial Networks (GANs), Transformers, and Diffusion Models. Each category is explored in detail, discussing various sub-models and their contributions to the field.\\

We also review the most commonly used datasets, such as COCO-Stuff, Visual Genome, and MS-COCO, which are critical for training and evaluating these models. Methodologies for scene generation are examined, including image-to-3D conversion, text-to-3D generation, UI/layout design, graph-based methods, and interactive scene generation. Evaluation metrics such as Fréchet Inception Distance (FID), Kullback-Leibler (KL) Divergence, Inception Score (IS), Intersection over Union (IoU), and Mean Average Precision (mAP) are discussed in the context of their use in assessing model performance.\\

The survey identifies key challenges and limitations in the field, such as maintaining realism, handling complex scenes with multiple objects, and ensuring consistency in object relationships and spatial arrangements. By summarizing recent advances and pinpointing areas for improvement, this survey aims to provide a valuable resource for researchers and practitioners working on automatic scene generation.
\end{abstract}



\maketitle

\section{Introduction}
\label{sec:introduction}
Automatic scene generation is the process of creating different world representations using different techniques from machine learning, deep learning, embedded systems, NLP, etc. Manual scene generation is time-consuming. Also, user has to learn different techniques and applications of graphics to generate any scene. This can limit the use of scene generation in different applications. Automatic scene generation can be useful for the user in the non-graphics domain \cite{seversky2006real}. Scene generation has many applications in different fields like robotics, recreation, visual representation, training and simulation, education, research, and many more.

Many research have taken place to generate 3D scenes or 2D images automatically in previous years. Some research works used text descriptors as input from the users and generated 3D scenes or 2D images based on the user specifications. Some took 2D images to generate 3D scenes using different deep-learning models \cite{abdul2018shrec}. There are some researchers who developed 2D images using bounding boxes. Many researchers used graphs to generate a scene. The graph nodes are mainly the objects of the scene and the edges of the graph represent the relations between the objects. Some researchers also used images of objects to generate a 2D image of the scene by merging those objects. Many researchers also added more flexibility to the system by creating an interactive method where user can modify the scene as per needed \cite{ashual2019specifying}. Where most of the works are considered as supervised classification problem by assigning class to each object of different categories, some researches considered it as semi-supervised classification problem having objects of unknown class \cite{jiang2023scenimefy}\cite{li2020semi}. Many researchers also used videos and audio signals as input and generated scenes by applying different techniques to extract features from those videos or audio signals. Some researchers also worked on scene layout by using different deep-learning models. Some researchers tried to develop a hybrid model that can take more than one type of input to generate the final scene. Sora, Gemini and ChatGPT are well known for their efficient scen

Most of the scene generation systems are developed using different deep learning-based algorithms. The most applied algorithms are GAN, VAE, Transformer, and Diffusion models. Different researchers modified the architecture of these base models to generate their systems. Based on the type of input, different types of datasets are used for both the training and evaluation of the systems. Most of the papers frequently leverage the COCO-Stuff \cite{caesar2018coco} and Visual Genome datasets \cite{krishna2017visual}. Additionally, other datasets play crucial roles in various paper proposals. Also, many loss functions and optimizers are implemented in different systems. Every system's performance is evaluated using many types of evaluation metrics. The most used metrics are Fréchet inception distance (FID), Kullback-Leibler (KL) Divergence, Inception score (IS), Intersection over Union (IoU), and Mean Average Precision (mAP). However, many authors also applied many other evaluation metrics for analyzing their system's performance quantitatively. Almost every paper also did a qualitative comparison with other existing state-of-art systems. Some authors also evaluated their method by the users' inspection to identify whether the output of the systems can fulfill the user's satisfaction or not. 

In the paper, we tried to summarize papers on scene generation of previous years as many as possible. 
The paper is divided into different categories based on the type of input of the papers as well as their implementation methods. The categories are i) Image to 3D, ii) Text to 3D, iii) UI/Layout Design, iv) Box to Image, v) Graph, vi) Mask to Image, vii) Interactive Scene, viii) Semi Supervised, ix) Text to Image, x) Video, xi) Image reconstruction, xii) Others. We also worked on finding out the datasets and models they used, the loss functions, optimizers, and metrics to evaluate their system's performances and their contributions. We tried to find out the limitations and challenges they faced during the development of the systems. 
To summarize, our contribution is:\par
1. We summarized the most recent papers on scene generation based on 13 categories and the method of developing the systems for scene generation.\par
2. We described the datasets and models the authors of the papers used to train and evaluate the system. Also, the loss functions, optimizers, and metrics mentioned in the papers are analyzed.\par
3. We tried to find out the challenges and limitations of the development of proposed systems and solutions to resolve the problem mentioned in the papers.

The following sections provide more details about the whole paper. Section \ref{sec:Base models} describes the models that are used for developing scene generation systems in different papers. Section \ref{sec:dataset} explains about the datasets used in different papers we summarized. Section \ref{sec:box to image} explains the papers that worked to generate images using bounding boxes. Section \ref{sec:graph} describes about the papers that used graphs as input. Section \ref{sec:hybrid} summarized the papers that developed a hybrid system for scene or image generation. Section \ref{sec:mask to image} summarized the papers that used masks to generate images. Section \ref{sec:interactive scene} describes papers developing interactive scenes. Section \ref{sec:semi supervised} explains about the papers that used semi-supervised methods for scene generation and section \ref{sec:text to image} summarizes the papers that took text descriptors as input to generate an image. Section \ref{sec:video} summarized the papers that used videos as input. Section \ref{sec:image reconstruction} explains papers related to image reconstruction. Section \ref{sec:ui} describes the papers that designed user interfaces or layouts for scene generation. Section \ref{sec:image to 3D} and Section \ref{sec:text to 3D} explain the papers for 3D scene generation that take image and text as input respectively. Section \ref{sec:others} summarizes other papers related to scene generation. Section \ref{sec:loss functions} and Section \ref{sec:evaluation techniques} describe the loss functions and evaluation techniques used in different papers we summarized respectively. Section \ref{sec:use cases} explains about the use cases of scene generation. Section \ref{sec:Challenges of scene generation} explains the challenges and limitations mentioned in the papers on scene generation. Finally, section \ref{sec:conclusions} concludes the importance of the scene generation survey and the overall contribution of our paper.

\section{Base models}
\label{sec:Base models}
\begin{figure*}[!ht]
\centerline{
\includegraphics[width= \textwidth]{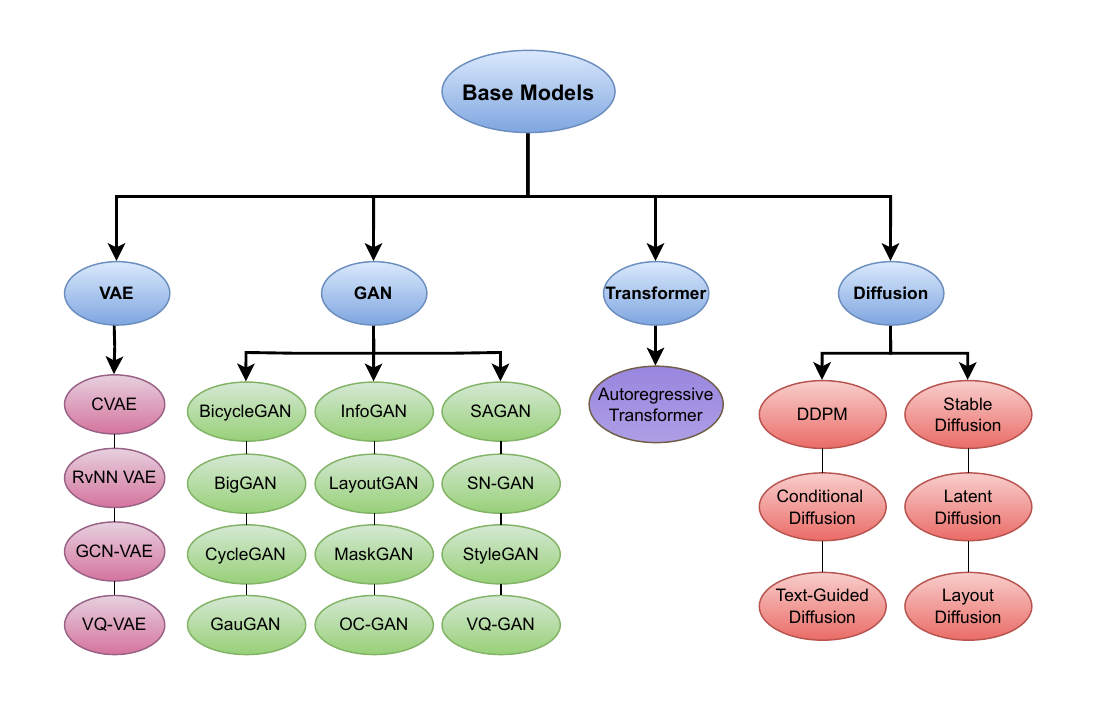}}
\caption{Tree Diagram of Base Models and Submodels.}
\label{Tree Diagram of Base Models and Submodels.}
\end{figure*}

Most of the scene generation models follow some basic architecture. On top of those base models, most of the other researchers customize or optimize their structure. Four base models for scene generation and their sub-models are mentioned in this section. Figure \ref{Tree Diagram of Base Models and Submodels.} shows the tree diagram of the most used models in scene generation and their sub-models.

\subsection{Variational Autoencoder (VAE)}

\begin{figure*}[htbp]
\centerline{
\includegraphics[width= .7\linewidth]{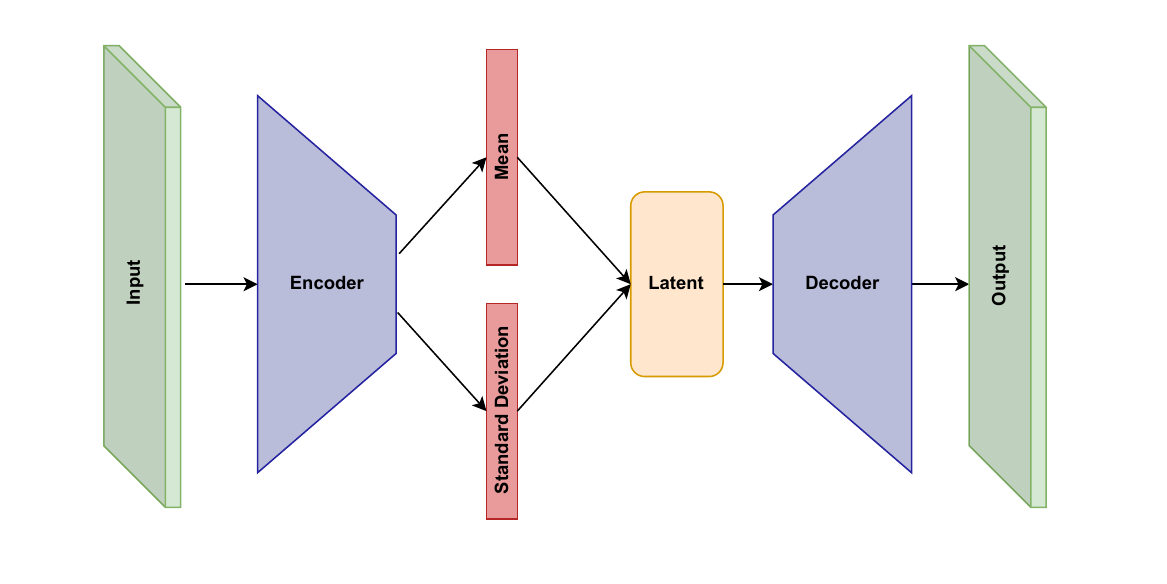}}
\caption{Basic VAE architecture.}
\label{fig:vae_archi}
\end{figure*}
The idea of autoencoders is a setting of an encoder and a decoder that stores information in a small dimension and reconstructs it when necessary.  However, the problem with autoencoders is that they can't generate new content. To solve this issue Variational Autoencoder (VAE) comes into action \cite{kingma2013auto}. It takes the same input as the autoencoder and transforms it into probability distribution within latent space instead of a single point. The decoder takes a sample from the distribution and tries to reconstruct it back to the original data. During training, it tries to adjust the weight of both the encoder and decoder to reduce the reconstruction loss. It tries to balance two things at the same time the reconstruction loss and the regularization term. The combined loss is mentioned in the equation \ref{equ:vae_loss}. This regularization term pushes the latent space to maintain the chosen distribution, preventing overfitting and promoting generalization. The basic architecture of VAE is mentioned in figure \ref{fig:vae_archi}.

\begin{equation}
\mathcal{L}_{VAE} = -\mathbb{E}{q(z|x)}[\log p(x|z)] + \text{KL}[q(z|x) || p(z)] 
\label{equ:vae_loss}
\end{equation}

Where, x represents an image, and z represents latent variables. $\text{KL}[q(z|x) || p(z)]$ is the Kullback-Leibler divergence term and $-\mathbb{E}{q(z|x)}[\log p(x|z)]$ is the reconstruction loss. 

\subsubsection{Conditional variational autoencoder (CVAE)}
One of the problems with VAE is the user doesn't have any control over the reconstruction process. The conditional VAE gives a way to guide image generation through class labels or prompt text \cite{doersch2016tutorial}. Conditional VAE takes additional input along with the input image from the user.

\subsubsection{Recursive Neural Network (RvNN VAE)}
Recursive Neural Network or RvNN is able to extract detailed and structured information from data \cite{li2019grains}. It uses the same weight again and again on the structured data. This structure is useful for tree-like structures. \cite{gao2023scenehgn} is using RvNN with VAE architecture.

\subsubsection{GCN-VAE}
A graph is a common structure to represent knowledge with interconnection. To grab graph information including all its relations to pass through the VAE architecture Graph Convolution Neural Network is used \cite{kipf2016variational}. \cite{dhamo2021graph}\cite{chattopadhyay2023learning} are a few examples of using GCN-VAE for scene generation.

\subsubsection{VQ-VAE}
VQ-VAE has two primary differences from VAEs: firstly, the encoder network produces discrete codes instead of continuous ones; secondly, the prior is learned instead of static \cite{van2017neural}. A discrete latent representation is learned by incorporating concepts from vector quantization (VQ).  By utilizing the VQ approach the model can avoid posterior collapse problems, which occur when latents are neglected and paired with a strong autoregressive decoder. By associating these representations with an autoregressive prior, the model can produce high-quality photos, videos.


\subsection{GAN}
\begin{figure*}[htbp]
\centerline{
\includegraphics[width= .9\textwidth]{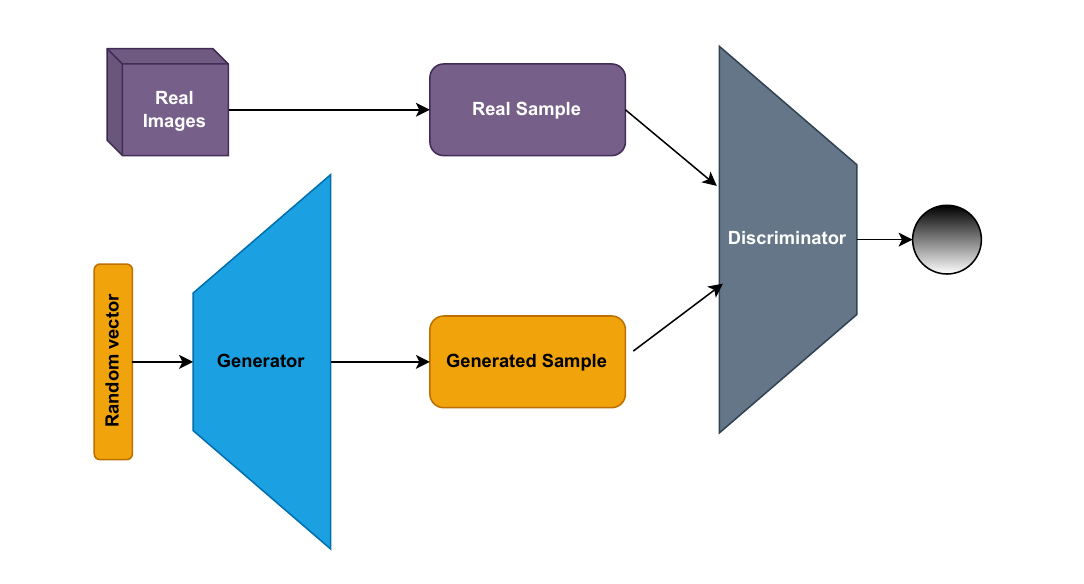}}
\caption{Basic GAN architecture.}
\label{fig:gan_archi}
\end{figure*}

The generative adversarial network is a two-player game cutting-edge approach for image generation \cite{creswell2018generative}. It consists of one generator and a discriminator. The generator tries to generate images as real as possible. For this, it takes a fixed-length vector with random variables from the Gaussian distribution. This vector is close to the latent variable for VAE. With the help of this vector, the generator generates the image of the same domain. The discriminator works as a pure binary classifier in the model. It takes each image and tries to predict whether it is real or fake. GAN is an unsupervised learning model. During training time, if the discriminator identifies the fake image then the discriminator stays the same but the generator gets updated. When the generator fools the discriminator, the discriminator gets updated. This process continues until the generator generates all images that the discriminator can't identify. After training, the discriminator is discarded and only the generator is used to generate new images. The mathematical statement to optimize for GAN is mentioned in equation \ref{equ:gan_opt}. The architecture of the GAN model is shown in figure \ref{fig:gan_archi}.

\begin{dmath}
\min_G \max_D V(D, G) = \mathbb{E}_{x\sim p{\text{data}}(x)}[\log D(x)] + \mathbb{E}_{z\sim p_z(z)}[\log(1 - D(G(z)))] 
\label{equ:gan_opt}
\end{dmath}

Here, \textit{D} represents the discriminator, \textit{G} represents the generator, $p_{\text{data}}(x)$ is the distribution of real data, and $ p_z(z)$ is the distribution of the noise vector ( z ).

\subsubsection{BicycleGAN}
To restrict few number of real samples conversing again and again in the output, BicycleGan was proposed \cite{zhu2017toward}. It adds a bijection between output and latent space. One way the output is generated from the latent space is that the output is again encoded to the latent space. This way it prevents the model generate only few number of samples from the input sample.

\subsubsection{BigGAN}
The main difference between other GANs and BigGANs is the size of the model \cite{brock2018large}. It increases the batch size and width of each layer for the model which improves its performance. It also adds a skip connection from the latent variable to the generator. This model also introduced a new variation of orthogonal regularization. 

\subsubsection{CycleGAN}
As the name mentioned, CycleGAN works in a cyclic order \cite{zhu2017unpaired}. First, from input, it tries to generate the image. After that, it works in the opposite direction, where it tries to generate an input image from the generated image. Then the mean squared error is calculated from the generated image and input image. This process is able to map between paired and unpaired images.

\subsubsection{GauGAN}
GauGAN is developed by NVIDIA. It is capable of taking text, semantic segmentation, sketch, and style within a single GAN and generating an image based on that \cite{park2019semantic}. In the example video, they sketched an area and selected a rock, and based on that it generated a rock at that place. The second version of this model is also published.

\subsubsection{InfoGAN}
InfoGan encourages the GAN network to learn interpretable and meaningful information by maximizing the mutual information between a small set of observation and noise variables \cite{chen2016infogan}. It is more interested in mutual information that helps the model to optimize it properly. As a result, it can separate writing style from the digit shape or visual concepts from the face like hairstyle, emotion, or the presence of eyeglasses.

\subsubsection{LayoutGAN}
LayoutGAN takes multiple layers of the 2D plane and tries to generate semantics and 2D relations among the objects \cite{li2019layoutgan}. It applies multiple self-attention layers on those planes to generate a meaningful layout from it. For this purpose, the rendering layer is the beginning challenge.

\subsubsection{MaskGAN}
MaskGAN maps semantic segmentation to target image \cite{fedus2018maskgan}. It also allows the property to manipulate segmentation maps interactively to generate new maps. These manipulations are learned as traversing on the mask manifold, that gives better result. This model is based on two components Dense Mapping Network and Editing Behavior Simulator. The author of this model also contributed by generating high-resolution face images.

\subsubsection{OC-GAN}
Complex scenes with multiple objects are hard to generate. By addressing this issue, OC-GAN \cite{perera2019ocgan} mainly tried to solve two problems i) some spurious objects are generated without corresponding bounding boxes and ii) Overlapping boxes merge two objects. The author proposed an Object-Centric Generative Adversarial Network to learn about individual objects and the spatial relationships among objects in the scene.

\subsubsection{SAGAN}
For picture generation challenges, attention-driven, long-range dependency modeling is made possible by the Self-Attention Generative Adversarial Network, or SAGAN \cite{zhang2019self}. Using only spatially localized points from lower-resolution feature maps, conventional convolutional GANs produce high-resolution details. With SAGAN, cues from every feature position can be used to produce details. Additionally, the discriminator may verify the consistency of highly detailed features located in distant regions of the image with one another.

\subsubsection{SN-GAN}
Spectral normalization in Generative Adversarial Networks (GANs) resolves problems like vanishing gradient, exploding gradient, and model collapsing using Spectral normalization. It also makes the training process easier and smoother. Spectral normalization, dividing by each weight spectral norm, acts as a form of normalization, that keeps the model mathematically predictive.

\subsubsection{StyleGAN}
One kind of generative adversarial network is StyleGAN \cite{miyato2018spectral}. Drawing on style transfer literature, it employs an alternate generator architecture for generative adversarial networks. Otherwise, it uses a training regimen that increases gradually, similar to Progressive GAN. Among its other peculiarities is that, unlike conventional GANs, it derives from a fixed value tensor rather than stochastically generated latent variables. After being converted by an 8-layer feedforward network, the stochastically generated latent variables are employed as style vectors in the adaptive instance normalization at each resolution. Finally, it uses mixing regularization, a type of regularization in which two style latent variables are mixed during training.

\subsubsection{Vector Quantised Generative Adversarial Network (VQ-GAN)}
VQGAN (Vector Quantized Generative Adversarial Network) combines the power of Vector Quantized with GAN to generate higher quality image \cite{verma2023vector}. The vector quantizer takes the output from the image generator and divide some small fixed number of discrete codebook vectors. Then again reconstruct the image from this vector. This reduces noise in the image and also able to add constraints on certain objects in a generation.

\subsection{Transformer}

\begin{figure}[htbp]
\centerline{
\includegraphics[width= \linewidth]{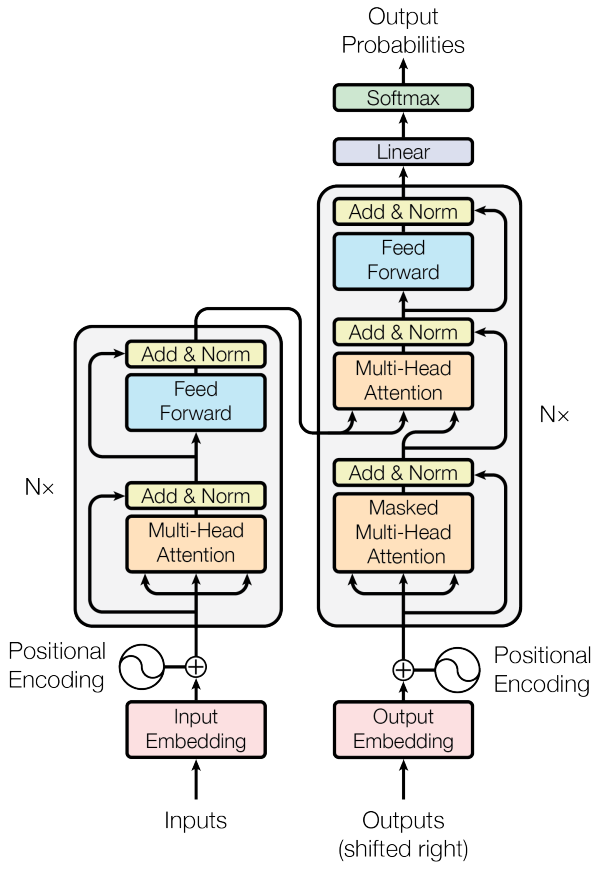}}
\caption{Basic Transformer architecture. \protect \cite{vaswani2017attention}}
\label{fig:trans_archi}
\end{figure}
Taking inspiration from RNN and CNN, Transformers \cite{vaswani2017attention} understand the context of data and generate new data. This consists of an encoder and decoder network. It takes sequential data as input for the beginning and converts it to a numerical representation called embedding. Embedding contains the semantic meaning of the content. Along with this embedding, some additional information as a vector feeds into the encoder model. This vector contains positional information in a specific pattern. Multiple self-attention is used as multi-head attention to capture different interrelations among the tokens. Softmax activation is used to calculate the weight of the attention. To stabilize and speed up training layer normalization and residual connections are used with multi-Head attention. To capture more complex patterns in the data a feedforward neural network is used. Multiple blocks of attention are stuck one after another to capture hierarchical and abstract features in data. The decoder network is almost the same as the encoder layer for the Transformer. Instead of input embedding it uses the output embedding. Additionally, it has a linear classifier and softmax layer to generate probabilistic output. A basic architecture form \cite{vaswani2017attention} is shown in figure \ref{fig:trans_archi}.

\subsubsection{Autoregressive Transformers}
Autoregressive models generate the next based on the previous measurement from the sequence \. It is closely related to knowledge-based statistical time-series analysis that predicts the next thing based on probability. The autoregressive technique can be applied with Transformers for sharpening, up-scaling, and reconstructing images while maintaining the quality. \cite{paschalidou2021atiss} is an example for  Autoregressive Transformers to generate indoor scene.


\subsection{Diffusion}

\begin{figure*}[htbp]
\centerline{
\includegraphics[width= \textwidth]{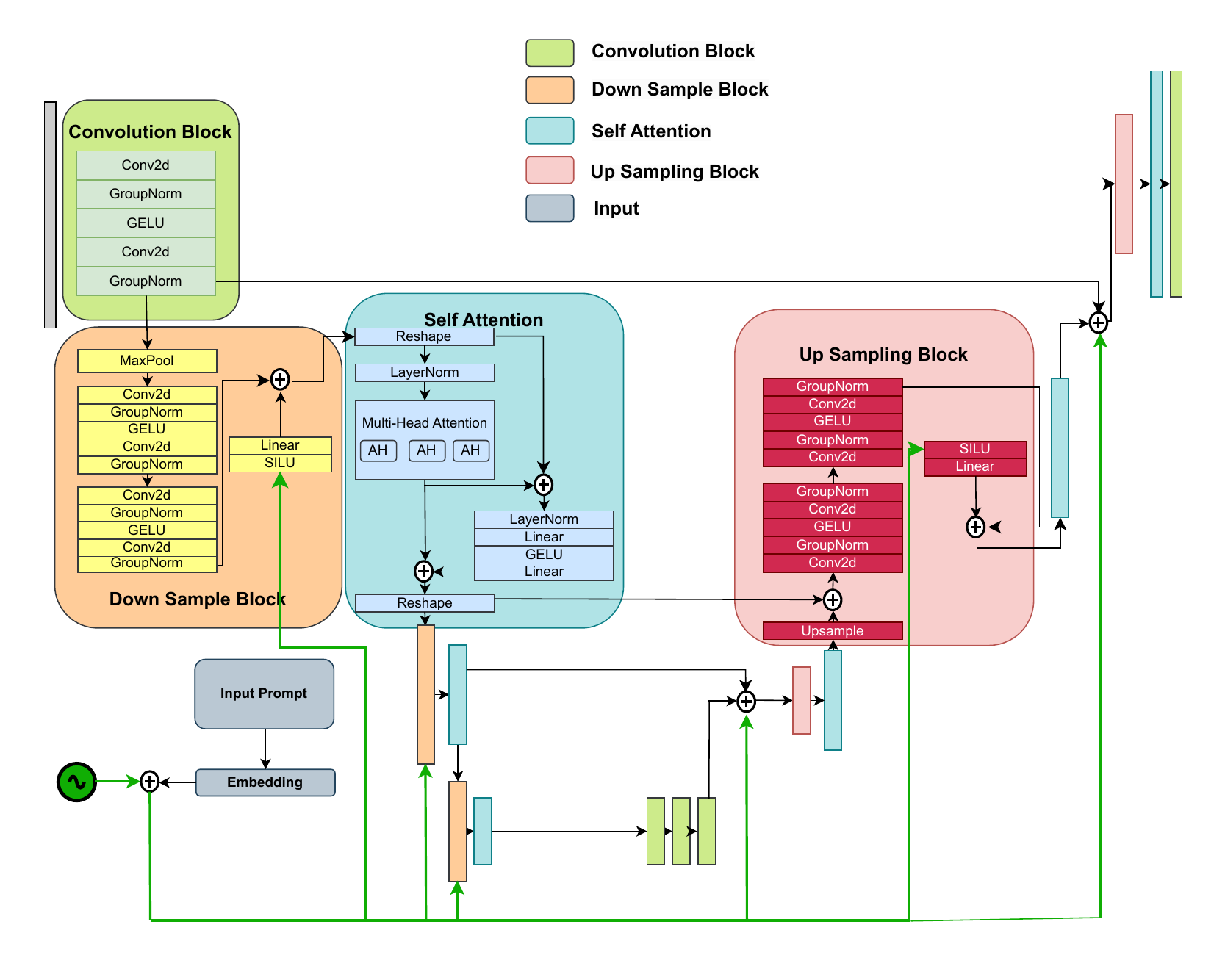}}
\caption{Basic Diffusion architecture.}
\label{fig:diff_archi}
\end{figure*}

Diffusion \cite{ho2020denoising} is a layer-by-layer approach trained in both forward and backward processes to generate new images. For forward diffusion models, it takes the input image and adds controlled Gaussian noise at each step. Using the reparameterization trick, all steps can be done at the same time. At the end of the network, it becomes only noise with very little information about the original image. In the backward process, at each step, the diffusion model tries to predict the noise of that step and subtract that noise from the input of that step. That makes it less noisy than it was as input of that block. The basic architecture of the diffusion model is shown in figure \ref{fig:diff_archi}.

\begin{equation}
 x_t \sim q(x_t|x_0) = \mathcal{N}(x_t,\sqrt{\Bar{\alpha}_t}x_0, (1-\Bar{\alpha}_t)I
\label{equ:diff}
\end{equation}

Noise are added through equation \ref{equ:diff} at each step. Here, \textit{I} is the identity matrix, $\mathcal{N}$ represents the normal distribution, $x_0$ is the initial image, and $x_t$ image includes noise after t-steps. $\Bar{\alpha}_t$ represents summation of all $1-\beta$ where $\beta$ is the standard deviation of noise for that step.

\subsubsection{Denoising Diffusion Probabilistic Model (DDPM)}
The diffusion model consists of a forward noising and backward denoising process. The author of this model was the first to implement Diffusion for high-quality image generation. They demonstrated that the greatest quality results are produced when a certain parameterization of diffusion models demonstrates an equivalency with annealed Langevin dynamics during sampling and with denoising score matching over several noise levels during training. \cite{bautista2022gaudi} is one example of utilizing this diffusion for image generation.

\subsubsection{Conditional Diffusion}
For conditional diffusion with the basic denoising training \cite{fu2024unveil}, it is possible to add an additional input variable as a condition. This variable will guide the model to generate specific things. This guide can be scaler input, class label or text sequence. CLIP embeddings \cite{khandelwal2022simple} is one of the most popular guidings for the diffusion model.

\subsubsection{Text-guided Diffusion}
Text-guided diffusion is another conditional diffusion that takes text as the condition \cite{kim2022diffusionclip}. Then the model converts the text to embedding and uses it as a guide. GLIDE \cite{halgren2004glide}, DALL-E \cite{ge2022dall} are some examples of text-guided diffusion models which are vastly popular.

\subsubsection{Stable Diffusion}
Stable Diffusion is capable of taking an image or text as an input prompt and converts to an image. It is a user-friendly model, that can be used with simple consumer-grade GPU. It is also capable of generating video or animation. Stable diffusion uses UNet architecture to predict noise.

\subsubsection{Latent Diffusion}
In general, the diffusion model works in pixel space, as a result it takes a huge space and time to convert image to image. To solve this, Latent Diffusion is introduced \cite{rombach2022high}. It has two key points i) Perceptual image compression: it converges images in lower dimensions without losing much information. ii) Latent Diffusion: This small dimensional data is used for further training. That makes the table stable and faster. 

\subsubsection{Layout Diffusion}
LayoutDiffusion depicts layout generation as a discrete denoising diffusion process since the layout is usually represented as a series of discrete tokens \cite{zhang2023layoutdiffusion}. It gains the ability to reverse a mild forward process, where in neighboring step layouts do not significantly differ from one another and layouts grow more chaotic as forward steps increase. Finally, a piece-wise linear noise schedule combined with a block-wise transition matrix.

\section{Dataset}
\label{sec:dataset}
Most of the papers frequently leverage the COCO-Stuff \cite{caesar2018coco} and Visual Genome datasets \cite{krishna2017visual}, as they provide rich annotations that facilitate a deeper understanding of relationships between objects within scene graphs or bounding boxes. Additionally, other datasets play crucial roles in various paper proposals, aiding in tasks related to scene understanding. Table \ref{tab:dataset} provides an overview of datasets utilized across multiple studies, each serving distinct research objectives.

\subsubsection{COCO-Stuff} 
The majority of papers utilize the COCO-Stuff dataset \cite{caesar2018coco}, which comprises approximately 118,000 annotated images for training and 5,000 for validation. This dataset serves as the foundation for understanding the relationship between objects used in Scene Graph Generation (Sg2Im) \cite{johnson2018image}. Within the COCO-Stuff dataset, there exist 91 \textit{stuff} categories and 80 \textit{thing} categories. Researchers leverage these annotations to construct scene graphs based on the 2D image coordinates of the objects, facilitating the development and evaluation of various computer vision models and algorithms.

\subsubsection{Visual Genome} Due to the presence of noisy annotations in the Visual Genome dataset \cite{krishna2017visual}, several researchers have opted to utilize the VG-MSDN dataset \cite{li2017scene}. This alternative dataset offers over 46,000 training images and a testing set comprising 10,000 images. Within the Visual Genome dataset, there are 108,077 images annotated with scene graphs, encompassing 92 \textit{thing} categories and 87 \textit{stuff} categories. Researchers leverage these datasets to address the challenges posed by noisy annotations and to advance the field of computer vision, particularly in tasks such as object detection, scene understanding, and image generation.

\subsubsection{HICO-DET} HICO-DET \cite{chao2018learning} is a framework designed for human-object interaction modeling, where all datasets are centered around humans. Within this framework, images are processed with a certain threshold to determine which ones to keep and which ones to discard. Specifically, out of the available images, 15,963 are utilized for training, while 4,034 are allocated for testing purposes.

\subsubsection{SUNCG} SUNCG\cite{song2017semantic} is a comprehensive dataset featuring synthetic 3D scenes, densely annotated with volumetric data. The training dataset comprises 53,860 bedroom scenes, with an average of 13.15 objects present in each scene.

\subsubsection{RPLAN} RPLAN \cite{wu2019data} is a manually collected large-scale densely annotated dataset of floor plans from real residential buildings which consists of more than 80k floorplan images of real residential buildings.

\subsubsection{Synscapes} Synscapes \cite{wrenninge2018synscapes} is a synthetic dataset for street scene parsing created using photorealistic rendering techniques and shows state-of-the-art results for training and validation as well as new types of analysis.

\subsubsection{Pfb} PfB \cite{richter2017playing} dataset in a graph format, it becomes easier to analyze and understand the complex interactions and dynamics of urban traffic within the synthetic environment.

\subsubsection{YCB} From the YCB dataset \cite{calli2015ycb}, one can select objects with diverse characteristics such as size, texture, and shape to create a varied and representative subset.

\subsubsection{ColorMNIST} Colored MNIST is a synthetic dataset derived from the original MNIST dataset, which consists of grayscale images of handwritten digits from 0 to 9. In ColoredMNIST, each grayscale image is transformed into a colored image by overlaying a color channel onto the original grayscale image. This

\onecolumn
\begin{center}
\begin{longtable}{| p{.10\textwidth} | p{.15\textwidth} | p{.65\textwidth} |} 
\caption{Dataset description}\\
\hline
\label{tab:dataset}
 \textbf{Dataset} &   \textbf{Number of Images} &   \textbf{Description} \\
\hline\hline
\endfirsthead
\hline
 \textbf{Dataset} &   \textbf{Number of Images} &   \textbf{Description} \\
\hline\hline
\endhead
COCO-Stuff \cite{caesar2018coco}& \makecell{ training: 118,000 \\ validation: 5,000} &   Scene understanding tasks like semantic segmentation, object detection and image captioning, and also shows the relationships between objects in Scene Graph. \\
        \hline
        Visual Genome \cite{krishna2017visual}& \makecell{ training: 70-80\% \\ validation: 10-15\%} &   Contains region descriptions and question-answer pairs associated with WordNet synsets. \\
         \hline
        HICO-DET \cite{chao2018learning}& \makecell{training: 15,963 \\ validation: 4,034}& Detecting Human Object Interactions (HOI) in image.  \\
         \hline
        SUNCG \cite{song2017semantic}& \makecell{training: 53,860 \\ validation: 10-20\%} & A large-scale dataset of synthetic 3D scenes with dense volumetric annotations. \\
         \hline
         RPLAN \cite{wu2019data}& \makecell{training: 118,000 \\ validation: }&  A manually collected large-scale, densely annotated dataset of floor plans from real residential buildings. \\
         \hline
        Synscapes \cite{wrenninge2018synscapes}&\makecell{training:000 \\ validation: 000}& A synthetic dataset for street scene parsing created using photorealistic rendering techniques. \\
        \hline
        Pfb \cite{richter2017playing}& \makecell{training:  \\ validation: }&  Complex interactions and dynamics of urban Traffic \\
        \hline
        YCB \cite{calli2015ycb}&\makecell{ training: 80,000  \\ validation: 5,000}& Designed for facilitating benchmarking in robotic manipulation, there are also mesh models and high-resolution RGB-D scans of the objects for easy incorporation into manipulation and planning software platforms.  \\
        \hline
        ColorMNIST &\makecell{training: 8,000 \\ validation: 8,000}&  A synthetic binary classification task derived from MNIST. \\
        \hline
        CLEVR-G & \makecell{training: 10,000; 64x64 \\ validation: 10,000}&  A synthetic Visual Question Answering dataset that also contains images of 3D-rendered objects. \\
        \hline
        CelebAMask \cite{lee2020maskgan}& \makecell{training: 30,000 \\ validation:} & High-resolution Face images \\
        \hline
        LAION-5B & \makecell{training: 5.85 million \\ Validation: }& An open, large-scale dataset for training next-generation image-text models. \\
        \hline
        CC12m & \makecell{training: 12 million \\ validation: }&  A dataset with 12 million image-text pairs specifically meant to be used for vision and language pre-training. \\
        \hline
        CC & \makecell{training: 3.3 million \\ validation: 16,000}&  A dataset containing image-URL and caption pairs is designed for the training and evaluation of machine-learning image captioning systems. \\
        \hline
        MS-COCO &\makecell{ training: 83,000 \\ validation: 41,000}&  A large-scale dataset for Object Detection, Segmentation, Captioning, and Key-Pointing detection. It consists of 328k images. \\
        \hline
       Cityscapes. and Cityscapes 25k \cite{cordts2016cityscapes}& \makecell{training: 2,975 \\  validation: 5,00}& Focuses on Semantic understanding of Urban Street Environments. \\
        \hline
        Indian Driving Dataset (IDD) \cite{varma2019idd}&\makecell{training: 10,000 \\ validation: }&  A dataset for road scene understanding in unstructured environments used for semantic segmentation and object detection for autonomous driving. \\
        \hline
        ADE20K \cite{zhou2017scene}& \makecell{training: 25,574 \\ validation: 2,000}& A dataset includes, for instance segmentation, semantic segmentation, and object detection tasks. \\
        \hline
        ModelNet \cite{wu20153d}& \makecell{training: 9,843 \\ validation: 2,468}&  The dataset contains Synthetic Object Points Clouds. \\
        \hline
        ShapeNet \cite{chang2015shapenet}& \makecell{training: 100,000  \\ validation: }& Large-scale dataset of 3D shapes. \\
        \hline
        ImageNet & \makecell{training: 80,000 \\ validation: 5,000}&  A large visual database designed for use in visual object recognition. \\
        \hline
        Flickr & \makecell{training: 70,000 \\  validation: 10,000}&   A standard benchmark for sentence-based image description \\
        \hline
        Places & \makecell{training: 98,721 \\ validation: 1,00}&  Designed following principles of human visual cognition and to build a core of visual knowledge for high-level visual understanding tasks, such as scene context, object recognition, action and event prediction, and theory-of-mind inference.  \\
        \hline
        ScanNet & \makecell{training: 1201 \\  validation: 312}&  An instance-level indoor RGB-D dataset that includes both 2D and 3D data. \\
        \hline
        3D-Front \cite{fu20213d}& \makecell{ training:  \\ validation: }& Large-scale, synthetic indoor scenes highlighted by professionally designed layouts and a large number of rooms populated by high-quality textured 3D models. \\
        \hline
        Matterport3D & \makecell{training: 4939 \\ validation: 456}&   A large RGB-D dataset for scene understanding in indoor environments. \\
        \hline
        YouTube 3D \cite{chen2019learning}& \makecell{training: 47,125 \\ validation: 1,525} &  This dataset was developed using the videos from YouTube and contains 3D vertex coordinates.  \\
        \hline
        OASIS \cite{chen2020oasis}& \makecell{training: 9664 \\ validation: 1120}& A Large-Scale Dataset for Single Image 3D in the Wild. \\
        \hline
        KITTI \cite{vasiljevic2019diode}& \makecell{training: 7,480 \\  validation: 7481}& Used in Mobile robotics and autonomous driving and also consists of traffic scenarios. \\
        \hline
        DIODE \cite{schops2017multi}& \makecell{training: 25458 \\ validation: 771}&  Contains high-resolution color images for the Indoor and outdoor scenes. \\
      \hline
      
\end{longtable}

  \end{center}
\twocolumn

\vspace{-5pt}

overlaying process creates a synthetic dataset where each digit image now has a color associated with it. In this case, they are using 8000 training and 8000 test images.

\subsubsection{CLEVR-G} CLEVR is a standard dataset containing rich compositionality and complicated scenes for examining reasoning in visual question answering, which contains 10,000 64x64 training images and 10,000 testing images.

\subsubsection{CelebAMask} CelebAMask-HQ \cite{lee2020maskgan} is a large-scale face image dataset that has 30,000 high-resolution face images selected from the CelebA dataset by following CelebA-HQ. Each image has a segmentation mask of facial attributes corresponding to CelebA. The masks of CelebAMask-HQ were manually annotated with a size of 512 x 512 and 19 classes, including all facial components and accessories such as skin, nose, eyes, eyebrows, ears, mouth, lip, hair, hat, eyeglass, earring, necklace, neck, and cloth.

\subsubsection{LAION-5B} The LAION-5B dataset represents an open large-scale dataset for the training of the next generation of image-text models. It comprises a vast collection of 5.85 billion image-text pairs, meticulously filtered through CLIP technology. Among these pairs, 2.3 billion are in English, while an impressive 2.2 billion samples span over 100 other languages. Furthermore, the dataset offers several nearest neighbor indices, enhancing the web interface for exploration and subset creation. Additionally, it includes detection scores for watermark presence and NSFW content, providing comprehensive utility for various applications.

\subsubsection{CC12m} The CC12M (Conceptual 12M) dataset represents a significant advancement in web-scale image-text pre-training, specifically designed to identify and understand the vast array of visual concepts present in the long tail of online content. With a collection of 12 million image-text pairs, this dataset serves as a foundational resource for pre-training models that integrate both vision and language. Distinguished by its size and scope, CC12M surpasses its predecessor, Conceptual Captions (CC3M), offering a much broader and more diverse range of visual concepts. This diversity makes it an invaluable asset for pre-training and end-to-end training of image captioning models, enabling them to capture and comprehend a wider spectrum of visual information.

\subsubsection{CC} The Conceptual Captions (CC) dataset is tailored for training and assessing machine-learned image captioning systems, comprising pairs of (image-URL, caption). Conceptual Captions offers an expansive collection, exceeding 3 million images paired with natural language captions. Its release includes two splits: a training set with around 3.3 million examples and a validation set with approximately 16,000 examples.

\subsubsection{MS-COCO} The MS-COCO (Microsoft Common Objects in Context) dataset is a comprehensive resource designed for large-scale tasks such as object detection, segmentation, captioning, and key-point detection. Originally released in 2014, the dataset comprised 164,000 images, divided into training (83,000), validation (41,000), and testing (41,000) sets. By 2017, the dataset underwent revisions, notably expanding the training and validation splits to 118,000 and 5,000 images, respectively. This adjustment aimed to enhance model training and evaluation by providing a larger and more diverse set of images for both tasks.

\subsubsection{Cityscapes. and Cityscapes 25k} The Cityscapes \cite{cordts2016cityscapes} dataset provides detailed insights into the semantic composition of urban street environments. With a focus on dense semantic segmentation, it offers polygonal annotations across 30 distinct classes. The dataset comprises 5,000 images meticulously annotated with fine details, along with an additional 20,000 images annotated with coarser labels. Among these, 2,975 images are designated for training, while 500 images serve as validation data. Each image is standardized at 256x512 pixels and presented as a composite, with the original photograph occupying the left half and its corresponding labeled segmentation output displayed on the right half. This dataset serves as a valuable resource for advancing research in urban scene understanding and semantic segmentation tasks. The current state-of-the-art on Cityscapes 25K \cite{cordts2016cityscapes} 256x512 is SB-GAN.

\subsubsection{Indian Driving Dataset (IDD)} Indian Driving Dataset (IDD) \cite{varma2019idd} is a novel dataset for road safety understanding in unstructured environments. It consists of 10,000 images and has already assigned 34 classes collected from 182 drive sequences on Indian roads. The images are mostly 1080p resolution. The dataset consists of images obtained from a front-facing camera attached to a car.

\subsubsection{ADE20K} The ADE20K \cite{zhou2017scene} dataset is a comprehensive collection of over 27,000 images sourced from the SUN and Places databases. These images are meticulously annotated with objects, covering a vast spectrum of over 3,000 object categories. The dataset comprises 27,574 images, with 25,574 allocated for training purposes and 2,000 for testing. These images span across 365 different scenes, providing diverse contextual backgrounds for object recognition tasks. Within the dataset, there are a staggering 707,868 unique objects categorized into 3,688 distinct categories. Moreover, the dataset includes annotations for 193,238 object parts and sub-parts.

\subsubsection{ModelNet} The ModelNet40 \cite{wu20153d} dataset comprises meticulously curated synthetic object point clouds, comprising 12,311 CAD-generated meshes distributed across 40 distinct categories. Notably, 9,843 meshes are allocated for training purposes, with the remaining 2,468 reserved exclusively for testing.

\subsubsection{Shapenet} Shapenet \cite{chang2015shapenet} is a large-scale repository that contains over 300 million models, with 220,000 classified into 3.135 classes arranged using WordNet hypernym-hyponym relationships.

\subsubsection{ImageNet} The ImageNet dataset is a massive collection of annotated images, totaling 14,197,122 entries. It serves as the foundation for the ImageNet Large Scale Visual Recognition Challenge (ILSVRC), a renowned benchmark in tasks like image classification and object detection. Within the dataset, there are 1,034,908 images with bounding box annotations, facilitating precise object localization. The WordNet hierarchy enriches the dataset with semantic information, comprising a total of 21,841 non-empty synsets. Additionally, to enhance feature representation, 1,000 synsets are endowed with Scale-Invariant Feature Transform (SIFT) features. These features are extracted from a subset of 1.2 million images, further refining the dataset's utility and depth for various computer vision tasks.

\subsubsection{Flickr} The Flickr30k dataset has emerged as a foundational benchmark for tasks centered around generating descriptions for images based on sentences. It comprises a vast collection of 158,000 captions extracted from Flickr images. Within this dataset, there are 244,000 coreference chains, aiding in resolving references within the captions and improving overall coherence. Moreover, the dataset includes 276,000 manually annotated bounding boxes, offering precise localization information for various objects depicted in the images. This combination of rich textual descriptions and meticulously annotated visual data makes Flickr30k a crucial resource for advancing research in image understanding and natural language processing.

\subsubsection{Places} The Places dataset is proposed for scene recognition and contains more than 2.5 million images covering more than 205 scene categories with more than 5,000 images per category.

\subsubsection{ScanNet} ScanNet is an instance-level indoor RGB-D dataset that includes both 2D and 3D data. It is a collection of labeled voxels rather than points or objects. ScanNet v2 is the newest version of the ScanNet dataset. It has collected 1513 annotated scans.The scans have an approximate 90\% surface coverage. In the semantic segmentation task, the dataset is marked with 20 classes of annotated 3D voxelized objects.

\subsubsection{3D-Front}  3D-FRONT (3D-furnished rooms with layouts and semaNTics) \cite{fu20213d} is a large-scale, comprehensive repository of synthetic indoor scenes. It features professionally designed layouts and rooms populated with high-quality, textured 3D models with style compatibility. Freely available to the academic community and beyond. It contains 18,797 diversely furnished rooms.7,302 furniture objects with high-quality textures. Floorplans and layout designs sourced from professional creations. Interior designs (furniture styles, colors, textures) curated using a recommender system to attain consistent styles as expert designs.

\subsubsection{Matterport3D} The Matterport 3D dataset is a large RGB-D dataset for scene understanding in indoor environments. It contains 10,800 panoramic views inside 90 real building-scale scenes, constructed from 194,400 RGB-D images. Each scene is a residential building consisting of multiple rooms and floor levels, and is annotated with surface construction, camera poses, and semantic segmentation.

\subsubsection{YouTube 3D}  The Youtube 3D dataset \cite{chen2019learning} contains 3D vertex coordinates of 50,175 hand meshes aligned with the wild images, comprising hundreds of subjects performing a wide variety of tasks. The training set was generated from 102 videos, resulting in 47,125 hand annotations. The validation and test sets cover 7 videos with an empty intersection of subjects with the training set and contain 1,525 samples each.

\subsubsection{OASIS}  The OASIS dataset \cite{chen2020oasis} for single-image 3D in the wild consists of annotations of detailed 3D geometry for 140,000 images.

\subsubsection{KITTI}  KITTI (Karlsruhe Institute of Technology and Toyota Technological Institute) \cite{geiger2012we} is a popular dataset for mobile robotics and autonomous driving. It contains traffic scenarios recorded with RGB cameras, stereo cameras, and 3D laser scanner. There is no ground truth for semantic segmentation in the original dataset. There is ground truth for 323 images with 3 classes: road, vertical, and sky, and there are also annotated 252 acquisitions with 10 object categories and labeled 216 images with 11 classes. Classes include buildings, trees, sky, cars, signs, roads, pedestrians, fences, poles, sidewalks, and bicyclists.

\subsubsection{DIODE}  Diode \cite{vasiljevic2019diode} (Dense Indoor/Outdoor Depth (DIODE)) is the first standard dataset for monocular depth estimation, comprising diverse indoor and outdoor scenes acquired with the same hardware setup. The training set consists of 8574 indoor and 16884 outdoor samples from 20 scans each. The validation set contains 325 indoor and 446 outdoor samples, with each set from 10 different scans. The ground truth density for the indoor training and validation splits are approximately 99.54\% and 99\%, respectively. The density of the outdoor sets is naturally lower, with 67.19\% for training and 78.33\% for validation subsets. The indoor and outdoor ranges for the dataset are 50m and 300m, respectively.

\subsubsection{ETH3D}  ETHD \cite{schops2017multi} is a multi-view stereo benchmark and 3D reconstruction benchmark that covers a variety of indoor and outdoor scenes. Ground-truth geometry has been obtained using a high-precision laser scanner. A DSLR camera as well as a synchronized multi-camera rig with varying field-of-view were used to capture images.

\subsubsection{Waymo Open}  The Waymo Open Dataset \cite{sun2020scalability} is composed of two datasets: the perception dataset with high-resolution sensor data and labels for 2,030 scenes, and the motion dataset with object trajectories and corresponding 3D maps for 103,354 scenes. The Waymo Open Dataset currently contains 1,950 segments.

\subsubsection{VGG-Sound} The VGG-Sound consists of more than 210k videos for 310 audio classes.

\subsubsection{VRD}  The Visual Relationship Dataset (VRD) contains 4000 images for training and 1000 for testing, annotated with visual relationships. Bounding boxes are annotated with a label containing 100 unary predicates.

\subsubsection{RICO} The Rico dataset contains design data from more than 9.3k Android apps spanning 27 categories. It exposes the visual, textual, structural, and interactive design properties of more than 66k unique UI screens.

\section{Box Layout to image}
\label{sec:box to image}
\begin{figure*}[htbp]
\centerline{
\includegraphics[width= \textwidth]{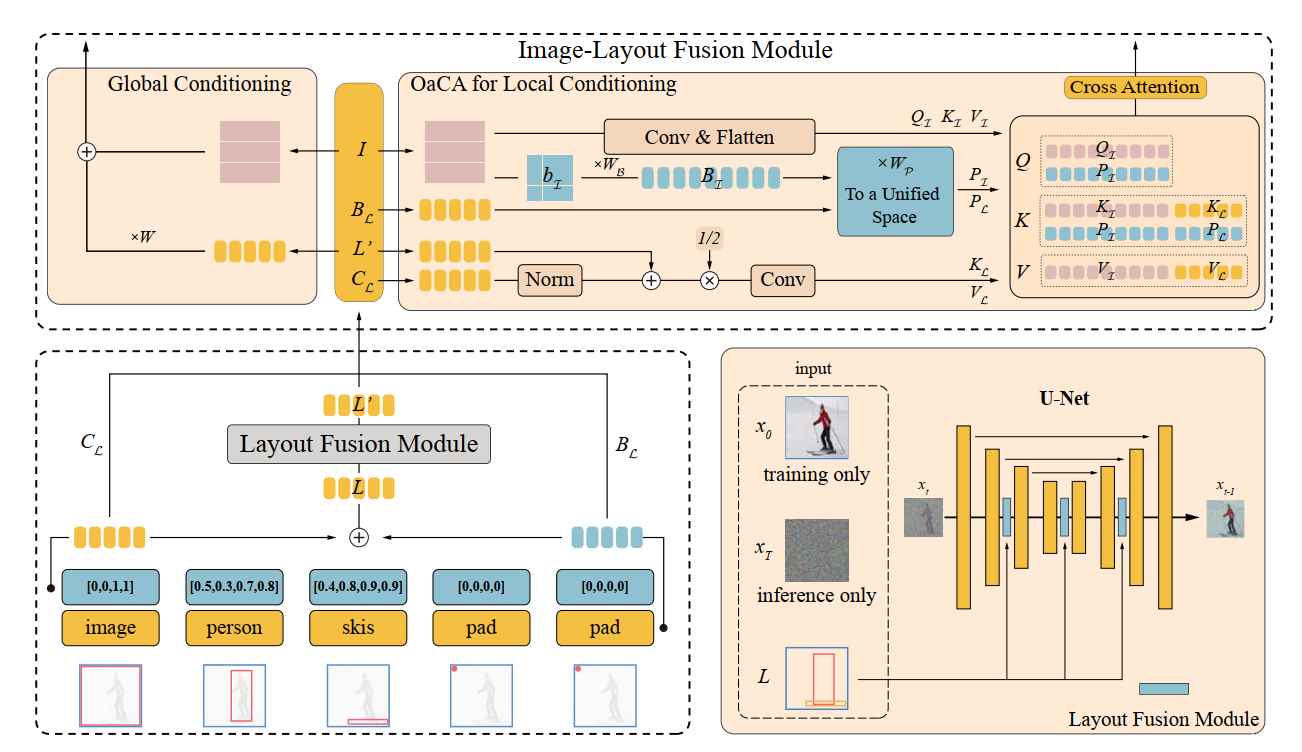}}
\caption{Flow Diagram of A Box to Image Generation \protect \cite{zheng2023layoutdiffusion}.}
\label{Flow diagram}
\end{figure*}

Generating realistic images from layouts can be challenging, especially when scenes are divided into boxes representing multiple objects. To address this, various models have been proposed to enhance image generation comfort. Many approaches involve segmenting each box and encoding its information, leveraging techniques like Transformer with Focal Attention (TwFA) \cite{yang2022modeling} to improve generation quality. Additionally, models often employ location-aware cross-object attention mechanisms to ensure accurate object placement within the scene. To enhance interaction with the layout, some methods utilize layout fusion modules, which help create a structural path for image generation. By composing feature maps extracted from the layout using techniques like LSTM, these models can generate more realistic output images. Overall, these advancements in layout-based image synthesis aim to provide more comfortable and accurate results, bridging the gap between scene representation and realistic image generation. Figure \ref{Flow diagram} presents a visual depiction of the process illustrating how these models are employed to generate realistic output images.

In this paper\cite{yang2023law}, the author used vanilla classifier-free guidance \cite{ho2022classifier} \cite{nichol2021glide} to generate an image from the layout. First, they segment each box and encode all information from the box into a fragment to learn layout embedding. To perform multihead attention on each fragmented box, they proposed location-aware cross-object attention. Then they regrouped those fragments based on their spatial locations.  To measure rationality and harmonious relations in the generated image, they mentioned a new evaluation matrix called Scene Relation Score. For this matrix, they generated graphs and labels using pre-trained VCTReeEB \cite{suhail2021energy} and compared this box with the original input. They used the same setting \cite{dhariwal2021diffusion} for their training procedure. However, this model still lacks a second-level style and a global context. 

To address the challenge of scene layout generation across diverse domains, this paper \cite{gupta2021layouttransformer} proposes a novel framework called LayoutTransformer. This framework maps different parameters of primitive elements to a fixed-length continuous latent vector and then uses a masked Transformer decoder to predict the next primitive in the layout (one parameter at a time) based on representations of existing primitives. The generative framework can start from an empty set or a set of primitives and iteratively generate a new primitive parameter at a time. It predicts whether to stop or generate the next primitive, allowing for variable-length layouts. LayoutTransformer takes layout elements as input and predicts the next layout elements as output. During training, teacher forcing is employed, using ground-truth layout tokens as input to a multi-head decoder block. The first layer of the block is a masked self-attention layer, enabling the model to see only previous elements while predicting the current element. The framework pads each layout with a special <bos> token at the beginning and <eos> token at the end. In comparison, LayoutGAN \cite{li2019layoutgan} uses a GAN framework to generate semantic and geometric properties of a fixed number of scene elements.

The paper \cite{zhao2020layout2image} introduces Layout2Im, a novel approach for layout-based image generation that aims to tackle the challenge of generating images with complex object layouts. The method combines Convolutional LSTM to merge individual object representations into a comprehensive layout encoding, which is then decoded into an image. The model addresses two key challenges: limited generalizability in generating complex, real-world images, and ambiguity in textual descriptions of object categories and appearances. To address these challenges, categories are encoded using word embeddings, while appearances are encoded into low-dimensional vectors sampled from normal distributions. Key steps include object feature map composition, object latent code estimation, and object-wise attention (OWA), which employs attention masks based on word embeddings to accurately represent object shapes and classes. The model fuses object feature maps into a hidden feature map for generating realistic images and employs an image decoder for the final output. To ensure consistency between latent codes and outputs, object latent code regression is used, while discriminators enhance realism and object recognizability by classifying input images as real or fake. Future directions include improving controllability in image generation through fine-grained attribute specification. The approach is validated with qualitative and quantitative results on COCO-Stuff \cite{caesar2018coco} and Visual Genome \cite{krishna2017visual} datasets, demonstrating the model's ability to generate complex images while maintaining respect for object categories and their layout.

This paper \cite{li2021image} presents a method for synthesizing images based on layouts, which consist of bounding boxes with object categories. The focus is on generating clean and semantically clear semantic masks to improve image generation from layouts, representing object positions, sizes, and classes. To address this, the paper proposes the Locality-aware Mask Adaptation (LAMA) module, designed to adapt overlapped or nearby object masks during generation. LAMA considers local relations between objects and scales mask values for each object individually with a learned matching mechanism. Empirical results demonstrate that LAMA improves mask clarity, particularly in scenarios where object masks overlap. The proposed model consists of two stages: layout-to-mask and mask-to-image. In the layout-to-mask stage, semantic masks are generated based on the layout of objects. In the mask-to-image stage, images are generated using the generated semantic masks. This step incorporates modifications to the mask-to-image component of LostGAN-V1 \cite{sun2019image}, including replacing Batch Normalization with BatchGroup group normalization (BGN) in ISLA-norm layers \cite{sun2019image}, applying noise injection \cite{karras2019style} after convolution layers, and using ReZero \cite{bachlechner2021rezero} to stabilize training. Experimental results indicate that LAMA substantially enhances mask clarity and contributes to overall performance improvement, particularly in terms of visual fidelity and layout alignment.

This paper \cite{he2021context} introduces an efficient method for generating complex images comprising various objects against natural backgrounds \cite{johnson2018image,ashual2019specifying,reed2016generative,park2019semantic}. Previous approaches, such as Layout-to-Image (L2I) generation models \cite{zhao2020layout2image,sylvain2021object,sun2021learning} and Generative Adversarial Networks (GANs) \cite{park2019semantic,karras2019style}, have faced limitations including broken relations between objects and background elements, as well as poorly generated occluded regions between objects. The proposed solution addresses these limitations by incorporating a context-aware feature transformation module into the generator of an L2I model. This module updates the generated features for each object and background element by considering their relationships with other objects and elements in the image through self-attention mechanisms. Additionally, instead of using location-insensitive globally pooled object image features in the discriminator, the method employs a Gram matrix computed from the feature maps of the generated object images. The performance of the proposed method is evaluated against existing L2I models \cite{sun2019image,sun2021learning,sylvain2021object,ma2020attribute}, the pix2pix model \cite{isola2017image}, and the Grid2Im model \cite{ashual2019specifying}. Results indicate that the proposed method outperforms all compared methods across various benchmarks and evaluation metrics, particularly excelling in Inception Score and Fréchet Inception Distance (FID). Notably, among 40 evaluation sets, the proposed method achieved superior performance in 32 sets, establishing a new state-of-the-art.

This paper \cite{lin2023ocf} introduces an effective approach for accurately locating objects in complex scenes and enabling editable contiguous synthesis. It achieves this by leveraging a novel object-wise constraint-free and location-lossless layout embedding technique, which combines Variational Autoencoder (VAE) \cite{van2017neural} architecture with the recently high-performing Vector Quantized Generative Adversarial Network (VQGAN) \cite{hampali2021monte}. The approach involves utilizing a cross-VAE structure and incorporating a pre trained VQGAN to define the latent space and directly project each region of the layout onto the image latent space. Subsequently, the quantization and decoder components of VQGAN are repurposed for synthesis. This integration of VAE within the VQGAN framework, termed Layout-DVGAN in this paper, enables the preservation of spatial object locations and eliminates the need for laborious manual interaction with objects. The training process consists of two stages. Initially, a vanilla VAE is trained to regularize each feature entity into a lower-dimensional latent space, which is then encoded by the pre-trained VQGAN encoder. This is followed by the standard quantization process of VQGAN for image generation. In the second stage, the bottom pipeline is trained with a strict location constraint to ensure lossless layout embedding and Cross-VAE for aligning layout-to-image features. Overall, this method offers a robust solution for accurately locating objects within scenes and facilitating seamless synthesis, all while minimizing the need for manual intervention.

This paper \cite{cheng2023layoutdiffuse} proposes LayoutDiffuse, which adapts a foundational diffusion model pre-trained on large-scale image or text-image datasets for layout-to-image generation. By employing this technique, the results demonstrate high perceptual quality and alignment with layouts, requiring less data. The paper defines layout as either a segmentation mask \cite{park2019semantic} or a collection of labeled object bounding boxes \cite{sun2019image}. Unlike other conditional signals such as tokenized class labels \cite{ho2022classifier} or text descriptions of images \cite{ramesh2021zero}, layouts are readily obtainable and capture essential coarse-level semantic structures like the type and location of objects. In contrast to Generative Adversarial Networks (GANs) \cite{he2021context,lee2020maskgan,li2021image,park2019semantic,richardson2021encoding,sun2019image}, which serve as building blocks for layout-to-image generation models but struggle with generating reasonable images when layouts are complex, the proposed diffusion model can generate images by iteratively refining the noisy signal with a likelihood function modeled via a U-Net. Since the diffusion model is based on maximizing likelihood \cite{rombach2022high}, it does not suffer from the mode collapse problem encountered by GANs \cite{goodfellow2020generative}. The algorithm, named LayoutDiffuse, incorporates the layout signal into the diffusion model by introducing a newly designed neural adapter \cite{houlsby2019parameter,jia2022visual}. This adapter comprises two components: layout attention and task-adaptive prompts. Experimental results demonstrate that LayoutDiffuse i) produces state-of-the-art high-quality images, ii) generates more recognizable objects, and iii) is efficient in terms of time and data requirements.

When generating images from layouts, segmentations, or scene graphs, a primary challenge lies in the lack of fidelity and semantic control to specify the visual appearance of objects at an instance level, as well as the general lack of consistency, particularly in spatial equivariance. To address this issue, the paper \cite{zhao2019image} proposes a method that enables instance-level attribute control. Specifically, the input to our attribute-guided generative model is a tuple containing: (1) object bounding boxes; (2) object categories; and (3), optionally, a set of attributes for each object. The output is a generated image where the requested objects are in the desired locations and possess the prescribed attributes. In comparison to text-to-image and scene-graph-to-image generation paradigms, the layout-to-image approach offers an easy, spatially aware, and interactive means of specifying desired content. By introducing a new framework for attribute-guided image generation from layout, which builds upon and significantly extends the backbone of previous work, the paper demonstrates that a series of simple and intuitive architectural changes, including incorporating optional attribute information, adopting a global context encoder, and adding an additional image generation path where object locations can be shifted, leads to instance-level fine-grained control over object generation while improving image quality and resolution. In this case, it refers to this model as attribute-guided layout2im. Furthermore, this paper demonstrates that the model ensures visual consistency in generated images even when bounding boxes in the layout undergo translation.

This paper \cite{zheng2023layoutdiffusion}introduces LayoutDiffusion, a diffusion model comprising two components: the Layout Fusion Module (LFM) and Object-aware Cross Attention (OaCA). These elements facilitate the construction of a structured image patch incorporating region information and its transformation into a spatial layout for unified fusion with a standard layout. The approach consists of four main components: (a) layout embedding for preprocessing the layout input, (b) layout fusion module for enhancing interaction among layout objects, (c) image-layout fusion module for creating the structural image patch and implementing object-aware cross-attention tailored for layout and image fusion, and (d) the layout-conditional diffusion model with training and accelerated sampling methods. With layout guidance, the diffusion model enables greater control over individual objects while maintaining higher quality compared to prevailing GAN-based methods.

This paper \cite{zhao2019image} proposes a novel approach for generating images from coarse layouts, which consist of bounding boxes and object categories, providing a flexible control mechanism for image generation. The approach addresses the challenge of generating complex, real-world images such as those found in COCO-Stuff \cite{caesar2018coco} and Visual Genome \cite{krishna2017visual}. The model takes ground truth images and their corresponding layouts as inputs. Objects are cropped from the input image based on their bounding boxes and processed with an object estimator to predict a latent code for each object. Object feature maps are then prepared based on the latent codes and layout, and processed through an object encoder, object fuser, and image decoder to reconstruct the input image. Additionally, an additional set of latent codes is sampled from a normal distribution to generate a new image. Objects in the generated images are used to regress the sampled latent codes. The model is trained adversarially using a pair of discriminators and a number of objectives. The representation of each object in the image is explicitly disentangled into a specified part (category) and an unspecified part (appearance). The category is encoded using word embedding, while the appearance is distilled into a low-dimensional vector sampled from a normal distribution. A feature map is constructed for each object based on this representation and the object's bounding box. These feature maps are composed using Convolutional LSTM into a hidden feature map for the entire image, which is then decoded into an output image. By disentangling the representation of objects into categories and sampled appearance, the model can generate a diverse set of consistent images from the same layout. The paper presents qualitative and quantitative results on COCO-Stuff \cite{caesar2018coco} and Visual Genome \cite{krishna2017visual} datasets, demonstrating the model's ability to generate complex images that respect object categories and their layout without using segmentation masks \cite{hong2018inferring,johnson2018image}. Comprehensive ablations validate each component of the approach. Ultimately, the method can generate realistic images with recognizable objects at the desired locations.

This paper \cite{yang2022modeling} presents the application of Transformer with Focal Attention (TwFA) in exploring dependencies within complex scenes composed of multiple objects of diverse categories, aimed at improving image generation. TwFA significantly boosts data efficiency during training while enhancing both quantitative metrics and qualitative visual realism compared to existing CNN-based and transformer-based methods. The proposed TwFA framework follows a structured pipeline involving tokenization, composition, and generation phases. Initially, the image is tokenized using a VQ-GAN encoder, yielding a collection of indices representing codebook entries. Subsequently, TwFA generates final patch tokens in an auto-regressive manner, step-by-step, during the composition phase. Finally, the VQ-GAN decoder is employed to reconstruct the RGB image during the generation phase. This framework demonstrates promising advancements in image generation, providing a robust solution for synthesizing complex scenes with improved efficiency and visual fidelity.

\section{Graph}
\label{sec:graph}

\begin{figure*}[htbp]
\centerline{
\includegraphics[width= \textwidth]{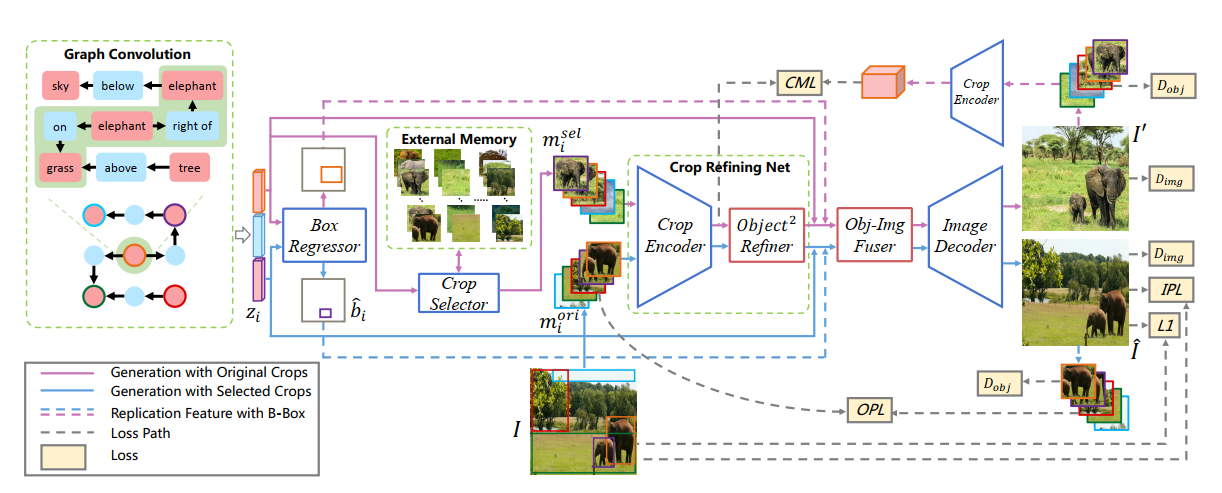}}
\caption{Flow Diagram of A Graph to Image Generation \protect\cite{li2019pastegan}.}
\label{Flow}
\end{figure*}

Most papers in the field focus on representing input as a scene graph, which describes objects and their relationships within a scene. The primary objective of these papers is to generate more realistic images from these complex scene graphs. In some cases, the scene layout, as well as the 3D scene layout, is constructed using bounding boxes and segmentation masks for all the objects in the graph. Initially, the scene graph may be converted into a feature vector for processing, while in other cases, it is used directly, with Convolutional Neural Networks (CNNs) employed for further processing. These CNNs utilize convolutional operations, often accompanied by Rectified Linear Unit (ReLU) activations, to enhance the representation. To generate images from the layout, Crop Refining Networks (CRNs) are commonly employed. These networks are supported by a crop encoder and an object refiner, which often requires an object image fuser. Overall, the use of unsupervised neural network architectures aids in producing more realistic images from the scene graph. Figure \ref{Flow} illustrates the procedural flow depicting the utilization of these models in generating real-world outputs.

For complex scene generation, managing messy layouts and avoiding object distortion are key challenges faced by most existing approaches, which primarily focus on fitting objects within an image. According to Tianyu et al., the relative proportions of objects remain consistent despite variations, prompting them to emphasize spatial layouts to preserve the semantics and relationships of objects. They converted the scene graph object feature vector and relationships and processed them through a Graph Convolutional Network. This paper \cite{hua2021exploiting} employed a relation-guided generator to handle appearances based on the relationships between objects. For their upsampling layer, they used ResBlock with the ISLA-Norm method \cite{sun2019image}. Finally, they proposed a discriminant architecture featuring a ResNet backbone with varied downsampling to maintain consistency between the scene and graph. Similar to traditional GAN-based methods, their output from ResBlocks passed through a linear layer to produce scalar outputs that identify whether the scene is real or fake.

This paper \cite{johnson2018image} introduces an efficient method for generating images based on scene graphs, which describe objects and their relationships within a scene. Despite recent advancements like StackGAN \cite{zhang2017stackgan}, which struggle to faithfully depict complex sentences with numerous objects, this paper proposes the use of scene graphs to overcome this limitation. Scene graphs have been previously employed in semantic image retrieval \cite{johnson2015image} and for evaluating \cite{anderson2016spice} and improving image captioning \cite{liu2017improved}. The primary objective of this paper is to generate complex images containing many objects and relationships by conditioning image generation on scene graphs. This allows the model to explicitly reason about objects and their interactions. However, there are three main challenges: processing the graph-structured input, ensuring that the generated images accurately reflect the specified objects and relationships, and guaranteeing realism in the synthesized images. To address these challenges, this paper employs a graph convolution network for processing the scene graph, utilizing ReLU for convolutional operations to pass information along graph edges. Subsequently, a scene layout is constructed by predicting bounding boxes and segmentation masks for all objects in the graph. Image generation from the layout is achieved using a cascaded refinement network (CRN) \cite{chen2017photographic}. Finally, realistic output images are generated by training the image generation network adversarially against a pair of discriminator networks, incorporating LeakyReLU \cite{maas2013rectifier} and batch normalization \cite{ioffe2015batch}. By comparing to leading methods that generate images from text descriptions, this approach demonstrates the advantage of reasoning explicitly about objects and relationships through structured scene graphs, resulting in the generation of complex images with numerous recognizable objects.

This paper \cite{luo2020end} introduces an efficient approach for generating 3D indoor scene layouts guided by scene graphs as input. Scene graphs serve as an abstract yet comprehensive representation of the scene's high-level attributes and relationships, aiding in the synthesis of diverse scene layouts while ensuring the relationships described by the scene graph are respected. Traditionally, unconditioned models have been employed for scene layout generation, leveraging implicit likelihood models. In contrast, this work proposes a conditional approach to layout synthesis using scene graphs. The method not only encodes object attributes and identities but also captures 3D spatial relationships within the scene. This model, named the 3D Scene Layout Network (3D-SLN), accounts for the stochastic nature of scene layouts conditioned on the scene graph's abstract description. Furthermore, the paper integrates a differentiable rendering module to refine the generated layout using only 2D projections of the scene. By leveraging 2.5D sketches (depth/surface normal) and semantic maps of the 3D scene, the model ensures a better fit between the generated layout and the target layout. The 3D-SLN combines a graph convolution network (GCN) \cite{johnson2018image} with a conditional variational autoencoder (cVAE) \cite{sohn2015learning} to model the posterior distribution of the layout based on the scene graph. The model can generate diverse scene layouts by sampling from the prior distribution, encompassing each object's size, location, and rotation. This approach extends previous methods that generated 2D bounding boxes from scene graphs \cite{johnson2018image} or text descriptions \cite{hong2018inferring} by creating complete 3D scene layouts, including 3D bounding boxes and rotations for each object. By producing not only 3D scene layouts but also 2.5D sketches and semantic maps, the model achieves higher accuracy and diversity in conditional scene synthesis. This approach allows for exemplar-based scene generation from various input forms, providing greater flexibility and control over the scene layout creation process.

The approach described in the paper \cite{li2019pastegan} involves using a model like PasteGAN to generate images with preferred objects and rich interactions based on a scene graph and object crop images. This method leverages a series of neural networks and attention mechanisms to encode spatial arrangements, visual appearance, and pairwise interactions of objects. These components are then fused into a scene canvas that is fed into an image decoder. The scene graph, which represents the relationships between objects in the scene, is processed with a Graph Convolutional Network \cite{johnson2018image}. Next, a Crop Refining Network encodes the spatial arrangements. This network is composed of two branches: a Crop Encoder (Extracts the main visual features of the object crops) and an Object Refiner (Consists of two 2D graph convolution layers that fuse the visual appearance of object crops based on relationships defined in the scene graph). Then, an Object-Image Fuser helps generate the visual appearances of the objects and their pairwise interactions into one latent scene canvas. It fuses all object integral features into the latent scene canvas. The decoder, which is based on a Cascaded Refinement Network (CRN), takes the latent scene canvas as input and generates the final output image while respecting the object positions and scene layout.To select the most compatible object crops from an external object tank and handle object appearances, a crop selector is used. Finally, to improve the realism of the generated images, a discriminator is used in conjunction with the generator (PasteGAN) in an adversarial training setup. This helps maintaining high image quality. Overall, this approach offers a comprehensive pipeline for image generation based on a scene graph and object crop images. The combination of GCN, Crop Refining Network, and Object-Image Fuser helps ensure the generated images are accurate in terms of both scene layout and object appearance, while the CRN-based decoder and discriminator help to improve the realism and quality of the final images.

This paper \cite{yang2021layouttransformer} proposes a generative model called LayoutTransformer Network (LT-Net) for text-conditioned layout generation. Previously, when translating textual inputs into layouts or images requiring explicit descriptions of each object, certain implicit objects or relationships were not adequately inferred during layout generation. In this context, the conversion of textual input into a scene graph is crucial \cite{johnson2018image}. Typically, the task of converting a scene graph to an image can be broken down into two steps: SG-to-layout \cite{herzig2020learning,jyothi2019layoutvae,lee2019neural} and layout-to-image \cite{li2019layoutgan,li2019pastegan,sun2019image,ashual2019specifying} generation. The latter aims to synthesize realistic images based on the provided layout configuration. To tackle text-to-layout generation, \cite{jyothi2019layoutvae} employs a Variational Autoencoder (VAE) to model the layout distribution of objects. Alternatively, \cite{lee2019neural,herzig2020learning} utilize Graph Convolution Network (GCN) \cite{duvenaud2015convolutional} to extract semantic information from the input scene graph and use VAE to generate the output layouts. Previously, there were challenges with avoiding possible collapse problems. To address this, following \cite{johnson2018image}, LT-Net begins with the scene graph as the input text description. To leverage implicit objects or relations in a scene graph input, a masked language model (MLM) based on BERT \cite{devlin2018bert} is employed. Regarding layout synthesis, LT-Net advances the transformer decoder, which sequentially produces bounding boxes for each object/relation. The distribution of these bounding boxes is modeled by Gaussian Mixture Models \cite{reynolds2009gaussian}. In this process, predictor P encodes the input scene graph into contextual features f, representing different semantic attributes. Generator G interprets these contextual features into layout-aware representation c, predicting bounding box information b with distributions matching learned Gaussian distribution models. Finally, the co-attention module jointly observes the generated bounding boxes and contextual representations to refine the final layout. Experiments were conducted on the COCO-Stuff dataset \cite{caesar2018coco} and the Visual Genome dataset \cite{krishna2017visual}, demonstrating the effectiveness of the LT-Net model over state-of-the-art layout generation methods both qualitatively and quantitatively.

This paper \cite{hu2020graph2plan} presents a learning framework for automated floorplan generation that combines generative modeling using deep neural networks and user-in-the-loop designs to enable human users to provide sparse design constraints. These constraints are represented by a layout graph. The core component of our learning framework is a deep neural network called Graph2Plan, which converts a layout graph and a building boundary into a floorplan that fulfills both layout and boundary constraints. The network processes the layout graph using a graph neural network (GNN) and the input building boundary and raster floorplan images using conventional neural networks (CNN). If a user wants to design their building according to specific room counts or connectivity, these constraints are best represented by a layout graph, similar to scene graphs for image composition. The model uses the large-scale floorplan dataset RPLAN \cite{wu2019data} to fulfill both layout and boundary constraints. The process begins with the user entering a building boundary and a set of constraints to guide layout retrieval and floorplan generation. To obtain layout graphs, the proposed system uses a retrieve-and-adjust approach based on the RPLAN dataset. This approach incorporates human design principles from real floor plans. The goal is to generate a new floor plan that instantiates the layout graph within the input building boundary. To achieve this, the authors train Graph2Plan to perform the retargeting based on design principles embedded in the floorplan dataset. The final output of Graph2Plan is a raster floorplan image and one bounding box for each room. A potential issue with the output boxes is that they may not be well aligned, and some boxes may overlap in certain regions. To address this, in the final vectorization step, they use the raster floorplan image to determine room label assignments in the regions with overlap.

In the \cite{wang2022interactive} paper, the authors propose the Panoptic Layout Generative Adversarial Network (PLGAN) to address issues such as missing regions in constructed semantic layouts and undesirable artifacts in generated images. These challenges remain significant, as existing models struggle with reasoning object locations and relations \cite{zhang2017stackgan}. Image synthesis from semantic layouts offers a promising approach to enhance computer-human interaction \cite{hong2018inferring,li2019object} and produce aesthetically pleasing results \cite{park2019semantic,wang2018high,zhu2020sean}. Grid2Im \cite{ashual2019specifying} introduced a scene-graph-to-image synthesis approach consisting of two stages: layout construction (creating an instance layout with per-object masks and bounding boxes) and image generation (where the instance layout is converted into a photo-realistic image). This approach is based on panoptic segmentation theory \cite{kirillov2019panoptic}, which divides objects into countable "things" and uncountable "stuff." Then, develop a panoptic layout generation (PLG) module, which includes a stuff branch for stuff layout construction and an instance branch for instance layout construction. The instance and stuff layouts are combined into a panoptic layout using an instance- and stuff-aware normalization (ISA-Norm) module. This module eliminates missing regions and improves the robustness of the layout generation process.

To address the challenge of generating complex scenes with rich primitive concepts, this paper \cite{deng2018probabilistic} proposes the Probabilistic Neural Programmed Networks (PNP-Net), a probabilistic modeling framework that combines and generates scene images based on text-based programs (or other language forms like sentences). This approach leads to compositional models of appearance that can be utilized across various scenes. The proposed model consists of two core components. Mapping Functions: These functions take either semantic concepts or distributions as input and generate distributions over a latent space, capturing their combined meaning. Probabilistic Modeling Framework: This framework performs inference and learning using the latent space. And the model has four core modules: (a) Combine: Takes two distributions and generates a compound distribution as output.(b) Describe: Models the interaction between distributions and generated output distributions.(c) Transform: Instantiates the size of a spatial appearance distribution and performs bilinear interpolation to generate a new appearance distribution.(d) Layout: Places two appearance distributions according to sampled offsets on a canvas and renders them using convolutional neural network (CNN) layers. The model takes a program derived from text concepts, performs compositional generation in the latent space, then decodes the latent code into image space. During training, an encoder (Reader) acts as a proposal distribution for learning the latent space. The Describe and Transform modules work together to emphasize that a Transform module is applied following a Describe module in the formulation.

This paper \cite{kumar2022graph} presents a novel framework for learning an effective strategy to explore cluttered environments by training scene generation and scene exploration agents that share a graphical representation of the scene based on a scene grammar. The framework uses scene grammar to represent the space of possible structured clutter scenes and generate stable scenes with hidden objects. The Scene Generation Agent is trained using deep reinforcement learning (deep RL) to create complex scenes that hide more objects. The Scene Exploration Agent is trained to explore cluttered scenes efficiently with minimal interactions. The training process adopts a "learning by cheating" framework \cite{chen2020learning}, which first trains a teacher agent with privileged information and then clones the learned behaviors to a student agent. The agents are modeled as message-passing neural networks (MPNNs) \cite{zhou2020graph} and trained with RL to maximize their respective objectives. The Scene Generation agent focuses on hiding objects in the scene, while the Scene Exploration agent tries to locate them. Each message-passing step consists of a message-passing layer followed by Global Attention \cite{li2015gated} with a skip connection. Finally, the framework is tested in real-world, cluttered environments, demonstrating successful sim-to-real transfer of the Scene Exploration Agent.

This paper \cite{ivgi2021scene} introduces a novel method for generating images from scene graphs by gradually producing the entire layout description to improve inter-object dependency. This approach addresses photorealism, correctness, and diversity issues. To tackle diversity challenges, the method shifts from supervised losses to adversarial ones, reducing dependence on supervised losses. Additionally, it introduces a new approach for high-resolution layout generation, which uses Graph Convolutional Networks (GCNs) \cite{kipf2016semi} to process variable-shaped structured graphs and contextualizes the state of all objects with CNN-based generators. The paper proposes the Contextualized Objects Layout Refiner (COLoR), the first model that directly generates layouts from scene graphs (SGs) without any intermediate steps such as boxes and masks. The method stacks multiple copies of a block to enhance the model's performance. COLoR achieves significant improvements in layout quality compared to previous works, resulting in accurate and photo-realistic images.

This paper \cite{savkin2021unsupervised} introduces a method for directly synthesizing traffic scene imagery without rendering, using domain-invariant scene representation and synthetic scene graphs as the internal representation. To produce realistic images, the approach uses an unsupervised neural network architecture. The challenge of costly data annotation and acquisition has led to the use of synthetic datasets \cite{ros2016synthia,richter2017playing,wrenninge2018synscapes} and simulation systems \cite{shah2018airsim,dosovitskiy2017carla}. This paper overcomes domain differences by eliminating the rendering part of the pipeline. Instead, it replaces rendering with an abstract scene representation, enabling the direct synthesis of realistic traffic scene images. The proposed method leverages synthetic 3D scene graphs with spatial components and supports realistic traffic scene generation from synthetic scene graphs through an unsupervised neural network architecture. By using unsupervised learning techniques, the method bypasses the need for expensive labeled data and renders photorealistic images directly from scene graphs. This process streamlines traffic scene image synthesis and enhances the efficiency and quality of the output.

\section{Hybrid/Multimodels}
\label{sec:hybrid}
\begin{figure*}[htbp]
\centerline{
\includegraphics[width= \textwidth]{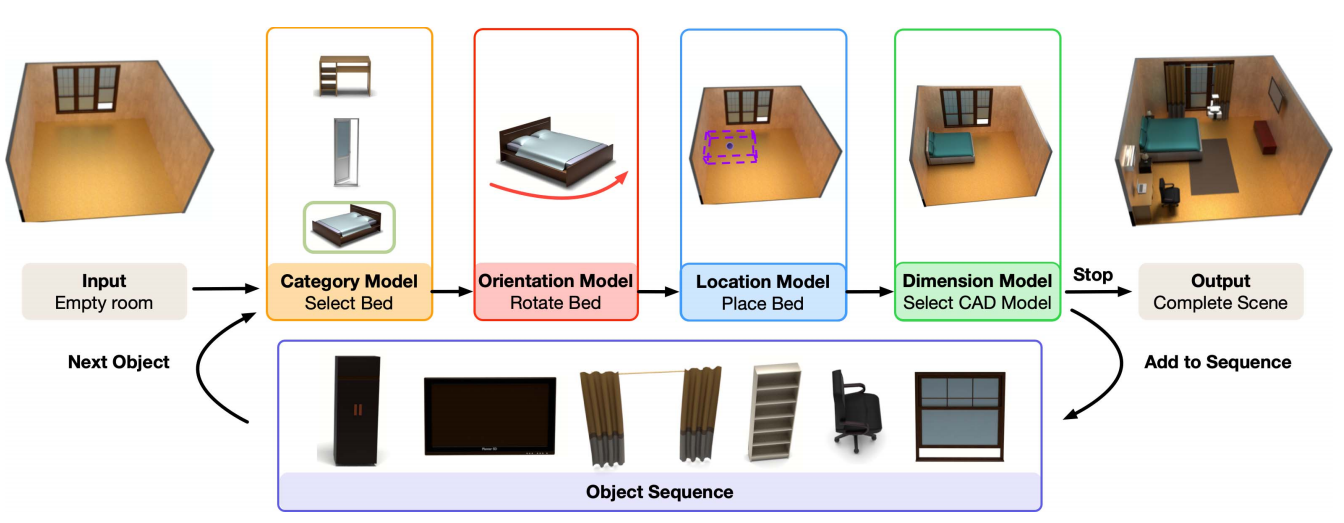}}
\caption{Flow Diagram of A Hybrid System \protect \cite{wang2021sceneformer}.}
\label{Flow diagram of SceneFormer.}
\end{figure*}

Many 3D scene generation systems follow text-to-image generation methods that show much capability in generating images for different applications. But many systems need more control by the user which is missing in these text-to-image generation models. Also, text can be represented in a vast number of synonyms words which may not be available in the training data used to develop those models. These limitations can produce erroneous outputs, That is why, most hybrid systems mainly combine two types of input to generate the final 3D scene or image. Fig \ref{Flow diagram of SceneFormer.} shows the flow diagram of a hybrid system called SceneFormer that takes two types of input to generate an indoor 3D scene. The inputs are mainly text descriptors and scene layout. The authors worked with different approaches for generating the scene. Mostly, the authors used a pre-trained method called Stable Diffusion and updated image generation by modifying the cross-attention layer, whereas in other cases the authors used transformers and classifier-free guidance for predicting the scene attributes. Some papers introduced different versions of the GAN network like ProGAN and GauGAN for predicting the semantic class and layout of the urban scene to generate the final image. However, the authors in one paper used a feed-forward neural network for generating the 3D scenes. Most of the hybrid models are implemented to create 3D scenes or images of the indoor environment like living rooms, bedrooms, etc. As the hybrid model is complex, most papers focused on a fixed type of scene instead of a general environment in the outdoors. Lack of available and appropriate datasets can be a reason for this.

In \cite{chen2024training} the authors proposed a layout-guided text-to-image generator where the layout can be controlled without further training or finetuning. The authors used the default pre-trained method called Stable Diffusion (SD) \cite{rombach2022high} which is publicly available as text to text-to-image generator using diffusion. They implemented two layout guidance methods that modify the cross-attention layer of the image generator. The first one is forward guidance which is already used in previous work \cite{balaji2022ediffi,singh2023high}. Second, one is backward guidance which is mainly the major contribution of the paper. Forward guidance directly manipulates the cross-attention layer to shift the activations to a desired pattern, whereas backward guidance mainly backpropagates to update the image latents so that they can match with the desired layout via energy minimization. The authors compared the performance of the forward and backward guidance method based on three benchmarks: VISOR \cite{gokhale2022benchmarking}, COCO 2014 \cite{lin2014microsoft}, and Flickr30K Entities \cite{plummer2015flickr30k} \cite{young2014image}. The authors also compared their works and their method gives better performance than others, especially based on the VISOR metric.

To maintain controllability, human perception quality and resolution control of the generated image Gafni et al. introduced a model that can generate scenes based on text prompt and scene layout(segmentation map) \cite{gafni2022make}. Through their text prompt, they can control the style, color, structure, and arrangement of the object. As a result, they can edit scenes, overcome unordered prompts, and generate story illustrations. As part of their improvement they modified the tokenization procedure, human preference in token space scene-based transformer, and classifier-free guidance. To generate a token form text prompt they used BPE \cite{sennrich2015neural} From segmentation they used the VQ\_SEG, a modified GQ-VAE, network and created three groups - panoptic, human, and face. To emphasize on face parts like eyes, nose, lips, eyebrows they employ a weighted binary cross-entropy face loss over the segmentation. Their model collects the token from text, scene, and image and uses it in an Auto-Regressive vene-Based Transformer to generate the final scene.

The paper in \cite{le2021semantic} developed a model for controlling the distribution of semantic classes in every pixel in a layout to an image generation model for urban scenes. The authors first take ProGAN \cite{karras2017progressive} as the layout generator. They provided a conditioning code for specifying the class distribution of the target alongside with a random noise as the input to the network. They also proposed a new semantically-assisted activation (SAA) module to propagate the constraints from the previous scale to the next and also a residual conditional fusion between adjacent subnetworks. They proposed the SAA module with two objectives: conditional objective and adversarial objective. Finally, they used GauGAN \cite{park2019semantic} for translating the layout to the photorealistic image. The authors also extended their framework for editing the subregions in existing semantic layouts partially. The authors used three datasets for evaluating their proposed framework: Cityscapes\cite{cordts2016cityscapes}, Cityscapes-25k, and Indian Driving Dataset (IDD) \cite{varma2019idd}. They also compared the performance of the framework with other state-of-the-art methods based on different types of metrics. The proposed framework outperformed other baseline works based on the evaluation metrics.

Wang et al. proposed an indoor 3D scene generation system called SceneFormer using a set of transformers \cite{wang2021sceneformer}. The authors developed the scene based on a sequence of objects which are predicted using transformers. The system can take two types of input: room layout or text descriptors. For room layout, the transformer decoder predicts the class, location, size, and orientation of each object of the scene cross-attention mechanism. Then,  the CAD model of each object is retrieved from the dataset based on class and size to place in the predicted location of the scene. For text descriptors, the authors used sentences describing the scene and tokenized them to generate a token sequence of a maximum of 40 tokens in length. Then they used GloVe \cite{pennington2014glove}, ELMo \cite{peters1802deep}, and BERT \cite{devlin2018bert} to embed each word using the Flair library \cite{akbik2018contextual} and 2 layer MLP to convert the dimension and find the text conditioned models. They used a series of transformer decoders for the location of category prediction of each text condition model. For training and evaluating the method, the authors used three datasets: ModelNet \cite{wu20153d}, ShapeNet \cite{chang2015shapenet}, and human-created scene datasets with synthetic objects \cite{song2017semantic}. The method is compared with other state-of-art work like DeepSynth \cite{wang2018deep}, PlanIT \cite{wang2019planit} and FastSynth \cite{ritchie2019fast}. The proposed method is faster and improved than other existing methods.

Xue et al. developed a system called Freestyle LIS (FLIS or FreestyleNet) \cite{xue2023freestyle}. It is mainly a layout-to-image synthesis method that can generate images of classes that are not present in the training dataset. There are two types of input in the system: layout and text. Layout is used to control object shapes and positions and text to define classes, attributes, and styles of the objects. The authors used Stable Diffusion \cite{rombach2022high} which is a pre-trained text-to-image diffusion network. The model has a text encoder called CLIP to extract text embeddings and uses denoising U-Net to generate the final result. The authors proposed a module called Rectified Cross-Attention (RCA) which is plugged into each cross-attention layer of the U-Net. The proposed module mainly integrates the input layout into the image generation process. The authors used two datasets for evaluation: COCO-Stuff and ADE20K. They compared their model with other state-of-art baselines and their model works better based on Frechet Inception Distance (FID) and Mean Intersection-over Union (mIoU).

In the \cite{zhang2020deep}, Zhang et al. proposed an approach for generating indoor 3D scenes using a feed-forward neural network. The network uses a random sample of objects from a normal distribution of the latent space as input and generates a scene as an arrangement of objects. They solved the challenges of implementing the network by different techniques. Firstly, each scene is generated using a subset of objects from the superset of the abstract object in the dataset and the network determines the geometric shape, size, orientation, and location of each object. They mainly introduced a sparse dense generative network to solve the overfitting of the network. Then they train the network using two types of loss function. The first one is standard VAE-GAN \cite{larsen2016autoencoding} and the second one is to project the scene into an image domain and capture the relation of adjacent objects by using a discriminator in the convolution layer of the network. Finally, they proposed latent pose variables to train the instance and optimize the instance together with the generator. The authors used a part of the SUNCG \cite{song2017semantic} dataset having only bedroom and living room scenes for training and evaluating the method they proposed. They also compared their method with other baseline methods qualitatively and their proposed method performed comparatively better than other baseline methods.

\section{Mask to Image}
\label{sec:mask to image}
Scene generation based on semantic segmented mask is another popular approach for image generation. Here object type, location, size, shape, and interaction with other objects are predefined from the input mask. Only need to predict the texture, with a meaningful, consistent, and realistic way. Most of the authors utilize GAN architecture \cite{azadi2019semantic}\cite{gao2020sketchycoco}\cite{drew_Gene} with different variations to achieve new scenes from the mask. All instance segmentation datasets can be used as training data for this process.

\begin{figure*}[htbp]
\centerline{
\includegraphics[width= \textwidth]{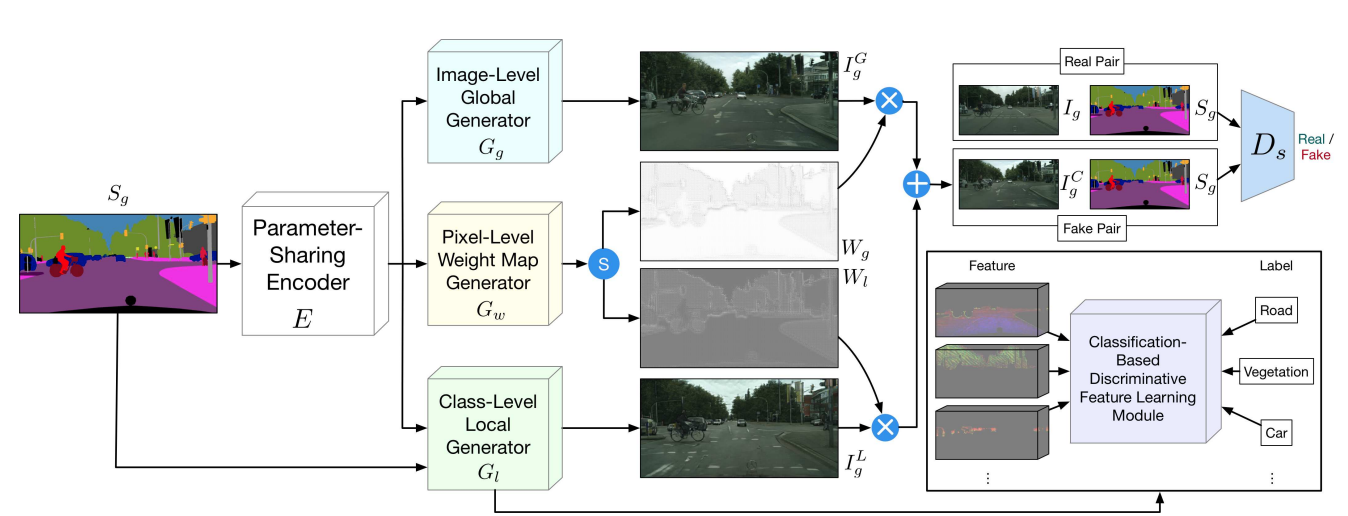}}
\caption{Flow Diagram of A Mask to Image System \protect \cite{tang2020local}.}
\label{fig:mask_to_image}
\end{figure*}

Azadi et al \cite{azadi2019semantic} tried to generate the texture of the complex scene when pixel-wise semantic segmentation of that image is given. They proposed an unconditional image generator called semantic bottleneck GAN (SB-GAN) for this purpose. They begin with an unconditional progressive segmentation generation network to estimate pixel-wise semantic segmentation from scratch. They created this layer following the progressive GAN architecture\cite{karras2017progressive} and accompanied by improved WGAN loss function\cite{gulrajani2017improved}. Then they utilized the conditional segmentation-to-image synthesis network to generate images from semanic segmentation. They created this layer by taking advantage of the Spatially-Adaptive Normalization \cite{park2019semantic} model. As the SB-GAN model is divided into two stages, it is possible to i) maintain shape at the semantic segmentation generation state and ii) generate meaning full texture at image generation. According to their claim, they achieve state-of-the-art performance on the Cityscapes\cite{cordts2016cityscapes} and ADE-Indoor\cite{zhou2017scene} datasets with this model.

Rather than generating images in a single shot, Turkoglu et al. used their model to generate sequentially\cite{turkoglu2019layer}. They started with the background, then added foreground, and then populated the scene one by one. They control the object of the scene using external conditions such as pixel-by-pixel semantic segmentation.  They divided the background generation process into two branches. The first branch generates a background image from a noise vector. The second branch used the scene generated from the first branch and a semantic segmented map like an image translator mentioned in Pix2Pix\cite{isola2017image}. For foreground generation, it uses a noise vector, object mask, and background as input. They used Zhu et al\cite{zhu2017unpaired} for noise vector representation.

Gau et al. tried to generate complex images from freehand sketches in their SketchyCOCO research \cite{gao2020sketchycoco}. In their research, they contributed in two ways. As a technical contribution, they proposed an attribute vector bridged Generative Adversarial Network called EdgeGAN, and dataset contribution they created a large-scale composite dataset called SketchyCOCO based on MS COCO Stuff \cite{caesar2018coco}. They divided their scene generation work into two stages. The first stage is foreground generation as these objects are defined by the sketch artist. They located each object using the sketch segmentation method[2]. The second stage is background generation and filling the empty space in the scene using pix2pix \cite{isola2017image}. They trined their foreground and background generation methods separately.  They started their EdgeGAN by converting the sketch image into embedding into shared latent space, then translated this data as an image. For this, they only forced on focal loss \cite{lin2017focal} and skipped the adversarial loss. The architecture of EdgeGAN is similar to WGAP-GP \cite{gulrajani2017improved} with Edge Encoder like BicycleGan \cite{zhu2017toward}. 

In general semantic label maps are used once for generating new images, This might lose some information to mitigate this SPADE \cite{park2019semantic} used activation in the normalization layer.  Also, different convolutional kernels can capture a variety of information in the convolutional layer of the discriminator.  Based on these two ideas, Liu et al. propose to add condition l from the segmented label map on the convolutional kernel and propose a feature pyramid semantics-embedding discriminator to generate scene \cite{liu2020learning}. According to their claim, they have achieved state-of-the-art performance on Cityscapes, COCO-stuff, and ADE20K datasets.

Hudson et al. used an object-oriented transformer with GAN architecture for their GANformer2 model for scene generation. This is a modification of the GANformer \cite{drew_Gene} model. They divided their image generation task into two stages: sequential planning and parallel execution. They started the planning stage using a latent variable, they converted it to a schematic layout. This layout contains information about the semantic class, position, shape, order, and unsupervised depth orientation. In the planning stage, they followed the Recurrent Layout Generation, Object Manipulation, Layout Composition, and Noise for Discrete Synthesis stages. The execution stage uses this information and applies bipartite attention \cite{drew_Gene} to guide the content and style to generate photo-realistic images. In this stage, they used Layout Translation, Layout Refinement,  Semantic-Matching Loss, and Segment-Fidelity Loss. According to the author’s claim, GANformer2 achieved high versatility over different datasets.

In their study\cite{DBLP_Yang}, Yang et al. tried to demonstrate the internal relationship that appears in the deep generative representations from state-of-the-art GANs trained for scene synthesis, such as StyleGAN \cite{8953766} and BigGAN \cite{brock2018large}. They are able to quantify the causal relationship between the layer-wise activations and the semantics contained in the output image by probing the per-layer representation with a wide range of semantics at various abstraction levels.  Changes in illumination and scene viewpoint are two examples of human-understandable variation elements that may be quantified and further used to guide the generating process. To quantify the relationship between different layers, they proposed a rescoring technique. Their techniques have some limitations like i) lack of thorough and precise off-the-shelf classifiers ii) only linear SVM is used for boundary search. iii) Only limited to visual scene understanding.

Tang et al. focused on generating detailed scenes including information on small objects and detailed local textures from semantically guided scenes \cite{tang2020local}. They designed a Local class-specific and Global image-level Generative Adversarial Network (LGGAN)  to consider the local context from semantic segmented images. This generator has three sub-generators that use the same encoded feature map from the input image. The first one is the global generator, it learns the global appearance of the object in the image. Then the second one is a class-specific local generator, which focuses on generating individual objects. The final one is the fusion weight-map generator, which combines the rest two generators. In the training stage, they combined these methods and trained end-to-end fashion. In that research, they mainly focused on semantic image synthesis and cross-view image translation. For the input encoder they followed the style of GauGAN \cite{park2019semantic} and SelectionGAN \cite{tang2019multi}.  The model architecture for this approach is given in Fig \ref{fig:mask_to_image}.

\section{Interactive scene}
\label{sec:interactive scene}
\begin{figure*}[htbp]
\centerline{
\includegraphics[width= \textwidth]{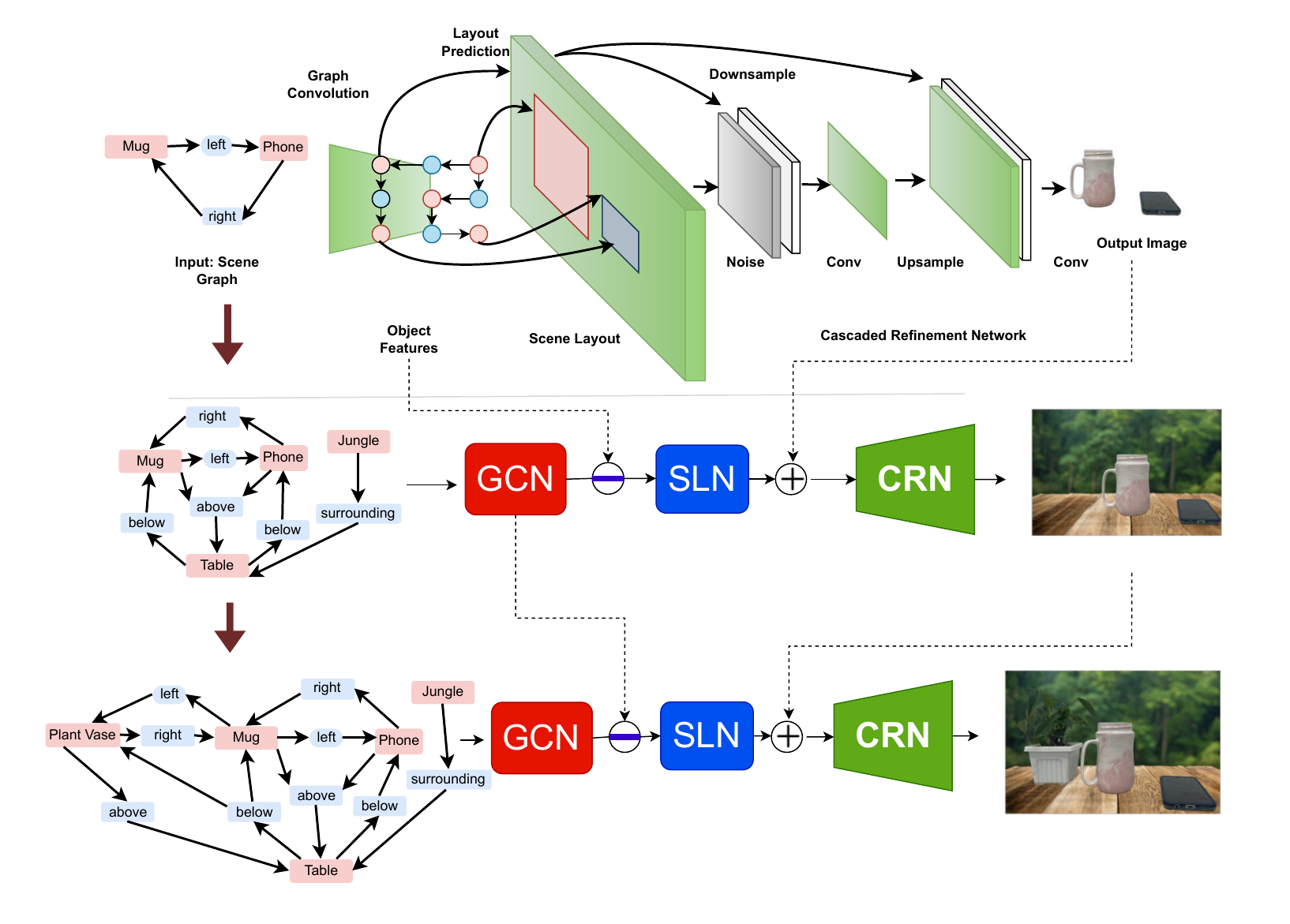}}
\caption{Flow Diagram of A Interactive System \protect \cite{tang2020local}.}
\label{fig:intractive}
\end{figure*}

Most of the current models generate model images from direct input. The user does not have much space to modify some part of this if he/she does not like it. The user has to give a full prompt again and also there is no surety of getting the expected image as it is giving one image for one input. That is where we need an interactive way to modify the full or some part of the image. An example of an interactive approach for image generation is given in figure \ref{fig:intractive}

Ashual et al. prepared a model to generate scenes interactively \cite{ashual2019specifying}. The method separates between a layout embedding and an appearance embedding and a combination of these generates high quality and complex images. They showed two ways of interaction i) importing an object from other images and ii) navigation in the object space, by selecting an appearance. They represented the scene in the graph for further processing. This graph has 3 parts i) an encoded vector of object type (ii) a coarse 5 × 5 grid and size of the location of the objects (iii) the appearance embedding. The age of the graph denotes “right of”, “left of”, “above”, “below”, “surrounding”, and “inside” from the related objects. The steps the author followed for this work are (i) A graph convolutional network to convert scene graph to object embedding (ii) A CNN to generate the object mask from embedding where the Least Squares GAN (LS-GAN \cite{mao2017least}) was used as discriminator (iii) A parallel network to convert the embedding to a bounding box (iv) An appearance embedding CNN that converts image information into an embedding vector (v) A multiplexer to combine the object masks and the appearance embedding into one multidimensional tensor,  (vi) An encoder-decoder residual network to generate the scene image and compared by \cite{johnson2016perceptual} with real image. This method is mostly related to \cite{johnson2018image} without human interaction part and some minor changes.

Mittal et al. tried to generate images through continuous interaction with users \cite{mittal2019interactive}. They used a Graph Convolutional Network(GCN) with a Generative adversarial image translator to convert the graph into an image. For their baseline architecture, they used the architecture given by \cite{johnson2018image}, where i) a Graph convolution neural network: that takes a graph as input and represents as a vector for each node and edge ii) a Layout Prediction Network: Takes the vector from the GCN and generates the layout of the image and iii) Cascade refinement Network(CRN): converts the layout to actual images. Then it comes to the part of generating sequential images. For this, they change the noise passer of CRN to the previous RGB image passer. This RGB image from the previous step also guides image generation in the next step. But there is a huge issue with this system, it only takes the same graph with the previous image as input. As a result, it is not possible to generate new images according to user requirements. In the future, any form of natural language can be taken as input for this.

On the other hand, Ashual et al. \cite{ashual2020interactive} introduced a dual embedding to change the size, location, and appearance of the object based on one external input independently. Initially, they took the layout graph and encoded it in a neural network. The first part of the encoder is placement, relative position and other global image features. The second part is the appearance of the object. The neural network has multiple subparts i) A graph convolutional network: to convert graph to object embedding ii) A CNN: converts the local embedding to object mask iii) A parallel network: generates local embedding to bounding box location to place mask, iv) An appearance embedding CNN: to convert image information to embedding, v) A multiplexer: to combine the object masks and appearance embedding and finally vi) encoder-decoder residual network: to generate an output image. This architecture adds stochasticity before mask creation. So it is possible to manipulate the graph easily.

Interactive narration can generate the trigger path of the story. To elaborate on this idea, Kumaran et al. proposed a model SceneCraft \cite{kumaran2023scenecraft} to narrative about scene objectives, non-player character (NPC) traits, NPC location, NPC background and NPC narrative variations using a Large language model. For their narration, they used Ink (https://www.inklestudios.com/ink) dialogue script and then read by StoryLoom Engin \cite{mott2019designing}. They used LLM (GPT 3.5) to process the prompt given by the template. Then a BranchingDialogue Graph Generator combines the dialogue scripting of any narrative variations. This model can generate multiple dialog variations for each utterance, emotion, and gesture. In the model, it is possible to incorporate 6 emotions and 39 gestures. There are two main issues with this model i) they mainly depend on GPT 3.5 and don’t consider modifying it according to their need and ii) to explain internal relations they need a high-level plot than their current plot.

Virtual reality uses 3d scenes for games, movies, and other activities. To help virtual reality in 3D Chen et al. proposed EntangleVR \cite{chen2021entanglevr}, which generates scenes based on human interaction by focusing on the complex relationship between objects and entities. EntangleVR provides a method to create an entanglement-based virtual programming interface where users can select and create nodes, compose graphs, and interact with other objects. Users can participate in this entangled environment and produce an observer effect \cite{billinghurst19973d}. EntangleVR creates the environment by utilizing nodes that contain the object instances, logical statements, and control events. As a result, it becomes a relational graph. They used 7 types of nodes for that: Super Object, Qubit, Gate, Observer, Entangler, Super Location, and Avatar. Where the Qubit represents the basic unit of computation, the Super Object represents the higher-level mapping of a qubit’s computation, the observer node provides an interaction method to measure and collapse the linked super object, the gate node provides a selected set of common single-qubit quantum gates and finally the entangler is built to simulate the unique quantum phenomenon of entanglement that is not available.

Cheng et al. proposed a model to generate composite images step by step \cite{cheng2019interactive}. They took natural language as input and followed different steps like natural language processing, spatial layout generation finally scene generation for final image scene generation. In the segmentation layer, they used instance segmentation, applied a discrete event system (DES), and employed supervisory control to control unexpected events from data. The natural language processing pipeline has 2 sub-modules i) semantic roles are labeled to identify verbs using AllenNLP \cite{he2017deep} ii) the argument is parsed to identify properties of objects and spatial relations using Stanford Log-linear POS Tagger \cite{klein2003accurate}. Spatial layout generator try to generate a reasonable layout for the image. The scene generator tries to create the actual scene and add objects cascading according to the layout. Finally, the motion planner takes the generated scene as input, and outputs of the robotic painter to draw the scene in 2d. In the context of supervisory control, a DEDS is modeled using a five-tuple automaton. The tuples are states abstracted from the system, the set of events, transition function, initial state and start set like context-free grammar. In this research, events are divided into 2 types: controllable and default.

\section{Semi Supervised}
\label{sec:semi supervised}
\begin{figure*}[htbp]
\centerline{
\includegraphics[width= \textwidth]{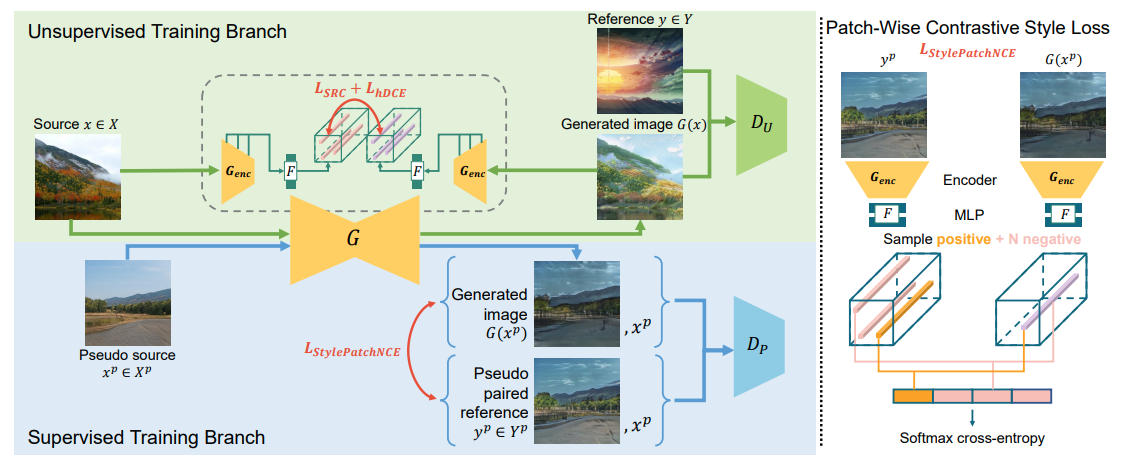}}
\caption{Flow Diagram of A Semi Supervised System \protect \cite{kipf2016variational}.}
\label{fig:semi_supervised}
\end{figure*}

Missing enough properly labeled data is one of the key challenges for any kind of supervised machine learning algorithm. The most efficient solution comes when it's possible to complete the task with unsupervised learning. But it is not always possible. An efficient solution can be labeled with a few data to at least guide the model in the right direction so the model can learn itself from the rest of the data. Also for scene generation, if someone tries something new other than the popular approach he will face the same issue with data. Some of the researchers solved it by using semi-supervised methods. An approach of image generation in semi-supervised approach is given in figure \ref{fig:semi_supervised}.

Lack of details, unable to understand semantics, and inconstant styling are some key problems for high-quality animated scene generation. To resolve these, Jiang et al. \cite{jiang2023scenimefy} proposed a new architecture called Scenimefy for unsupervised image-to-image translation. They used StyleGAN \cite{karras2020analyzing} guided by CLIP \cite{radford2021learning} and VGG \cite{simonyan2014very} to capture complex features and generate real and animated image pairs. They used this CLIP to reduce the cosine distance between two images. To maintain overall semantics they used a perceptual loss \cite{zhang2018unreasonable}. They used Mask2Former \cite{cheng2022masked} as a segmentation-guided filter to remove low-quality images. For Scenimefy they used only the high-quality data to map pixel-by-pixel comparisons between real and generated images. This step is done by a patch-wise contrastive styled loss inspired by CUT \cite{park2020contrastive}. They divided their training process into two steps. For the supervised training procedure, they used a conditional GAN framework \cite{mirza2014conditional}. For the unsupervised training stage, they used the approach proposed by Jung et al. \cite{jung2022exploring} to tackle heterogeneous relations in semantics between two images. This model still lacks external control to manage the degree of style needed to change.

To improve modality information learning and reduce full dependency on training data Li et al. proposed an end-to-end semi-supervised cross-modal image generation method \cite{li2020semi}.  This method uses two semantic networks with an image generation network. To learn semantic information they used a deep mutual learning strategy. Among the two semantic networks, one is image modality with a Convolutional neural network from AlexNet \cite{krizhevsky2012imagenet} and another is a Deep neural network for non-image modality. The reason for using two modalities is that they might provide complementary important information that improves the model \cite{chen2017deep}.  Image generation is conditioned by semantic features and labels generated from semantic networks. It uses a deep convolutional conditional generative adversarial network to generate images. Their image generator has 3 parts: 1 generator and 2 discriminators. One of the discriminators is used to determine the difference between the real image and the generated image. The second one is used for reducing the reconstruction loss. 

Dai et al. \cite{dai2021spsg} in their SPSG, presented an approach to generate a 3D model of the real world from an incomplete RGB-D image using a self-supervised approach. This self-supervise takes a 2D signal provided by RGBD frames as input as these frames are high resolution and self-consistent. Then they reconstruct the 3D model from this 2d image as a Truncated Signal Distance Function (TSDF) using volumetric fusion \cite{curless1996volumetric}. From this TSDF, final mesh can be extracted using Marching Cubes \cite{lorensen1998marching}. This approach helps to train on incomplete real-world data. As their semi-supervised approach, they removed some frames from their RGB-D frame and learned to generate those through the training process, ignoring the unobserved places through came. To maintain sharpness, high-quality image shape, and color, they added a 2d adversarial and 2D perceptual loss. However, this model fails to generate a very large-scale image.

Toutouh et al. \cite{toutouh2023semi} used a coevolutionary algorithm with Semi-Supervised Learning (SSL) for their image generation process. Their model can also work as a classifier. They extended the Lipizzane \cite{schmiedlechner2018lipizzaner} by combining the SSL-GAN with coevolution and gradient-based learning. The generator loss of SSL-GAN is the same as any other GAN. But for the discriminator, the loss function is modified to discriminate between generated images and unlabeled images. They added majority voting classification, spatial diversity loss, and balanced label batching samples with their model. Balanced label batching sample has two rules i) Data labels are not changed in each batch ii) Classes are balanced. As each batch, all images are the same, they used a majority voting classification scheme to label that batch based on the most frequent label for that batch. The Spatial diversity loss is to balance the weighted sum of supervised and unsupervised loss.  They call this model Lipi-SSL \cite{hemberg2021spatial} and it is mainly designed for fully unsupervised learning. With their model, they have achieved 96\% accuracy with MNIST training data after using only 100 labeled data among 60000 data points, which is only 0.167\% of total data. As the labeled samples are fewer they train their model in two phases. In the first stage, it trains with only the sigmoid layer, and in the second stage, it trains with the softmax layer.

Pavllo et al. used a weakly-supervised approach to generate images where they rely on the semantic segmentation for object shapes and classes in a complex image \cite{pavllo2020controlling}. They used masks and text as input to control the image generation process. The shape and classes of the object are controlled by the mask created by Mask R-CNN \cite{hu2018learning} and the style and textures are controlled by the text generated by using BERT base \cite{devlin2018bert}. They introduced a semantic attention module to add a textual description as a condition in their model. As a result, they can generate scenes even after moving, scaling, or deleting the foreground object. They also used a large vocabulary object detector as an annotator to use unlabeled data in the training. They divided their process into two steps. In the first step, they generated the background of the image. In the second step, they generated the foreground object based on the background when the background was kept frozen. For their conditional mechanism, they kept SPADE\ cite{park2019semantic} and used a multi-scale discriminator like \cite{wang2018high}. The main issue with this model is that it can not differentiate instances of the same class and does not access the position information.

Katsumata et al. tried to create an open-set image generation using Conditional GANs with a semi-supervised process \cite{katsumata2022ossgan}. For open set data the newly generated image does not belong to any predefined label. They used the entropy regularization process to quantify the sample-wise importance of GANs condition to allow using unlabeled data. To determine the similarity between images during training, they designed a model called Open-Set Semi-supervised GAN (OSSGAN). They independently train a conditional GAN for image generation where OSSGAN integrates the unlabeled data in the conditional GAN. They started by using 2 GAN models. First is RejectGAN which deals with only the closest of the labeled data and unlabeled data that has a confidence, of being one of the closed sets, higher than a threshold. The second one is OpensetGAN which deals with lower confident unlabeled data, it assigns new classes to those data. But the threshold-based model is not good to use, because this value may change based on the dataset, and also confidence changes after each iteration. So, they moved to their own created GAN architecture OSSGAN. Along with the generator and discriminator like other GAN, it has an additional auxiliary classifier. They used this classifier label as the original label. The main issue with this technique is that they mostly rely on the auxiliary classifier. If the classifier gives an incorrect answer, the full system will fail.

Fidelity, diversity, and controllability are the main challenges of GAN-based model. Bodla et al. \cite{bodla2018semi} tried to maintain all of these using their FusedGAN. FusedGAN is a combination of two GAN models. One is for unconditional image generation and another is for conditional image generation. Both of them use the same latent space. Unconditional GAN is mostly responsible for generating the shape of the object and it is fully independent from the conditional part. Conditional GAN is responsible for generating texture, and color and tries to maintain the condition provided by the input pipeline. As the unconditional GAN can generate shape in a single stage, it can be effectively trained with semi-supervised data. The main challenge with this architecture is that the unconditional GAN can generate inconsistent and unrealistic structures for unlabeled data, which can reduce the performance of the full system.

\section{Text to image}
\label{sec:text to image}
\begin{figure*}[htbp]
\centerline{
\includegraphics[width= \textwidth]{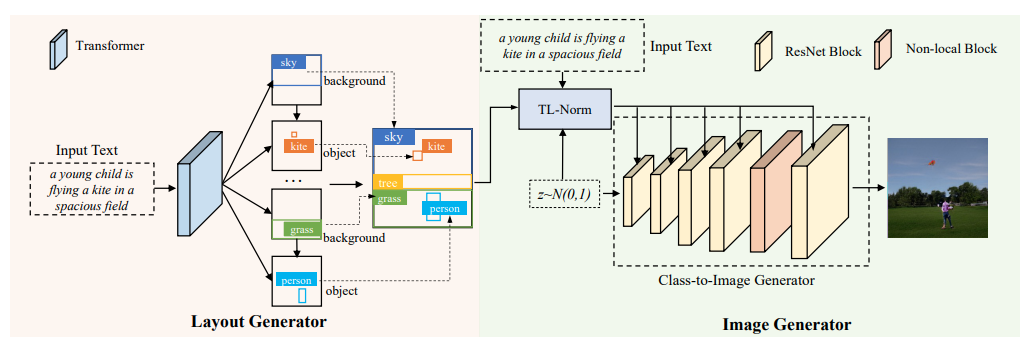}}
\caption{Flow Diagram of A Text to image generation System \protect \cite{chen2022background}.}
\label{fig:text_to_im}
\end{figure*}

Recent GAN-based models for scene generation mainly face two problems i) background scenes are unrealistic and ii) objects are distorted or miss key points. Chen et al. proposed a two-stage Background and Object Generative Adversarial Network to resolve this \cite{chen2022background}. The first stage incorporates the background with object layout using a transformer-based seq2seq architecture. In the second stage, a text-attended Layout-aware feature Normalization is applied to transfer objects from the class-to-image model to the layout-to-image model. They have applied the Instance Specific and Layout-Aware feature Normalization(ISLA-Norm) \cite{sun2019image} for layout-to-image generation. ISLA-Norm consists of the following steps i) Label Embedding, ii) Instance-specific Projection iii) Mask Prediction iv) ISLA $\lambda(L)$ and $\lambda(L)$ computation. The authors also applied multi-head attention to encoded information for layout-to-image generation. They used KL-Divergence between the generated image and training image as a loss function. An approach to generate image from text is given in Fig. \ref{fig:text_to_im}.

If any person is asked to draw a picture based on a text prompt, they will start with a top-level design overview of the general picture. Then they will fill in the necessary details according to requirements. Qu et al. followed the same procedure for their LayoutLLM-T2I \cite{qu2023layoutllm}. They started a large language model, for their work it is ChatGPT (gpt-3.5-turbo), to understand the context and layout planning. Then they presented a feedback-based model that selects information adaptively and learns from it. In this state, they used a parameter frozen Stable Diffusion GLIGEN \cite{li2023gligen} model where semantic relation information was injected into its backbone. In their text-to-image generation process, they used a pre-trained CLIP as text encoder and Fourier \cite{tancik2020fourier} mapping as bounding box encoder. They used self-attention to integrate object relationship information. For optimization, they started with Monte Carlo Sampling \cite{shapiro2003monte} and then applied the REINFORCE policy gradient algorithm \cite{williams1992simple} for final guidance.

Hong et al. proposed a hierarchical approach to generate text to image \cite{hong2018inferring}. They did not directly create images from the text, they divided this process into multiple steps. In the first step, they created a bounding box for each object based on the input text prompt. They used an auto-regressive decoder for the box generation where they imposed conditions based on LSTM \cite{hochreiter1997long}. This step is guided by reducing the negative log-likelihood. In the second step, they created a semantic map segmentation of objects inside of each bounding box to generate a more detailed image structure. This shape generator used a convolutional recurrent network \cite{shi2015convolutional}. This model takes an image CNN encoder and then passes through the bidirectional convolutional LSTM. Here, they used two conditional adversaries to enforce global and instance-wise constraints. These two steps already generate size, shape, and class information based on input text. In the final step, only pixel-by-pixel synthesized images according to the mask in pending. They used the conditional encoder-decoder network \cite{isola2017image} for this purpose. They used different types of evaluation for their models, one of the unique ones is caption generation using \cite{vinyals2015show}.

It is hard to generate a scene based on the label set. This label set just contains the name of the object and does not contain any relationship among them. So, this model requires spatial and count relationships to be learned. To manage this situation, Jythi et al. proposed their LayoutVAE \cite{jyothi2019layoutvae}, to guide the scene generation process. This model has two components to solve two problems, i. the count of objects and ii. spatial relationships among the objects.  For maintaining the count of the object, they used countVAE and for the bounding box of each object, BBoxVAE was used. CountVAE is a type of conditional VAE that predicts the count of the first object in the label set, then predicts the count of the second object continues till the end in an autoregressive manner. BBoxVAE generates bounding boxes from left to right in sequential order.

Text2Scene \cite{tan2019text2scene}, proposed by Tan et al., is a new interpretable model based on sequence to sequence framework \cite{sutskever2014sequence} that predicts and adds objects in the scene at each time step, generated from the text prompt. This model consists of the following parts i) Text encoder: maps text input to latent object representation, ii) Image encoder: encodes current scene using convolutional Gated Recurrent Units (GRUs) iii) Convolutional recurrent model: passes current step to next step. iv) Attention model: applied on text input v) Object decoder: predict next object to the scene vi) Attribute decoder: assign attributes to the new object. The Text Encoder is a Bidirectional recurrent network accompanied by Gated Recurrent Units (GRUs). To get spatial contexts for already added objects they used convolutional networks like \cite{xu2015show}. They trained their models for three types of scenes, (1) Cartoon-like scenes from Abstract Scenes dataset \cite{zitnick2013learning}, (2) Object layouts using the COCO dataset \cite{lin2014microsoft}, and (3) Synthetic scenes using the same COCO dataset \cite{lin2014microsoft}. As they are generating objects sequentially, they might lose small objects generated at the beginning.

Grounded-Language-to-Image Generation (GLIGen) \cite{li2023gligen} is a method that adds conditions on the pre-trained model to generate large-scale text to image. They did not update the value of the pre-trained model, just injected the grounding information through a new trainable layer. Here they used bounding boxes after encoding as grounding information and gated Transformer layers \cite{vaswani2017attention} as trainable layers. They applied cross-attention on the encoded caption and used it as a trainable layer.  They used  Latent Diffusion Models (LDM) on LAION \cite{schuhmann2021laion} as their baseline model.  According to the researchers claim that their model significantly outperforms all other models in zero-shot performance on layout2image generation. One of the limitations of this model is, that the generated style might change after adding a new layer.

Zakraoui et al. tried to generate images based on story text \cite{zakraoui2021improving}. They divided this story visualization task into 3 different subtasks i) understanding semantic text ii) prediction of object layout and iii) generation of object. Initially, they generated a graph by taking information from the story. As a story, they used COCO caption story text. They focused on each sentence of the story, gathering key points from the story and converted into a semantic representation. Following the sg2im model \cite{johnson2018image}, they gathered information about objects for graphs through an object layout module. This module consists of a graph convolution neural network. Then they passed this information to an image generation framework. They started this image generation process by generating low-regulation pictures using scene graph framework \cite{reed2016generative}. Finally, they used the object layout module, a vector of 1024 dimensions of text word embedding using Char-CNN-RNN and low regulation picture for high-quality image generation. This framework is conditioned by StackGAN \cite{zhang2017stackgan} and the object layout module. To assess the quality of the image they used K-nearest neighbor score. This approach still has a lot of scope for improvement as this model sometimes generates unrealistic images without understanding the local context.

To generate images from text Ji et al. used a semi-supervised approach \cite{ji2020text}. For a semi-supervised process, all images do not need to be labeled. They treated the unlabeled image as a pseudo-text Text Feature and generated an image using this. It is a combining image feature and text features by maintaining the semantics. As the feature may come from 2 different domains, image or text, the model should maintain the following things i) Modality invariation:  network should not be able to tell from where the feature is coming ii) Semantic consistent: semantic information from both features should reflect the same thing. To achieve this and take advantage of domain adaptation \cite{ganin2015unsupervised} and gradient reversal layer (GRL), they designed a Modality-invariant Semantic-consistent Module. For the baseline model they used \cite{reed2016generative}, for image encoding they used CNN, and for text encoding, they used Recurrent Neural Network (RNN).

To solve problems with complex text prompts Farshad et al. proposed SceneGenie \cite{farshad2023scenegenie}, a novel guidance approach that impacts the sampling process in the diffusion model. SceneGenie is a layout-based approach that incorporates the bounding box and segmentation map for the diffusion model. This bounding box and segmentation map are predicted by Graph Neural Network. For bounding box generation the injected Gaussian noise outside the region of interest (RoI). For segmentation guidance, they took help from the first-stage autoencoder. They took the text as input and split it to generate the graph. Then they collect the embedding using CLIP \cite{radford2021learning} features from each node. Then they feed this information to the Graph Neural network. This encoded data is used to generate the scene using the diffusion model. The Scene Graph to Segmentation is optimized on three objective functions i) box loss ii) mask loss iii) segmentation loss. This approach faces three main problems i) fail to generate high-quality images for complex structures ii) high time consumption for generation iii) this model needs segmentation maps and bounding box information for accurate generation.

\section{Video}
\label{sec:video}
\begin{figure*}[htbp]
\centerline{
\includegraphics[width= \textwidth]{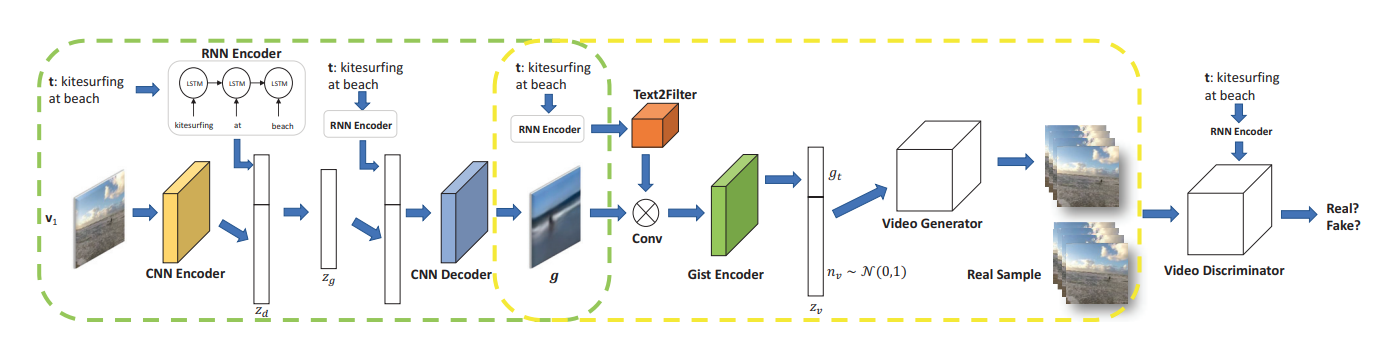}}
\caption{Flow Diagram of A video generation System \protect \cite{li2018video}.}
\label{fig:video}
\end{figure*}

Handling video generation is much more difficult than images. Firstly, a generative model must learn the plausible physical motion models of objects in addition to their appearance models, as a video is a spatiotemporal recording of visual information of objects doing diverse activities. A wrong learned object motion model might result in objects in the output movie making motions that aren't physically conceivable. Secondly, a great deal of variance is introduced by the temporal dimension. Think about how much range of motion an individual can have when doing a squat. Despite the fact that the individual in each movie has an identical appearance, each speed pattern produces a unique video. Third, motion artifacts are especially noticeable as humans have evolved to be sensitive to motion. The approach for video generation is given \ref{fig:video}.

Singer et al. in their Make-A-Video3D (MAV3D) \cite{singer2023text} method tried to generate moving 3D scenes and combine them with video from text description. For this purpose, they used a Text-to-Video diffusion-based model that is optimized for appearance, density, and motion consistency. This is a 4D dynamic Neural Radiance Field. The output of this model can be viewed from any angle or location. It is also possible to compress those in 3D spaces. To overcome Non-Rigid Structure from Motion (NRSfM), they optimize a dynamic Neural Radiance Field (NRF) by generating video with a multi-camera setup. The dynamic NRF faces some problems like i) effective representation of learnable 3d scenes ii) source of supervision and iii) Need memory and compute effective representation in both time and space dimensions. For effective representation, they combine static \cite{sun2022direct} and dynamic \cite{cao2023hexplane} approach to generate NeRFs \cite{mildenhall2021nerf}. To supervise, they proposed a multi-stage training pipeline by optimizing dynamic scenes using Score DIstillation Sampling \cite{poole2022dreamfusion}. Dynamic NeRFs are inefficiently converted for real-time implementation to a series of discontinuous meshes. Additionally, the quality of the depiction has increased with the use of super-resolution data. The T2V model's capacity to produce videos from different points of view determines the quality of the representation of Text-To-4D Dynamic Scene Generation.

Kasten et al. generated long-term multi-scene video from text description with zero shore that includes the scene and camera pose using their SceneScape architecture \cite{fridman2024scenescape}. This model does not need any training data as it utilizes the pretrained diffusion \cite{rombach2022high} and depth prediction model \cite{ranftl2020towards}\cite{ranftl2021vision} vision. The input of this model is the description and the trajectory of the camera described in the text. Following the text prompt this model generates one frame at a time and generates a new frame as the camera moves. The depth prediction model determines the geometry of new content. As two separate frames can be generated with different depths, so the authors fine-tuned the depth of the model at the time of testing. Then they used this newly generated part to update the missing mesh and rendered it in the 2d scene. The challenge with this model is that each frame depends on the previous frame, so little change in each frame can cause huge changes after some time. Another challenge is to portray dramatic depth discontinuities, such as sky vs. ground in outdoor situations because we represent the scene with triangle meth.

To generate photo-realistic, diverse, and coherent motion Bar-Tal et al. proposed Lumiere \cite{bar2024lumiere}, a video generator from text using the diffusion model. Taking inspiration from \cite{cciccek20163d}, they introduced a U-Net of Space-Time architecture (STUNet)  that generates video from at a single go.  Most of the previous models tried to generate keyframes first with the base model and then added internal frames to complete the video using temporal super-resolution (TSR). This process misses consistency between frames. But for Lumiere's case, the research generated overlapping frames using Multidiffusion \cite{bar2023multidiffusion} to maintain consistency using STUNet. So they do not need TSR anymore. This process can generate 80 frames at 16fps. They created this model on top of the pre-trained model \cite{hong2022cogvideo}\cite{singer2022make}\cite{saharia2022photorealistic}. However, this model is not optimized for multiple shots or transitions between scenes.

Vondrick et al. in their research focused on the scenes transforming with time \cite{vondrick2016generating}. They segregated the foreground and background generation process. As a result, the background becomes static and they can only focus on the foreground. This helps the generative adversarial network to understand which object is moving and which object is not. The authors started with some principles like i) the network will not change based on space and time ii) it will generate high dimension images from low-resolution video and iii) the camera is at a static position. They combined spatiotemporal convolutions \cite{ji20123d}\cite{tran2015learning} with fractionally straided convolutions \cite{zeiler2010deconvolutional}\cite{radford2015unsupervised} to generate video. To generate background they applied a small sparsity on the mask. They also modified the discriminator for two purposes i) to differentiate between real scenes and generated scenes and ii) to identify realistic motion in the video.  Following this process they are able to generate videos up to a few seconds.

Ostonov et al. applied the proximal policy optimization (PPO) algorithm for scene generation and indoor planning as a reinforcement learning architecture \cite{ostonov2022rlss}. By applying greedy search they wanted to reduce the action space. Reinforcement learning is the preferred suit where the number of data is not the issue, the issue is the labeling. For their rewarding process, they added conditions on the number of objects generated. If the model fails to meet the condition then -1 is rewarded and if it is able to meet the condition then 1 is rewarded.  PPO is faster than other algorithms so it is used for both policy and value-based updates. This model is based on reinforcement learning and it sequentially generates objects and places in the scene. Due to this nature of the model they called it the Reinforcement Learning algorithm for the Sequential Scene generation (RLSS) model.

To generate video from text, Li et al. used a hybrid architecture including Conditional Variational Autoencoder (CVAE) and Generative Adversarial Network \cite{li2018video}. Their idea is to generate a video with a consistent background with a small motion of the object. They took the input text and applied CVAE to it, this generates the sketch text-conditioned background color and object layout structure, This part is called the ‘gist’ of the video. Then they combined the ‘gist’ and feature vector from the input text to generate the final video. For their GAN architecture, they adopted the idea of \cite{vondrick2016generating}, the feature map from text is generated by using the idea from Isola et al. \cite{isola2017image}. The diversity of the scene is introduced by concatenating isometric Gaussian noise. To collect data with matching text descriptions from YouTube, the authors use the same concept as Ye et al. \cite{ye2015eventnet} and cut the video at the time of scene change using \cite{vondrick2016generating}.

Any video is a composition of content and motion. To generate video Tulyakov et al. proposed a Motion and Content decomposed Generative Adversarial Network (MoCoGAN) \cite{tulyakov2018mocogan}, that generates video frames sequentially. It has two parts, the content part is a mapping from a sequence of random vectors from the content subspace and the motion part is sampled to a sequence of frames from the motion subspace. As content remains the same in the whole space, they used Gaussian distribution to generate this. Motion space is generated by recurrent neural networks. MoCoGAN architecture consists of 4 sub-networks, the recurrent neural network, the image generator, the image discriminator, and the video discriminator. The input of the RNN is generated from a one-layer GRU network \cite{chung2014empirical}.

\section{Image reconstruction}
\label{sec:image reconstruction}
Image reconstruction is another popular use case for image generative models. There might be different reasons to generate the image again from another image. The reasons can be enhancing the images, reducing noise, or improving resolution. This applicable in different fields like medical imaging, satellite imagery, surveillance, photography, art restoration, forensic analysis, and remote sensing.

\subsection{From different angle}
For navigation or path playing, sometimes we need to decide based on visual information from only one side. To fix this kind of issue Novotny et al. proposed PerspectiveNet \cite{novotny2019perspectivenet} which uses geometric properties to generate views from different angles using RGBD view from one side and a viewpoint. They used the differentiable point tracker \cite{tulsiani2018layer} to project the image in Westview. The rest of the test views are filled with training a deep denoising RGBD autoencoder. For image parametrization, they followed \cite{bloesch2018codeslam}, who demonstrated that a deep latent coding of depth images can overcome the need for complex regularizers. They used Reprojection consistency loss that projected back the projected value in a common space and tested the difference. The style loss \cite{gatys2015neural} has been shown to facilitate more realistic results for image generation \cite{chen2017photographic}.

DeVries et al. designed a model to recompose local regions to project on moving cameras from different angles \cite{devries2021unconstrained}. They used a generative model Generative Scene Networks (GSN) to decomposite details and distribution coverage. This model takes a camera pose along with the camera instance parameter as input. Their model is the same as traditional GAN models, but it has two generator and sub-generators for the original generator. One local generator and one global generator. The global generator maps information to a 2D grid of local latent space. Each grid contains only information about that local area. Then they plotted the 2D points to 3D local conditional radiance field \cite{mildenhall2021nerf} using bi-linear sampling \cite{jaderberg2015spatial} and positional encoding. To tackle sampling from an invalid location, the authors stochastic weighted sampling. They generated all candidate camera poses and then applied Softmin on this to reduce the likelihood of sampling invalid locations. Their discriminator has the same architecture as StyleGAN2 \cite{karras2020analyzing} In this work, they also showed that decomposing a latent code into a grid of locally conditioned radiance fields results in an expressive and robust scene representation.

In order to begin, Lu et al. \cite{lu2023wovogen} trained an autoencoder model \cite{van2017neural} to convert a 3D world volume captured in a single frame into a 2D latent representation. The picture distributions in this latent space are modeled by a denoising diffusion probabilistic model (DDPM) \cite{ho2020denoising}. By including temporal versions of its residual and cross-attention blocks, they significantly improve the UNet's ability to handle time-varying data while using the vehicle control sequence as the conditional context. They used the autoencoder's decoder to generate the future 2D temporal latent and then decode it back into the 4D world voxel volume. After generating the future 4D world voxel volume, they transform it into a 4D world feature using a mix of 3D sparse CNN and CLIP. After that, this feature is geometrically modified to represent each time step's associated sample 3D picture volume for each camera. These volumes of 3D images are then compressed into features of 2D images, which are used as conditional inputs for ControlNet \cite{zhang2023adding}. In addition to visual cues, they utilize textual prompts as scene guidance, similar to those found in Stable Diffusion \cite{ho2020denoising}, to control the general scene parameters, including location, weather, lighting, and scene composition. Textual cues are used more specifically to control item position throughout the scene with better accuracy. We use the label names as textual criteria to transfer the 4D world volume labels onto each 2D pixel, providing objective guidance over each pixel's properties. They concatenate surround-view pictures into a meta-image and use the diffusion model to learn the distribution of real-world multiview image sequences in order to achieve inter-sensor consistency. They use the same temporal Transformer blocks that were previously defined in order to guarantee temporal coherence.

\subsection{Image modification}
In another paper, Gu et al. introduced knowledge-based to incorporate commonsense into the model \cite{gu2019scene}. This model extracts useful features from images. They used ConceptNet \cite{speer2013conceptnet} and to exploit multihop reasoning they used Dynamic Memory Network (DMN) \cite{kumar2016ask}. To generate a graph from the image they started extracting a set of objects from images using a Region Proposal Network (RPN) \cite{ren2015faster} and a subgraph is used to reduce computation \cite{li2018factorizable}. Then these features are refined with external knowledge. They introduced a feature refinement method for this. They predict the object label from the object and match it with corresponding entities in knowledge-based (KB). They identified the top-8 related items where the weight of that context is given by KB. They converted these relations as embedding and fed them into an RNN-based encoder \cite{xiong2016dynamic}. After identifying the most reasonable embedding using \cite{xiong2016dynamic} a graph is generated. In the final stage, they used a GAN-based model to generate images from this graph.

To manipulate the semantics of existing images Dhamo et al. provide an approach that shows a way for the user to change the node of the scene graph or its semantic edge to modify the original image \cite{dhamo2020semantic}. The whole system can be divided into three steps. The first step is representing the existing scene in the graph. For this research, they used F-Net \cite{li2018factorizable} which represents the graph as the combination of embedding of category, 4 corners of the bounding box of that object, and a feature encoding of visual representation. The second step is to modify the graph by the used. Users can modify object categories, locations, or relations among objects. They introduced a spatio-semantic scenegraph network (SGN) that helps to generate new training images without manual effort. The final step is to generate a new image from the modified graph. This procedure started by generating a layout from the graph and then image synthesizing using 2 different decoder architectures: cascaded refinement networks (CRN) \cite{chen2017photographic} (similar to \cite{johnson2018image}), as well as SPADE \cite{park2019semantic}.

\section{UI}
\label{sec:ui}
Layout scene design is mainly aligning different components like images, texts, shapes to gain the interest and attention of users. However, the challenging part of layout design is to properly localize the components based on their interdependent relationship \cite{lee2020neural}. Different research works on automatic layout generation are done to find out the relationship to overcome this challenge. Most of the research works implemented Generative Adversarial Networks (GAN) and CNN models for prediction. Many works considered designing a graph using the components as vertices and the relations as the edges and used the VAE model and Graph Convolution Network to predict the location where the components must be placed in the final layout as well as the implicit relations between objects. Sometimes, webpages of online sites are also used as input for feature extraction and transformer as the prediction model to design visual-text layout. For video game content generation, a Genetic algorithm is used to create new content after each level of the game. 

\begin{figure*}[htbp]
\centerline{
\includegraphics[width= \textwidth]{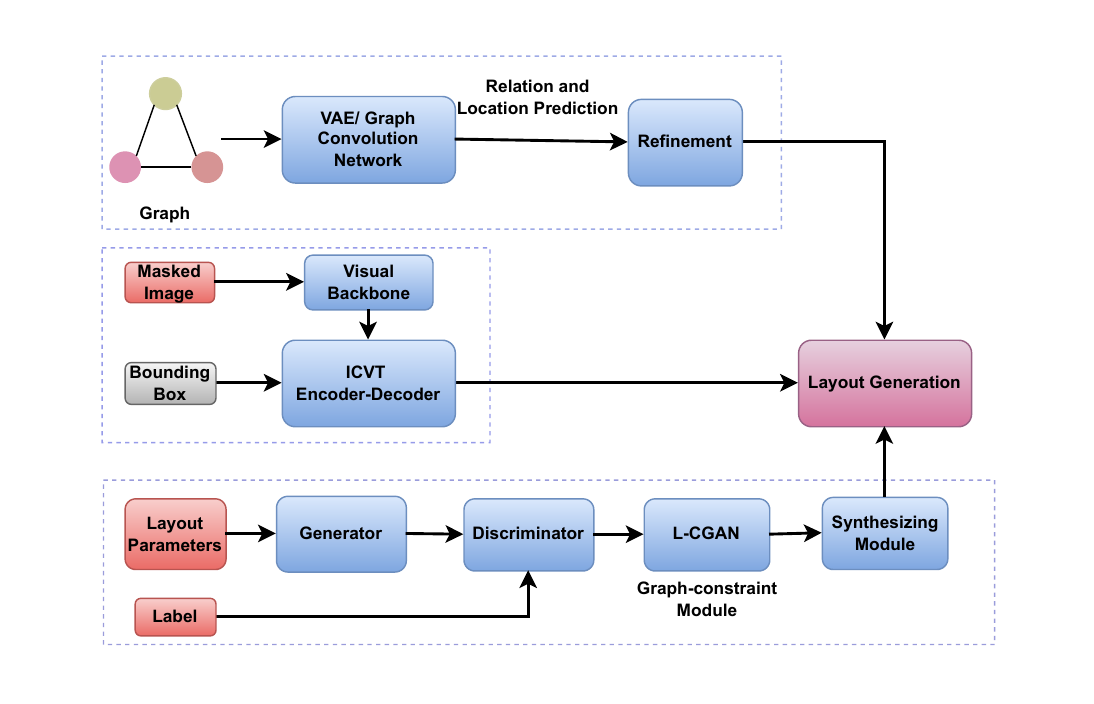}}
\caption{Flow Diagram of A Overall UI Design System.}
\label{UI}
\end{figure*}

Figure \ref{UI} shows the overall layout generation system that combines all implemented methods.

The authors in \cite{shi2022conditional} proposed a layout generation model that implements the Conditional Generative Adversarial Networks (L-CGAN) model. The model allows scaling, positioning, and flipping the primitives which makes the layout more simple and efficient. The authors also proposed a preprocessing algorithm for input data to be unrestricted in size so that they are not fixed. The system is implemented in different stages. Firstly, the generator outputs a mapping function from the prior distribution of the given data and the discriminator outputs a scalar value if the output of the mapping function came from training data or not. The G and D modules are simultaneously trained using the CGAN network. The proposed Graph-constraint module combines the G and D models for optimizing the layout design. Finally, the synthesizing module implements the proposed algorithm for preprocessing the input data and minimizes the gap between the structured input and the 2D image output. The authors evaluated the models in three different types of domains: handwritten layout, scene layout, and document layout. The MNIST Layouts dataset is used for handwritten layouts. They used the baseline code of LayoutGAN to compare their model using the handwritten dataset. The proposed method is a supervised learning method so it gives output with labels but LayoutGAN gives results randomly. For scene generation layout, they trained AbstractScene-Layouts which is an extended version of Abstract Scene Dataset v1.1 \cite{zitnick2013bringing}. Both the used study and ablation study are conducted for evaluating the scene layout generation using the proposed system. Finally, for documental layout, PubLayNet \cite{zhong2019publaynet} is used and they did qualitative and quantitative comparison of their method with other baseline methods based on coverage and overlapping. The proposed method performs best among other baseline methods.

Petrovas et al. \cite{petrovas2022procedural} tried to develop an autonomous Procedural Content Generation method based on Machine Learning (PCGML) for video game content generation. They used a genetic algorithm for iterative scene generation for different levels and objects of the game and the weighted aggregated sum product assessment (WASPAS) algorithm in a single-valued neutrosophic set environment (SVNS) as the fitness function. The development of the system is divided into 3 stages. The first stage is Game Scene Encoding Modeling where a set of object types is selected based on game-level design principles and they are encoded in a matrix. The numbers in the matrix grid represent different object types and each object can take one cell. The single-scene layout creates a single genetic algorithm chromosome. The second stage is the Game Scene Procedural Generation Criteria List. Here, a set of criteria is chosen based on recurrence in the literature, game principle designs, and creative definitions. The normalization of the criteria is in the range of [0,1] using fuzzy logic. Different types of fitness functions are used for calculation of fitness of different creative key terms like aesthetics, usefulness, etc. Finally, in the evaluation stage, the fitness functions are combined and used in the WASPAS-SVNS fitness function of the genetic algorithm at each iteration. In each iteration of the genetic algorithm, fitness for each game-level grid is calculated and the best-performing grids are selected. The model generates a variety of unexpected scenes which are different from the training data. The authors developed a framework from scratch for the research using the Unity game engine and visual game object assets from the Unity Asset Store.

Lee et al. \cite{lee2020neural} developed a graphical layout design system using neural networks. The system is divided into three modules. Firstly in the relation prediction module, the system takes design components and user-defined constraints as inputs and generates a relational graph. The nodes in the graph are the components and the constraints are the edges. Then the CNN model generates a complete graph between nodes having unknown relation edges. In the layout generation stage, the complete graph is used to create the layout of components by predicting the bounding box of all nodes. This stage is completed by implementing the graph-based iterative conditional VAE model. Finally, in the layout refinement stage, the authors fine-tuned the bounding boxes to make them optimal and better aligned with better visual quality. For evaluating their work, the authors used three datasets: Magazine \cite{zheng2019content}, RICO \cite{deka2017rico}, and Image banner ads by searching using keywords. They compared their method with sg2im and LayoutVAE algorithms qualitatively and quantitatively. In both ways, the proposed method gives competitive outputs.

In another paper, Cao et al. proposed an Image-Conditioned Variational Transformer (ICVT) model for generating a visual-textual layout in an image \cite{cao2022geometry}. The proposed model has three components. Firstly, visual features like categories and geometric parameters of the input image are extracted using Vision Transformer (ViT) in the visual backbone stage. The features of each element are projected into the embedded vector so that the whole layout can be transformed into a sequence of vectors which makes it a conditional sequence autoregressive prediction problem. Then in the ICVT encoder stage, a transformer CVAE encoder is implemented with a self-attention layer, cross-attention layer, and feed-forward network (FFN) to find the relationship between layouts and input image. ICVT encoder mainly models the posterior distribution in the CVAE framework. Finally, the ICVT decoder, having the same architecture as the encoder, is performed as an autoregressive generator which allows it to generate layouts by latent space sampling. The authors also proposed a geometry alignment module that extracts geometric information of the input image and aligns it with the layout elements. For training and testing the models, the authors build a dataset of 117,624 visual-textual advertisement poster images. They compared their method with other methods like GAN, Layout Transformer \cite{gupta2021layouttransformer}, and VTN \cite{arroyo2021variational} by reproducing them and training them with the dataset. The proposed ICVT method performs better based on alignment, occlusion rate, and FID metric.

Yang et al. \cite{yang2016automatic} proposed an automatic visual-textual layout generation system that is topic dependent. The authors firstly proposed a set of templates with specific topics having aesthetic principles based on domain experts and secondly designed a computational framework for generating final layouts with key elements. For template design and aesthetic principles, two aspects are considered for template designing: spatial layout and topic-dependent style after taking interviews with seven domain experts. The authors collected the 8 most used topics for template design. The framework consists of four modules. Firstly, the system uses an algorithm proposed by \cite{yin2013automatic} to extract the features of the webpages given as input to find out the domain and the key sentences. The user can also insert images and texts that can be used as input for the framework. Secondly, the image is analyzed to create a visual perception map having the key regions and  the image is resized to match with the size of the target layout preserving the key elements according to the visual perception map. Thirdly, the texts are overlaid into the image by using the spatial layout by an energy optimization process. Finally, the text color is selected by analyzing the color palette of the image, topic attribute, semantic attributes, global color harmonization, and local readability. The text coloring is implemented by applying different types of hue/tone models based on the topic of the input image. The authors created a dataset having 104 news on 8 topics from three different sources: CNN, Bing News, and Google News. The authors used rating scores of users for comparing their model with other methods: MM12 \cite{kuhna2012semi}, IUI13 \cite{jahanian2013recommendation}, and by designers. The proposed method has better practical use than other methods based on mean score and standard variance under different criteria.

\section{Image to 3D}
\label{sec:image to 3D}
\begin{figure*}[htbp]
\centerline{
\includegraphics[width= \textwidth]{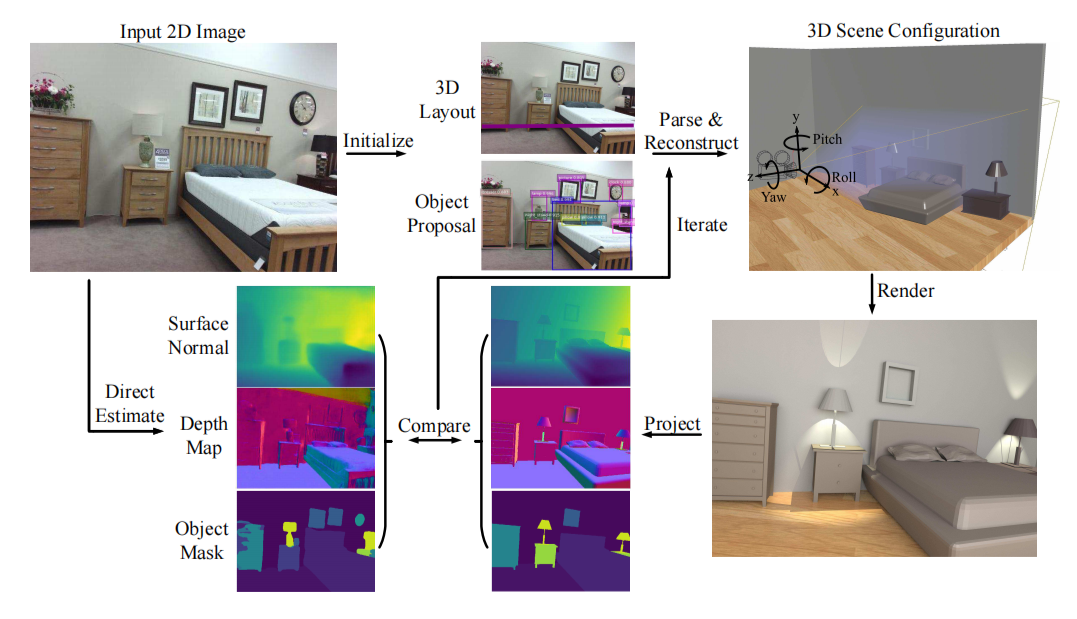}}
\caption{Flow Diagram of A Image to 3D Scene Generation System \protect \cite{huang2018holistic}.}
\label{fig:image_to_3D}
\end{figure*}

Generating a 3D scene from 2D images is a very common technique in the field of scene generation. However, extracting features from images and computing the depth value for the 3D scene is not an easy task. Many authors use different techniques for identifying the objects from the images and measuring the depth value. The most used models for this application are CNN, GAN, and VGG models. Some authors also implemented the ResNet.-18 model in both 2D instance segmentation and depth estimation. Most of the research works mainly use 2D images to find objects and their classes and searches relevant 3D scenes from the available dataset. Some systems use cameras to capture real-time images from different angles and preprocess the images using different versions of the SIFT algorithm for creating the final 3D scene. LiDar-based 3D object detecting method hybrid 2D semantic scene generation is also implemented. Figure \ref{fig:image_to_3D} shows an image to 3D scene generation system.

The authors in \cite{lienard2016embedded} proposed an inexpensive UVA control-based system for 3D scene construction during a flight with a duration of 20 minutes. The system is designed in two steps. One is collecting images from two cameras. A web camera is used for capturing images with low resolution and a high-resolution GoPro camera is triggered with a serial communication. The system is developed on Intel’s Minnowboard Max single board cameration that is combined with a custom-made baseboard and Computer Model implemented in \cite{meier2012pixhawk}. The image processing is implemented using different open-source implementations of the SIFT algorithm like Ezshift, OpenCV, and SIFT\_PyOCL \cite{lowe1999object}. The flight is controlled by two methods: Guided and Manual. In the guided method, the onboard computer continuously communicates with the flight controller to collect images from different angles and backtracking. The manual mode is controlled by humans and is only used for takeoff, landing and any emergency situation. The complete system is evaluated by using a virtual environment and an active real-time environment. The result shows that SIFT\_PyOCL performs better among the used SIFT algorithms. But it takes around 8s for one high-resolution image. This high workload problem can be solved by cropping the image or reducing the resolution of the image.

Abdul et al. implemented a system called SceneIBR which takes 2D images as input and searches for relevant 3D scenes from a dataset \cite{abdul2018shrec}. The input 2D image dataset contains 10,000 2D images collected from ImageNet and the output 3D images are collected from Google 3D Warehouse. Both dataset has ten classes containing 1000 2D images and 100 3D scenes per class. Then the dataset is distributed to six groups for the application of retrieval methods.  Due to different challenges, three groups completed the given task. The first group proposed a model called Mean Discrepancy domain adaption based on the VGG model (MMD-VGG) that has two steps: Data Preprocessing and Feature Extraction. They used SketchUp for the preprocessing of the 3D image dataset. For feature selection, they used both learning and non-learning-based settings. They implemented the VGG \cite{li2014comparison} model pre-trained on the Places \cite{li2017sketch} as the initial parameters in the deep network. They used one fully connected layer as the output layer of the image. They used Principal Component Analysis for domain shift reduction and measured the similarity of query and target image using Euclidean Distance. They used the VGG model pre-trained on the Places for direct feature extraction in the non-learning setting. Another group implemented a deep neural network that combines two CNN networks. They implemented triplet center loss (TCL) and softmax loss for learning a discriminative and robust representation. They also used the Euclidean Distance for measuring distances. The last group implemented a system called the RNIRAP algorithm. Here, they used the ResNet18 model in their fully connected neural network with one hidden layer. They also worked on improving the accuracy of classifying the label of the 2D image by implementing five classification models having the same structure and combined their results using a voting scheme. The five models are Saliency-Based 2D Views Selection, Place Classification Adaptation for 3D Models, Rank List Regeneration, and BoW: Bag-of-Words Framework Based Retrieval (BOW1 \& BOW2). Finally, they evaluated the proposed settings of three groups using seven parameters. They are Precision-Recall (PR) diagrams, Nearest Neighbor (NN), First Tier (FT), Second Tier (ST), E-Measures (E), Discounted Cumulated Gain (DCG) and Average Precision (AP). Based on the parameter values, RNIRAP performs best followed by the MMD-VGG model and TCL model.

Abdul et al. developed a system called SceneIBR2019 which is mainly an updated version of SceneIBR2018 \cite{abdul2019shrec}. They updated the dataset with 30,000 2D images and 3,000 3D scenes. Also, the classes are extended to 30 classes having 1,000 2D images and 100 3D scenes per class. Then 700 images and 70 scenes from each class are randomly selected for training purposes and the rest are for testing purposes. Six groups from different countries registered for the task but three groups completed the task within a deadline of 1 month. The first group used the RNIRAP algorithm for 3d scene retrieval which was also implemented in the previous version. The second group implemented CVAE and CVAE-VGG methods for the task. CVAE is mainly a semi-supervised method for classifying the image using neural networks. They used the sum of the negative Evidence Lower Bound (ELBO) and a classification loss as the loss function. Finally, the last group implemented VMV-VGG which utilizes VGG-16 architecture.  They mainly created 13 scene views from each 3D scene and 12 are selected uniformly. Then they augmented the dataset to avoid overfitting and extending the data. They implemented the VGG2 algorithm with domain adaptation for both 2D and 3D dataset and used majority voting for selecting the final class. The three methods are evaluated using seven parameters which were also used for the previous version of the system as well. Here the RNIRAP method performs best followed by the VMV-VGG and CVAE methods.

In another paper \cite{murez2020atlas}, Murez et al. proposed a method that constructs a 3D scene by taking a set of posed RGB images as inputs. Firstly, the method randomly takes a set of RGB images with intrinsics and pose information of each image. Secondly, the 2D CNN model is used to extract features from the images and the features are back-projected into a 3D canonical voxel volume. Thirdly, the back-projected features are accumulated using a running average. Finally, the features are sent to a 3D Convolutional Encoder-Decoder Network model as input so that it can be regressed to directly produce TSRF values at each pixel to reconstruct the final 3D scene. The authors also tried to do experiments for the prediction of semantic segmentation of the scene. They trained and tested their model using ScanNet \cite{dai2017scannet} having 2.5M images across 707 distinct spaces. For performance evaluation, they compared their results with four other existing methods: COLMAP \cite{schonberger2016structure}, MVDepthNet \cite{wang2018mvdepthnet}, GPMVS \cite{hou2019multi}, and DPSNet \cite{im2019dpsnet}. The proposed model is competitive with other models and sometimes outperforms the existing ones. As other methods take depth as input which makes the 3D construction much easier, they sometimes give better results. However, the proposed method is much faster than COLMAP and DPSNet.

Dahnert et al. \cite{dahnert2021panoptic} proposed a Panoptic 3D Scene construction method that combines the process of geometric reconstruction, 3D semantic segmentation, and 3D instance segmentation. The method takes one RGB image as input and uses ResNet-18 to predict both 2D instance segmentation and depth estimation. Mask R-CNN is implemented for prediction of 2D object bounding boxes, class category labels, and instance masks. A depth estimation encoder is used for depth estimation. By using camera intrinsics and depth estimations, the 2D image is lifted to 3D estimates. Then the 3D estimates are encoded using a sparse generative approach which produces a refined and high-resolution 3D scene output. For training and evaluating the proposed method, both synthetic and real-world dataset are used. 3D-Front is used for synthetic dataset \cite{fu20213d} and Matterport3D is used for real world dataset \cite{chang2017matterport3d}. They evaluated their method using panoptic reconstruction quality (PRQ) metrics as an average measurement among the class categories. The proposed method is compared quantitatively and qualitatively with existing models like SSCNet, Mesh R-CNN \cite{gkioxari2019mesh} and Total3D \cite{nie2020total3dunderstanding}. The proposed method significantly performs better than the existing methods. 

The authors in \cite{yin2021learning} developed a framework for monocular 3D shape estimation using a single image. The framework consists of two modules. One is a depth prediction module (DPM) and the other is a point cloud module (PCM). The system is implemented by training the modules separately by different data sources and finally combined in the inference time. The DPM takes a single RGB image as input and trains CNN to predict a depth map with an unknown scale value and shift. A 3D point cloud is predicted using the initial depth map. The PCM takes the 3D point cloud and initial estimated focal length as inputs and develops the PVCNN model \cite{liu2019point} to adjust the shift in the depth map and focal length for better geometric reconstruction of the 3D shapes. The system uses 9 different datasets to use for the input of DPM. As the data are collected from multiple sources, the paper proposes two types of losses: Image-level normalized regression (ILNR) loss for unknown depth scale and shift, and Pair-wise normal (PWN) loss for improving geometric features. The authors also compared their work with other existing state-of-art methods and the proposed system gives better performance.

Popov et al. \cite{popov2020corenet} proposed a simple 3D object reconstruction method that takes a single RGB image as input. They used a simple neural network model to reconstruct the full volume of the acne and extract the object meshes. They extended the model with three more features. Firstly, they implemented Ray-traced skip connections that work to propagate 3D information to 3D volume and help to reconstruct sharp details. Secondly, a hybrid 3D volume representation is created to build translation equivariant 3D models using standard convolutional blocks and encode fine object details. Thirdly, they proposed a reconstruction training loss using an intersection-over-union metric (IoU) that captures the overall geometry of the objects. The authors also worked on generating multiple objects available in the 2D image. Firstly, they tried to predict the semantic classes of the objects. Then they worked to detect occlusion and resolve them using hallucination. Finally, they ensured that every point in 3D space has only one object. They achieved this work by predicting the probability distribution of semantic classes in 3D space. The authors evaluated the system by using two types of data: synthetic data from ShapeNet and real images from Pix3D. They compared their model with other state-of-art methods like 3D-R2N2 \cite{choy20163d}, Pix2Mesh \cite{wang2018pixel2mesh}, and again ONN \cite{mescheder2019occupancy}, Pix2Vox \cite{xie2019pix2vox}, Pix2Vox++ \cite{xie2020pix2vox++}. The proposed method outperforms the existing works.

Chen et al. developed a system called SceneDreamer \cite{chen2023scenedreamer} that can generate unbounded 3D scenes from a collection of in-the-wild 2D images without declaring the camera positions. The framework is developed in three modules. Firstly, they proposed a Bird’s-Eye-View (BEV) Scene Representation. For learning the 3D GAN model, 3D GAN generates a height field and a semantic field that represents a 3D scene from BEV representation. Secondly, a generative hash grid which is a hash based encoding function, is introduced to parameterize the space-varied and scene-varied latent features of the scene. Finally, a style-based volumetric renderer implemented using GAN is learned to generate a photorealistic scene by using the latent features from the hash grid. The authors also created a dataset having 1,135,662 natural images by searching using the keywords on the internet. The proposed system is compared with other state-of-art works like GANcraft \cite{hao2021gancraft}, EG3D \cite{chan2022efficient}, Inf-Zero \cite{li2022infinitenature} and Inf-Nature \cite{liu2021infinite} in both quantitative and qualitative ways using different evaluation metrics. The proposed system gives comparatively good results.

The paper described an SSGNet system which is a LiDAR-based 3D object detection method using hybrid 2D semantic scene generation \cite{yang2023lidar}. The system has three modules. Firstly, the BEV features are extracted from the initial scene using a 3D backbone network. Any 3D backbone network can be used in this stage. The authors experimented with different networks like CenterPoint (sparse 3D Conv) \cite{yin2021center}, PillarNet (sparse 2D Conv) \cite{shi2022pillarnet}, VoxSeT (transformer) \cite{he2022voxel}, and PVRCNN \cite{shi2020pv}. The second stage is the BEV refinement module where the hybrid 2D semantic scene generation method is implemented to refine the initial BEV features that improve the performance of the 3D object detection in the final module. The authors implemented both implicit and explicit networks to produce new refined features. Finally, any LiDAR-based detection head can be used to detect the 3D objects. For evaluation, the authors used a version of Waymo Open dataset \cite{sun2020scalability} having 158,081 training samples and 39,987 validation samples and nuScenes dataset \cite{caesar2020nuscenes}. The proposed hybrid 2D semantic scene generation method is infused into four most popular LiDAR-based detection heads \cite{yin2021center}\cite{shi2022pillarnet}\cite{he2022voxel} \cite{shi2020pv} and compared with IA-SSD \cite{zhang2022not} and other methods \cite{yan2018second}\cite{lang2019pointpillars}. The proposed SSGNet model gives comparatively better performance based on mAP value. 

The authors proposed a data-driven approach for recovering a complete 3D model and its objects from an RGBD image \cite{zou2019complete}. RGBD image is mainly the RGB image having depth value. The authors first try to predict the support heights and classes of the objects using support CNN and classification CNN. Then similar objects are identified by using Shape retrieval CNNs so that the most similar region can be matched from the training set based on the object shape. Finally, the subset of the proposed objects is aligned using the optimization method to present the overall scene according to the occlusion, image appearance, depth, and layout consistency for evaluating the method. They used the code from \cite{tighe2013superparsing} to find the major orthogonal scene orientations for aligning the scene and obtaining the height value for each pixel. The authors also contributed to refine the NYUd v2 dataset \cite{silberman2012indoor} with detailed 3D shape annotations for the objects in every image of the dataset \cite{guo2013support}. The authors compared their method with previous work of theirs \cite{guo2015predicting}, Deep Sliding Shapes (DSS) \cite{song2016deep} and Semantic Scene Completion (SSCNet) \cite{song2017semantic}.

In the paper \cite{huang2018holistic}, the authors proposed an algorithm for holistic 3D scene parsing and reconstruction of indoor scenes. The system takes a single RGB image as input and simultaneously reconstructs the functional hierarchy and the 3D geometric structure of the scene from the image. The authors also proposed a Holistic Scene Grammar (HSG) for representing scene structure. HSG takes three features: latent human context, geometric constraints, and physical constraints. The scene is represented as a parsing graph where the nodes contain a Markov random field (MRF) and a hierarchical structure. MRF mainly captures the contextual relations between objects and room layout. The system uses a maximum posterior probability (MAP) estimate to find the optimal solution of the observed image simultaneously. The authors used SUN RGB-D dataset \cite{song2015sun} for evaluating the system’s performance qualitatively and quantitatively. The system performs really well in both ways. 

\section{Text to 3D}
\label{sec:text to 3D}
\begin{figure*}[htbp]
\centerline{
\includegraphics[width= \textwidth]{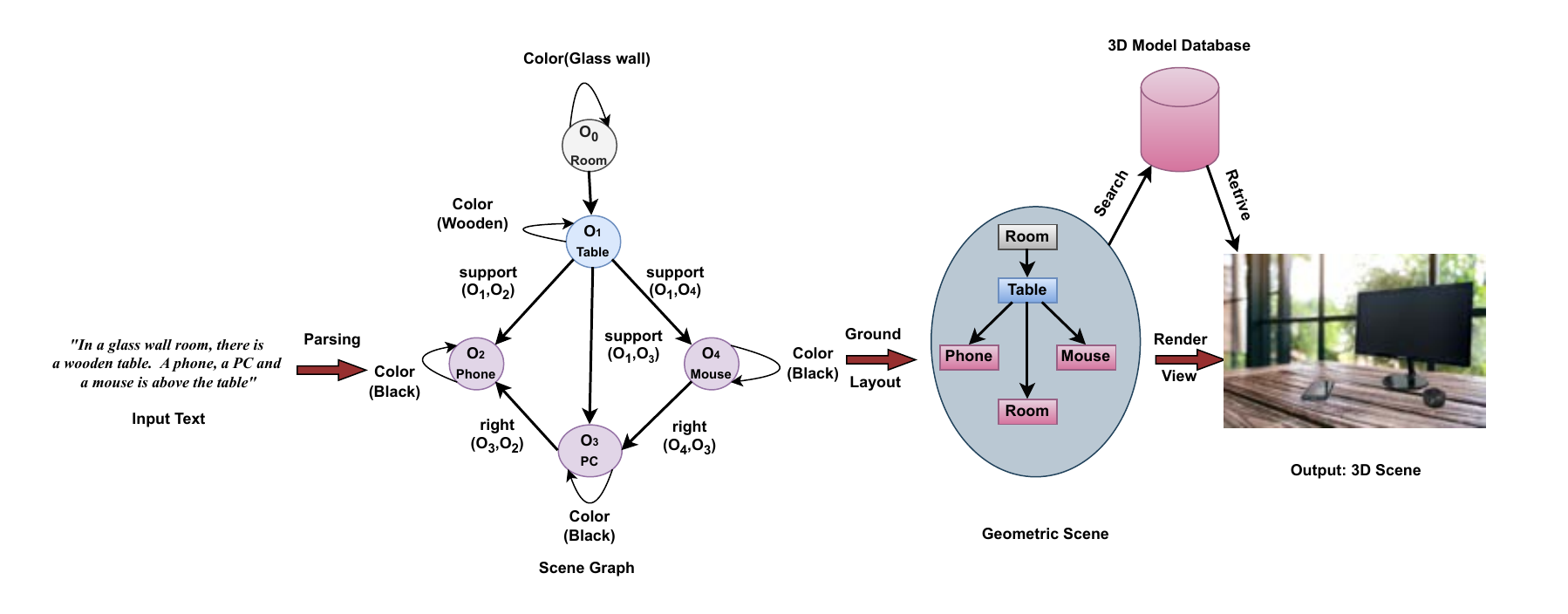}}
\caption{Flow Diagram of A Text to 3D Scene Generation System.}
\label{fig:text_to_3D}
\end{figure*}

Text to 3d scene generation is one of the most popular research technique in the field of scene generation. The system mainly takes text descriptors as input from the user and generate a 3D scene based on the text. The biggest challenge of text to 3D scene generation is that the input text does not contain the implicit relations or constraints of the objects needed for generating the scene. Many authors try to solve this problem in different ways. Some authors use a spatial knowledge representation database to store the spatial priors or relations among the objects, whereas others follow some constructive learning strategy to predict the 3D point clouds needed for the scene. Again, lexical knowledge representation is also used to find the semantic relations between objects. Sometimes, NeRF model and diffusion models are combined to create the framework for creating 360-degree scenes. Also, the SDS method is used to create 3D scenes by using 3D bounding boxes. Some systems also allow modification of the created scene by taking commands from the users. Figure \ref{fig:text_to_3D} shows a text-to-3D scene generation system called Text2NeRF system.

Chang et al. proposed a 3D scene generation system that uses natural language descriptions as input to produce the 3D scene \cite{chang2015text}. The authors created a new dataset with 1128 scenes and 4284 free-form natural language descriptions of them. For the training dataset, they provided a simple descriptor to participants to create a scene using the descriptors and asked other participants to describe those created scenes to get a more varied description of each scene. Both lexical learning grounding and rule based methods were used for developing the system. Firstly, one-versus-all logistic regression was used to identify the appropriate category or 3D model ID from the 3D model database. Then, a rule-based method was implemented for positioning and finding relations between the selected objects. For human evaluation of the final system, the Likert scale was used for giving rating to the scenes created by the system. The authors also introduced another evaluation metric called aligned scene template similarity (ASTS) which strongly correlated with human evaluation. The result shows that a combination of both learned model and rule based model gives better scene output than the methods separately.

The paper describes a cross-platform application for 3D scene generation \cite{seversky2006real}. The algorithm for designing the system describes the placement of objects based on spatial relationships. The algorithm has three stages. In the object mesh voxelization stage, polygon mesh representations of objects are converted into voxelized representations which consist of many cubes representing the shape of the main objects. The second stage is surface extraction \& spatial partitioning which draw out the object’s surfaces and partition them into spatially coherent regions. Finally, in the placement determination voxel collision between two objects is detected and placement of objects in the world coordinate system is done upon satisfying different conditions. The developed system interface has three sections. The main viewing window shows the current scene and is updated by adding new objects. In the right corner is the object viewer where the available objects are displayed. The button corner is for handling a single camera. The input section of the interface has two modes. One mode is for writing the scene description in text format and another mode is for giving audio descriptions of the scene.

The authors in \cite{chang2014semantic} proposed a 3D scene generation application that can generate scenes using texts and provides generated scene modification by the users using different commands. The system is developed using a semantic parsing mechanism. Firstly, the text is parsed to use as input for the scene template which provides different information like what objects are needed for the scene and relationships among the objects. To detect objects from a 3D model database, the system looks for noun phrases and uses different headwords. The objects are filtered with WordNet to find out which objects need to be visualized \cite{miller1995wordnet}.  The information is then used in the spatial knowledge base database for finding the implicit relations that are not mentioned in the input descriptions. Based on the relations provided, a 3D scene is generated in the application. The application also provides facilities for scene modification by using semantic parsing of the scene interaction commands given by the users like select, move, replace, insert, etc.

In another paper \cite{nasiri2022prompt}, Nasiri et al. introduced a simple and effective dynamic 3D scene generation method that can be used for 3D ZSL learning problems. As most of the previous works are done on pre-trained supervised models on 2D images, there is no system for unseen and unlabeled data on the 3D domain for ZSL learning. To solve this problem, the authors generated more 3D cloud samples for training the system. The paper proposed a prompt-based method that uses text captions for annotation. They used a constructive learning strategy for 3D scene annotation. The system takes text captions collected from pre-determined language prompts \cite{tsimpoukelli2021multimodal} and predicts which 3D point clouds are needed for the scene. The system also considers context and relations between objects which makes the system more generalized for unseen point cloud classes. For evaluating the system, the author used both synthetic (ModelNet40, ModelNet10) \cite{wu20153d} and real-world scene (ScanObjectNN) datasets \cite{uy2019revisiting}. The authors compared their results with other existing 2D image and 3D point cloud methods using both types of datasets. The method gives better accuracy than other methods in the synthetic dataset. However it gives a little less accuracy than some methods in the real-world dataset.

Zhang et al.proposed a text-driven photorealistic 3D scene generation framework that can also generate 360-degree scenes \cite{zhang2024text2nerf}. The framework is developed using the NeRF model combined with the pre-trained diffusion model \cite{mildenhall2021nerf}. Firstly, they used a diffusion model for 2D scene generation. They set the camera in 8  different positions to generate different support sets for the initial scene using the depth image-based rendering (DIBR) method and then they train the NeRF model with the support sets for initial 3D scene generation. Finally, they used the PIU strategy to expand the generated view and complete its missing regions using the diffusion model to create new support sets. These support sets are then used to update the NeRF model and this process goes iteratively. In practice, they created 30 inpainting results after each iteration and selected one of them as the final scene that is most similar to the initial CLIP space. The authors also proposed a two-stage depth alignment strategy to align depth values of different positions from different views so that the system can ensure depth consistency. The authors evaluated the system with 8 other baseline methods (CLIP-Mesh \cite{mohammad2022clip}, SJC \cite{wang2023score}, DreamFusion \cite{poole2022dreamfusion}, and DreamFusion-Scene, 3DP \cite{shih20203d}, PixelSynth \cite{rockwell2021pixelsynth}, and Text2Room \cite{hollein2023text2room}) and used three evaluation scores for comparison. The proposed model showed a higher CLIP value and lower BRISQUE, NIQE score value which indicates generating higher quality scenes than the baseline methods.

The authors in \cite{coyne2010frame} are deriving a system called Scenario-Based Lexical Knowledge Resource (SBLR) which is a new version of lexical knowledge representation. The work is mainly adding new features of two existing systems named WordNet and FrameNet into this system. The developing system has 15,000 nouns representing 3D objects, 2,000 3D objects, and 10,000 images. Firstly the authors tried to observe and find out the limitations of WordNet \cite{coyne2001wordseye} and FrameNet \cite{baker1998berkeley}. They worked on finding semantic relations between items via SBLR semantic relation. They tried to extend FrameNet’s notion of valence pattern to add different semantic and contextual constraints. The vignette mapping method was used to map semantic representations into 3D scenes. They are also working on graphical object properties and their relationship with other objects. The system described in the paper is mainly a work in progress and the software will be completed very soon. They are planning to evaluate the software’s performance with the partnership of a non-profit after-school program which is located in New York City.

Po et al. \cite{po2023compositional} proposed a 3D scene generation system that takes text prompts and 3D bounding boxes as inputs and gives a 3D scene as output. The authors proposed a new method called the locally conditioned diffusion model to generate 2D images from the input text. The method also helps to detect noise in an image and by using different denoising steps, the noises are removed from the final images. Then they use the SDS method which is an updated form of the SJC \cite{wang2023score} model to generate 3D scenes using 3D bounding boxes. The SDS method ensures images collected from all camera poses are merged in an image generated using text-conditioned diffusion prior. Also instead of using one text prompt, the system uses denoising steps for all text prompts for any camera pose. The authors used pre-trained models named GLIDE \cite{nichol2021glide} and StableDiffusion \cite{rombach2022high} to apply their proposed locally conditioned diffusion model and use a Voxel NeRF \cite{liu2020neural} to represent the 3D scene model. The proposed system’s outcome is compared with different previous SJC methods. The previous SJC-based system could not capture some certain scenes while merging multiple components but the proposed system has this ability. Again, SJC methods also fail to rebuild different geometric figures with details and the proposed system solves this problem.  The authors also used an attribute value called CLIP R-Precision to compare the system with other single-text prompt SJC models. In each case, the proposed system gives better results.

The authors in \cite{chang2014learning} proposed a text-to-3D scene generation system with the ability to learn implicit constraints. They also proposed a spatial knowledge representation system to find the implicit constraints that are not mentioned in the input text. The task is divided into four stages. Firstly, 3d objects and a set of explicit constraints are extracted by parsing the input text in the template parsing stage. Secondly in the inference stage, the set of constraints is expanded by extracting the implicit constraints that are not mentioned in the text using the spatial knowledge representation of learned spatial priors. Thirdly, based on the constraints list, the set of objects needed for creating the scene is selected from the 3D model dataset in the grounding stage. Finally, the objects are arranged and placed based on the constraints set in the scene layout stage. The authors used the Google 3D Warehouse dataset for 3D models \cite{fisher2012example}. They also compared their system’s performance with other simpler baselines qualitatively and their system performs better.

\section{Others}
\label{sec:others}
Other than the previously mentioned systems, there are some other approach taken for scene generation. Those approaches are rarely used or few researchers actually focused on them. As a result, very few research papers are published with this. So, we have limited information about those. Some examples of these are generating scenes from audio \cite{sung2023sound} or Embedder AI \cite{zhao2021luminous}. These researches are unique and can be a new direction of research in the future.

Sung-Bin et al. proposed a method called Sound2Scene \cite{sung2023sound} to generate a scene from latent space based on input audio. However, the challenges they faced were i) the audio does not contain all the descriptions required for the image ii) The correlation is incongruent. They started with prestrained conditional generative adversarial network \cite{casanova2021instance} with the image encoder, and then they tried to generate the encoder from the audio input. As an image encoder they used ResNet-50 \cite{he2016deep}, as an image generator they used BigGAN \cite{brock2018large} and Audio encoder they used ResNet-18 To consider different samples for contrastive learning, they used InfoNCE \cite{oord2018representation}. For their training, they used unlabeled videos. This approach faces two main downsides i) generates a single/blended object when audio of closely related sound is given ii) the quality of human-related images is not that good. Though the second issue is common for most of the GAN models \cite{li2021collaging}.

The main challenges for most of the existing models for scene generation are robustness, quality, and realism. To solve this, Zheng et al. tried to propose a text-to-4D dynamic 3D scene using their Dreadm-in-4D model \cite{zheng2023unified}. This uses 2D and 3D diffusion guidance to learn high-quality features. They divided their approach into two stages. The first one is for high-quality static scene representation and the second one is the dynamic stage for learning motion. For the first stage, they used 3D-aware \cite{liu2023zero}\cite{shi2023mvdream} and standard image diffusion models \cite{rombach2022high}. For the second stage, they used video diffusion guidance. For 4D content generation, they are a type of deformable neural radiance field (D-NeRF) \cite{pumarola2021d}. To generate 3D from text, they used MVDream through SDS loss \cite{poole2022dreamfusion} guidance. For the dynamic stage, they propose a full disentangling of the static model and the motion by freezing the NeRF network learned in the static stage and only learning the deformation field to match the motion described in the text prompt.

Lee et al. proposed a Multi-scale Contrastive Discriminator (MsConD) \cite{lee2023multi} for pixel-level multi-scale local discriminator for the GAN-based model to generate more accurate data.  This model is based on the ResNet-based discriminator of StyleGAN2 \cite{karras2020analyzing} and uses augmentation, the same as StyleGAN2ADA \cite{karras2020training}. In this research, they worked on improving and enforcing an auxiliary self-supervision task for the discriminator. To process local patches of different scales, they designed a multilevel discriminator. Each level consists of three things: residual block, classification head, and projection head. This model has a backbone network in discriminator to convert input images in a multi-scale feature map. An auxiliary task was assigned to improve the region-level representation. To calculate pixel level difference between the real and generated images at the local level, they wrap the pixel value in vector and measure the Euclidean distance between them.

Indoor scene synthesis for Embodied AI (EAI) mainly faces the following problems: i) the Object must be placed based on household tasks ii) the randomized layout must be normalized; iii) generated data should be accessible. LUMINOUS \cite{zhao2021luminous} proposed by Zhao et al. uses Constrained Stochastic Scene Generation (CSSG) to generate an arbitrary number of new indoor scenes by solving these problems. Then they present the Challenge Definition Format (CDF), which offers an easy-to-use task definition that includes the necessary objects, their spatial connections, and high-level descriptions of activities that downstream EAI will facilitate. Along with CSSG, LUMINOUS uses 3D-SLN \cite{luo2020end} and Deep-synth \cite{wang2018deep}, for scene generation. This LUMINOUS executor consists of three parts: i) EAI task executer, ii) task instruction generator module, and iii) video rendering toolkit. This novel approach is able to generate huge multimodal datasets in high dimensions for advanced machine/deep learning models.

\section{Loss Functions}
\label{sec:loss functions}
The loss function serves to quantify the disparity between predicted and actual outcomes. It's selection depends on the particular task at hand. For instance, in different machine learning tasks, various loss functions are employed, ranging from KL divergence \cite{zhao2019image} to image reconstruction loss \cite{zhao2019image}, RGB loss \cite{zhang2024text2nerf}, perceptual similarity loss \cite{mittal2019interactive}, and others. Each loss function caters to the specific requirements and nuances of the task, enabling efficient optimization and model performance. The Loss Function Table \ref{Loss Function Table} offers a comprehensive overview of various loss functions, accompanied by concise discussions and their respective equations.

\begin{table*}[]
\begin{center}
\caption{Loss Function and Equation}
\label{Loss Function Table}
\begin{tabular}{|l|l|l|}
\hline
\textbf{SI No} & \textbf{Loss Function Name}                                          & \textbf{Equation}                                                                                                                                             \\ \hline
1.             & KL Divergence                                                                      & $  \mathcal{L}_{KL} = \sum_{i=1}^{O} E[D_{KL}(Q(z_{ri}|o_i)||N(z_r))] $                                \\ \hline
2.             & Image Reconstruction Loss                                                                      &  $\mathcal{L}^{img}_1 = ||I - \hat{I}||_1$          \\ \hline
3.             & Object Latent Code Reconstruction Loss                                                            & $\mathcal{L}_1^{\text{latent}} = \sum_{i=1}^{Q} ||z_{s,i} - z'_{s,i}||_1$                               \\ \hline
4.             & Image Adversarial Loss                                                                      & $\mathcal{L}_{\text{GAN}}^{\text{img}} = \mathbb{E}_{x \sim P_{\text{real}}} \log D(x) + \mathbb{E}_{y \sim P_{\text{fake}}} \log(1 - D(y))$                                                                                                   \\ \hline
5.             & Object Adversarial Loss                                                                      & $ \mathcal{L}_{\text{GAN}}^{\text{obj}} = \mathbb{E}_{x \sim P_{\text{real}}} \log D(x) + \mathbb{E}_{y \sim P_{\text{fake}}} \log(1 - D(y))$ \\ \hline
6.             & Object Attribute Classification                                                                     & $ \mathcal{L}_{\text{GAN}}^{\text{obj}} = \mathbb{E}_{x \sim P_{\text{real}}} \log D(x) + \mathbb{E}_{y \sim P_{\text{fake}}} \log(1 - D(y))$                                                                                                                   \\ \hline
7.             & Auxiliar Attribute Classification
                         &  $\mathcal{L}_{\text{GAN}}^{\text{obj}} = \mathbb{E}_{x \sim P_{\text{real}}} \log D(x) + \mathbb{E}_{y \sim P_{\text{fake}}} \log(1 - D(y))$                                                         \\ \hline
8.             & Crop Matching Loss                                                                     & $\mathcal{L}_1^{\text{latent}} = \sum_{i=1}^{n} ||v_{i} - v'_{i}||_1$                   \\ \hline
9.             & Box Regression Loss                                                                  & $ \mathcal{L}_1^{\text{box}} = \sum_{i=1}^{n} ||b_{i} - \hat{b}_{i}||$                                                                                                                   \\ \hline
10.            & Appearance Loss                                                                   & $ \mathcal{L}_{\text{app}} = \mathbb{E}_{I \sim P_{\text{fake}}} [-D_{\text{app}}(A^{I}|I^{j})]$                                                                                                                                           \\ \hline
11.            & Adversarial Appearance Loss                                                                   & $ \mathcal{L}_{\text{app}}(D_{\text{app}}) = \mathbb{E}_{I^r \sim P_{\text{real}}} [\max(0, 1 - D_{\text{app}}(A^r|I^r))] + \mathbb{E}_{I^f \sim P_{\text{fake}}} [\max(0, 1 + D_{\text{app}}(A^f|I^f))]$                                                                                                                                           \\ \hline
12.            & Box Loss                                                                   & $ \mathcal{L}_{\text{box}} = \sum_i^n || b - b' ||$                                                                                                                                           \\ \hline
13.            & Cross Entropy Loss                                                                   & $\mathcal{L} = -\frac{1}{N}\left[\sum_{j=1}^{N}t_j\log(p_j) + (1 - t_j)\log(1 - p_j)\right]$                                                                                                                                           \\ \hline
14.            & L1 Pixel Loss                                                                   & $ \mathcal{L}_{\text{pixel}} = || I_G - I_{\text{final}} ||$                                                                                                                                           \\ \hline
15.            & Perceptual Similarity Loss                                                                   & $ \mathcal{L}_{\text{perceptual}} = \left\| P^{\phi} \left( I_k \right) - P^{\phi} \left( I_G \right) \right\|_2$                                                                                                                                           \\ \hline
16.            & RGB Loss                                                                   & $  \mathcal{L}_T = \|T(t) \cdot m(t)\|_2$                                                                                                                                           \\ \hline

\end{tabular}
\end{center}
\end{table*}

\subsection{KL Divergence Loss} 
KL (Kullback-Leibler) divergence is used to define the difference between two probability distributions, often between a predicted distribution and a target or true distribution. It has many applications in various fields, including image recognition, natural language processing, and speech recognition.
\begin{equation}
    \mathcal{L}_{KL} = \sum_{i=1}^{O} E[D_{KL}(Q(z_{ri}|o_i)||N(z_r))] 
   \label{KL3}
\end{equation}
Equation \ref{KL3} computes the KL-Divergence between the distribution \(Q(z_r|O)\) and the normal distribution \(N (z_r)\).\\
where,\\ 
\(o\) is the number of objects in the image/layout.

\subsection{Image Reconstruction Loss} 
Image Reconstruction Loss is used in various tasks such as image denoising, super-resolution, inpainting, and autoencoding. It discovers the dissimilarity between the original input image and its reconstructed version, which is generated by a neural network model.
\begin{equation}
    \mathcal{L}^{img}_1 = ||I - \hat{I}||_1
    \label{img}
\end{equation}
Equation \ref{img} penalizes the \(\mathcal{L}_1\) difference between ground-truth image \(I\) and reconstructed image \(\hat{I}\).
\subsection{Object Latent Code Reconstruction Loss} 
Object-Latent Code Reconstruction Loss is used in various tasks which are related to object manipulation, representation learning, or synthesis in computer vision and graphics.
\begin{equation}
    \mathcal{L}_1^{\text{latent}} = \sum_{i=1}^{Q} ||z_{s,i} - z'_{s,i}||_1
    \label{latent}
\end{equation}
Equation \ref{latent} penalizes the \(\mathcal{L}_1\) difference between the randomly sampled \(z_s \sim N(z_s)\) and the re-estimated \(z_{s}^'\) from the generated objects \(O^'\)
\subsection{Image Adversarial Loss}
The image adversarial loss, also known as the generator's adversarial loss, is where the generator gradually learns to produce images that closely resemble real images.
\begin{equation}
    \mathcal{L}_{\text{GAN}}^{\text{img}} = \mathbb{E}_{x \sim P_{\text{real}}} \log D(x) + \mathbb{E}_{y \sim P_{\text{fake}}} \log(1 - D(y))
    \label{gan}
\end{equation}
where in Equation \ref{gan},\\ \(x\) is the ground truth image \(I\), \(y\) is the reconstructed image
\(\hat{I}\) and sampled image \(I^'\).
\subsection{Object Adversarial Loss} 
Object Adversarial Loss mentioned in equation \ref{obj} is similar to image adversarial loss.
\begin{equation}
     \mathcal{L}_{\text{GAN}}^{\text{obj}} = \mathbb{E}_{x \sim P_{\text{real}}} \log D(x) + \mathbb{E}_{y \sim P_{\text{fake}}} \log(1 - D(y))
     \label{obj}
\end{equation}
where,\\
\(x\) is the objects
\(x\) cropped from the ground truth image \(I\),
\(y\) are \(\hat{O}\) and \(O^'\) cropped from the reconstructed image \(\hat{I}\) and sampled image \(I^'\).
\subsection{Auxiliar Attribute Classification Loss} 
For better generalization, Auxiliar attribute classification loss is used.\\ Equation \ref{fake} shows the equation for Auxiliar Attribute Classification Loss.
\begin{equation}
     \mathcal{L}_{\text{GAN}}^{\text{obj}} = \mathbb{E}_{x \sim P_{\text{real}}} \log D(x) + \mathbb{E}_{y \sim P_{\text{fake}}} \log(1 - D(y))
     \label{fake}
\end{equation}
\(\mathcal{L}_{AC}^{\text{obj}}\) \text{ from } \(D_{\text{obj}}\) encourages the generated objects \(\hat{O_i}\) and \(O'_i\)  to be recognizable as their corresponding categories.

\text{The final loss function \ref{final} from Equation \ref{KL3},\ref{img},\ref{latent},\ref{gan},\ref{obj},\ref{fake} is:}
\begin{equation}
    \mathcal{L} = \lambda_1 \mathcal{L}_{KL} + \lambda_2 \mathcal{L}_{1}^{img} + \lambda_3 \mathcal{L}_{1}^{latent} + \lambda_4 \mathcal{L}_{adv}^{img} + \lambda_5 \mathcal{L}_{adv}^{obj} + \lambda_6 \mathcal{L}_{AC}
    \label{final}
\end{equation}

\subsection{Crop Matching Loss} 
For the purpose of respecting the object crops’ appearance, 
\begin{equation}
    \mathcal{L}_1^{\text{latent}} = \sum_{i=1}^{n} ||v_{i} - v'_{i}||_1
    \label{CML}
\end{equation}
Equation \ref{CML} penalizes the \(\mathcal{L}_{1}^{\text{latent}}\) difference between the object crop feature map and the feature map of the object re-extracted from the generated images.

\subsection{Object Perceptual Loss}
Image perceptual loss \(\mathcal{L}_P^{\text{img}}\) penalizes the \(\mathcal{L}_{1}\) difference in the global feature space between the ground-truth image \(I\) and the reconstructed image \(\hat{I}\), while object perceptual loss \(\mathcal{L}_P^{\text{obj}}\) penalizes that between the original crop and the object crop re-extracted from \(\hat{I}\).

\subsection{Box Regression Loss} 
Box Regression Loss is used in Object detection tasks such as YOLO, SSD, etc.
\begin{equation}
    \mathcal{L}_1^{\text{box}} = \sum_{i=1}^{n} ||b_{i} - \hat{b}_{i}||
    \label{L_1}
\end{equation}
Equation \ref{L_1} penalizes the \(\mathcal{L}_{1}^{\text{box}}\) difference between ground truth and predicted boxes.\\ Therefore, the final loss function of the model is defined in equation \ref{ILO}:
\begin{dmath}
  \mathcal{L} = \lambda_1 \mathcal{L}_1^{\text{img}} + \lambda_2 \mathcal{L}_1^{\text{latent}} + \lambda_3 \mathcal{L}_{\text{GAN}}^{\text{img}} + \lambda_4 \mathcal{L}_{\text{GAN}}^{\text{obj}} + \lambda_5 \mathcal{L}_{\text{AC}}^{\text{obj}} + \lambda_6 \mathcal{L}_{\text{P}}^{\text{img}} 
  + \lambda_7 \mathcal{L}_{\text{p}}^{\text{obj}} + \lambda_8 \mathcal{L}_{\text{box}}
\label{ILO}  
\end{dmath}

Where,\\
           \(\lambda_i\) are the parameters balancing losses.
\subsection{Appearance Loss} 
Appearance Loss is used in various tasks such as image generation, image-to-image translation, etc.
\begin{equation}
    \mathcal{L}_{\text{app}} = \mathbb{E}_{I \sim P_{\text{fake}}} [-D_{\text{app}}(A^{I}|I^{j})]
    \label{AL}
\end{equation}
This Equation \ref{AL} represents the appearance loss ($L_{\text{app}}$), which measures the discrepancy between the appearance of fake/generated images ($I$) and their corresponding attributes ($A^{I}$) compared to the ground truth images ($I^{j}$).

\textbf{Adversarial Appearance Loss Function}
\begin{dmath}
    \mathcal{L}_{\text{app}}(D_{\text{app}}) = \mathbb{E}_{I^r \sim P_{\text{real}}} [\max(0, 1 - D_{\text{app}}(A^r|I^r))] + \mathbb{E}_{I^f \sim P_{\text{fake}}} [\max(0, 1 + D_{\text{app}}(A^f|I^f))]
    \label{DL}
\end{dmath}
This Equation \ref{DL} represents the adversarial appearance loss, which is formulated using adversarial training with a discriminator network $D_{\text{app}}$. It consists of two terms: one for real images ($I^r$) and one for fake/generated images ($I^f$).
\subsection{Box loss} Box loss is a loss function used in object detection to measure how well a model predicts the size and location of bounding boxes around objects.
\begin{equation}
    \mathcal{L}_{\text{box}} = \sum_i^n || b - b' ||
    \label{Boxx}
\end{equation}
The Equation \ref{Boxx} penalizes the \(\mathcal{L}_{box}\) distance between ground truth boxes from MS COCO vs predicted labels.From this method, it's easy to determine how well a model is.

\subsection{Cross Entropy Loss} 
Cross-entropy, also known as binary cross-entropy mentioned in equation \ref{cross}, helps models determine how wrong they are and improve themselves based on the wrong.Here, cross-entropy loss function measures the error between two probability distribution.
 \begin{equation}
     \mathcal{L} = -\frac{1}{N}\left[\sum_{j=1}^{N}t_j\log(p_j) + (1 - t_j)\log(1 - p_j)\right]
     \label{cross}
 \end{equation}

For $N$ data points, where $t_i$ is the truth value taking a value 0 or 1 and $p_i$ is the Softmax probability for the $i^{th}$ data point.

\subsection{Mask loss}
Penalizes the difference between the masks predicted vs. the ground truth masks, using cross-entropy loss in Equation \ref{cross}.

\subsection{L1 pixel loss} 
Penalizes the difference between the ground truth image from MS COCO \cite{caesar2018coco} and the final generated image at the end of the incremental generation. L1 pixel losses are also used to penalize the difference between the previous and current generated images.
\begin{equation}
    \mathcal{L}_{\text{pixel}} = || I_G - I_{\text{final}} ||
    \label{L1}
\end{equation}
From Equation \ref{L1}, it shows how to calculate between the previous and current generated images.\\
where, \\
The generated image $I_G$
and the ground truth image $I_{\text{final}}$.

\subsection{Perceptual Similarity Loss} 
Perceptual similarity serves as a regularization to ensure that the images generated at different steps are contextually similar to each other. Since we do not have ground truth for intermediate steps, we introduce an additional perceptual similarity loss between the final ground truth image and the different images generated in the intermediate steps.
\begin{equation}
    \mathcal{L}_{\text{perceptual}} = \left\| P^{\phi} \left( I_k \right) - P^{\phi} \left( I_G \right) \right\|_2
    \label{perceptual}
\end{equation}

The Equation \ref{perceptual} shows the model to ‘hallucinate’ for the intermediate steps in such a way that the contents of the image are similar to the ground truth\cite{mittal2019interactive}.\\
where,\\
\(p^{\phi}\) is the function computing a latent embedding that captures the perceptual characteristics of an image.

\subsection{RGB Loss, A Depth Loss, and A Transmittance Loss} 
A RGB loss, a depth loss, and a transmittance loss are used \cite{zhang2024text2nerf} to optimize the radiance field of the 3D scene. Here some authors also design a depth-aware transmittance loss \(\mathcal{L}_T\) to encourage the NeRF network.
\begin{equation}
    \mathcal{L}_T = \|T(t) \cdot m(t)\|_2
    \label{RGB}
\end{equation}
The RGB loss, in Equation \ref{RGB}\\
\( \mathcal{L}_{RGB}\)  is defined as an \(\mathcal{L}_T\) loss between the rendered pixel color \(C^R\) and the color \(C\) generated by the diffusion model \(f_d\).\\
Where,\\
    \(m(t)\) is a mask indicator and
    \(T(t)\) is the accumulated Transmittance\\
The total objective is then defined in Equation \ref{RDT}.
\begin{equation}
   \mathcal{L}_{\text{total}} = \mathcal{L}_{\text{RGB}} + \lambda_d \mathcal{L}_{\text{Depth}} + \lambda_t \mathcal{L}_T
   \label{RDT}
\end{equation}
Where,\\ 
      \(\lambda_d\) and \(\lambda_t\) are constant hyperparameters balancing between terms.

\section{Evaluation Techniques}
\label{sec:evaluation techniques}
Every system must be evaluated based on a metric or attribute value to measure the quality of the system's performance. Every scene generation system also follows different evaluation metrics for comparison with other similar existing systems. Fréchet inception distance (FID), Kullback-Leibler (KL) Divergence, Inception score (IS), Intersection over Union (IoU), and Mean Average Precision (mAP) are the five most used evaluation metrics for automatic scene or image generation systems using different GAN based models and other deep learning models. Kernel Inception Distance (KID) metric is also used in different systems \cite{farshad2023scenegenie}\cite{huang2018holistic}. However, \cite{dahnert2021panoptic} proposed a new metric called Panoptic Reconstruction Quality (PRQ) for 3D scene generation system evaluation for their proposed system. \cite{po2023compositional} used CLIP R-Precision and attribute value to compare the proposed system performance with other single text prompt SJC models. \cite{abdul2018shrec} used both the Precision-Recall (PR) diagram and the Discounted Cumulated Gain (DCG) metric for evaluating the implemented system. \cite{cheng2019interactive} used CIDEr, ROUGH-L, and METEOR as the evaluation metric of their proposed system. Table \ref{evaluation metric table} shows the quantitative evaluation metrics used in different papers and their related equations. Almost all the scene generation systems evaluated their system qualitatively by displaying the images or scenes generated by their system and others. Some systems are also evaluated by the users inspection.
\begin{table*}[]
\begin{center}
\caption{Evaluation Metrics and Equations}
\label{evaluation metric table}
\begin{tabular}{|c|c|c|}
\hline
\textbf{SI No} & \textbf{Evaluation Metric Name}                                          & \textbf{Equation}                                                                                                                                             \\ \hline
1.             & FID                                                                      & $d^{2} = \left| \varphi_{x} - \varphi_{y} \right|^{2} + tr (\sum_{}^{}x + \sum_{}^{} y - 2(\sum_{}^{}x * \sum_{}^{} y)^{1/2})$                                \\ \hline
2.             & KID                                                                      & \begin{tabular}[c]{@{}c@{}}$KID = MMD (f_{generated},f_{input})^{2}$\\ $MMD^{2} (x,y) = \left\| \varphi_{x} - \varphi_{y}\right\|^{2}$\end{tabular}           \\ \hline
3.             & KL Divergence                                                            & $D_{KL}\left( P||Q \right) = \sum_{x\in X}^{} P\left( x \right) \log\left( \frac{P\left( x \right)}{Q\left( x \right)} \right)$                               \\ \hline
4.             & IoU                                                                      & $IoU = \frac{\text{Area of Overlap}}{\text{Area of Union}}$                                                                                                   \\ \hline
5.             & PRQ                                                                      & $PQR^{c} = \frac{\sum_{\left( i,j \right)\in TP^{^{C}}}^{}IoU\left( i,j \right)}{\left| TP^{c} \right|+ 0.5 \left| FP^{c} \right|+ 0.5\left| FN^{c} \right|}$ \\ \hline
6.             & mAP                                                                      & $mAP = \frac{1}{n} \sum_{k=1}^{k=n} AP_{k}$                                                                                                                   \\ \hline
7.             & \begin{tabular}[c]{@{}c@{}}Precision-Recall (PR) \\ Diagram\end{tabular} & \begin{tabular}[c]{@{}c@{}}$Precision = \frac{TP}{TP+FP}$\\ $Recall = \frac{TP}{TP+FN}$\end{tabular}                                                          \\ \hline
8.             & DCG                                                                      & $DCG_{P} = \sum_{i=1}^{p} \frac{rel_{i}}{\log_{2}\left( i+1 \right)} = rel_{1} + \sum_{i=2}^{p} \frac{rel_{i}}{\log_{2}\left( i+1 \right)}$                   \\ \hline
9.             & ROUGE-L                                                                  & $R_{lcs} = \frac{LCS\left( X,Y \right)}{m}$                                                                                                                   \\ \hline
10.            & METEOR                                                                   & $M = F_{mean}(1-p)$                                                                                                                                           \\ \hline

\end{tabular}
\end{center}
\end{table*}

\subsection{Fréchet inception distance (FID)}
Fréchet inception distance (FID) is a metric to assess the image generated by different versions of the GAN model \cite{yu2021frechet}. FID  is measured by computing the proximity or similarity of the features of the generated image and input image of the GAN model. FID is computed in four stages. Firstly, the generated image and input image are preprocessed to make them in the same size and normalize the pixel value. Secondly, the images are passed to the Inception-v3 model to transform the pixel values into numerical vectors that can extract features like lines, borders, and other shapes. Thirdly, the mean and covariance matrix of the extracted features of each image is calculated by statistical analysis and the difference of mean and covariance matrix of each image is calculated. Finally, the Fréchet distance is measured between the generated and input image using the equation \ref{FID}. The lower value in Fréchet distance means the generated image is more similar to the input image.
\begin{equation}
   d^{2} = \left| \varphi_{x} - \varphi_{y} \right|^{2} + tr (\sum_{}^{}x + \sum_{}^{} y - 2(\sum_{}^{}x \sum_{}^{} y)^{1/2}) 
   \label{FID}
\end{equation}
Here, 
d is the Fréchet distance between two multivariate distributions of two images. 

$\varphi_{x}$  and $\varphi_{y}$ are the mean of the two images, and \par

$\sum_{}^{}x$ and $\sum_{}^{} y$ are the covariance matrices of the distributions.


\subsection{Kernel Inception Distance (KID)}
Kernel Inception Distance (KID) is used to assess the quality of the images generated using different deep learning \cite{binkowski2018demystifying}. KID compares the Maximum Mean Discrepancy (MMD) using the equation \ref{KID} of two datasets that contain the features of the two images. The features of the images are extracted in the same way as FID.  
\begin{equation}
   KID = MMD (f_{generated},f_{input})^{2}
   \label{KID}
\end{equation}
Here, MMD is the maximum mean discrepancy.
$f_{generated}$ and $f_{input}$ are the feature set of the two images.
Maximum Mean Discrepancy (MMD) is a kernel-based statistical test \cite{gretton2012kernel}. MMD mainly calculates the distance of the features of the two images x and y. Equation \ref{MMD} shows how MMD is computed.
\begin{equation}
   MMD^{2} (x,y) = \left\| \varphi_{x} - \varphi_{y}\right\|^{2}
   \label{MMD}
\end{equation}
where $\varphi_{x}$  and $\varphi_{y}$ are the mean of the features of the two images.

\subsection{Kullback-Leibler (KL) Divergence}
Kullback-Leibler (KL) Divergence which is also called relative entropy, is the measurement of the differences of two probability distributions in the sample space \cite{shlens2014notes}. KL divergence is used for identifying how different the generated image or scene using the GAN models is from the actual input image or scene. KL divergence can also be used as a loss function for different deep learning networks \cite{zhao2020layout2image}.

KL divergence is denoted by $D_{KL}\left( P||Q \right)$ that indicates the measurement of how a probability distribution P is different from another probability distribution Q.  $D_{KL}\left( P||Q \right)$ is calculated by the equation \ref{KL}.
\begin{equation}
   D_{KL}\left( P||Q \right) = \sum_{x\in X}^{} P\left( x \right) \log\left( \frac{P\left( x \right)}{Q\left( x \right)} \right)
   \label{KL}
\end{equation}
which is equivalent to equation \ref{KL2}.
\begin{equation}
   D_{KL}\left( P||Q \right) = - \sum_{x\in X}^{} P\left( x \right) \log\left( \frac{Q\left( x \right)}{P\left( x \right)} \right)
   \label{KL2}
\end{equation}

\subsection{Inception score (IS)}
Inception score (IS) is a commonly used evaluation algorithm to determine the quality as well as the diversity of the generated image from GAN models. Inception score is introduced based on Google's "Inception" image classification network \cite{salimans2016improved}. Firstly, the image classification network gives the probability distribution of the generated image. Secondly, the inception score process compares the probability distribution of all generated images and compute marginal distribution. Marginal distribution mainly computes the amount of variety is present in all the generated images. Finally, the probability distribution and marginal distribution are combined into a final single score which is mainly the Inception score. The final score is calculated using the Kullback-Leibler (KL) Divergence.  

\subsection{Intersection over Union (IoU)}
Intersection over Union (IoU) is a metric for evaluating the performance of an object detector model or image segmentation model \cite{rahman2016optimizing}. For calculating the IoU metric value, both the ground-truth bounding box and the predicted bounding box are needed. IoU is calculated using the equation \ref{IoU} by overlapping the two bounding boxes. If the IoU value for an object detection or segmentation model is higher than 0.5 for the testing dataset, the model is considered to be performing good.
\begin{equation}
   IoU = \frac{\text{Area of Overlap}}{\text{Area of Union}}
   \label{IoU}
\end{equation}

\subsection{Panoptic Reconstruction quality (PRQ)}
Panoptic Reconstruction quality (PRQ) is a metric for 3D scene generation which is similar to 2D panoptic segmentation quality metric \cite{kirillov2019panoptic}. PQR for a class category c is calculated using the equation \ref{PQR}.
\begin{equation}
   PQR^{c} = \frac{\sum_{\left( i,j \right)\in TP^{^{C}}}^{}IoU\left( i,j \right)}{\left| TP^{c} \right|+ 0.5 \left| FP^{c} \right|+ 0.5\left| FN^{c} \right|}
   \label{PQR}
\end{equation}
where P, FP, and FN denote true positives, false positives, and false negatives for class c, respectively and and predicted segments are matched with ground truth via a voxelized IoU overlap of greater than 25\%.

\subsection{Mean Average Precision (mAP)}
Mean Average Precision (mAP) is a metric for measuring the performance of a model that detects objects and retrieves information from an image \cite{henderson2017end}. The process first labels the predicted object or score in classes and generates the confusion matrix for each class. The average precision for each class is calculated. Finally, the mean value for the average precision of all classes is calculated. Mean Average Precision (mAP) is calculated using the equation \ref{map}. The higher value of mAP indicates the better performance of the model to predict objects on an image. 
\begin{equation}
   mAP = \frac{1}{n} \sum_{k=1}^{k=n} AP_{k}
   \label{map}
\end{equation}
where n = the number of classes and $AP_{k}$ is the average precision of the class k.

\subsection{Precision-Recall (PR) diagram}
Precision-Recall (PR) diagram is used when the scene generation needs to predict the class of the object from the input. It is needed for supervised classification problems. Precision-Recall (PR) curve or diagram helps to identify how well the model predicts the positive class or accurate class of the object. The performance of different models can be compared by calculating the area under the PR curve value.

Precision-Recall curve is generated by plotting recall in the x-axis and precision in the y-axis \cite{beger2016precision}. The precision and recall are computed using equation \ref{precision} and equation \ref{recall} respectively from the confusion matrix.
\begin{equation}
   Precision = \frac{TP}{TP+FP}
   \label{precision}
\end{equation}

\begin{equation}
   Recall = \frac{TP}{TP+FN}
   \label{recall}
\end{equation}
where True Positive (TP) = Number of correctly identified positive class instances,

False Positive (FP) = Number of negative class instances wrongly identified as positive class instances, and

False Negative (FN)= Number of positive class instances wrongly identified as negative class instances.

\subsection{Discounted Cumulated Gain (DCG)}
Discounted Cumulated Gain (DCG) \cite{abdul2018shrec} is an evaluation metric to measure the quality of a recommendation and information retrieval system. It is mainly used for accessing the quality of different ranking systems where the most relevant items or objects are considered in the top positions in the ranking list. Discounted Cumulated Gain (DCG) at a particular rank position P is calculated using equation \ref{DCG}.
\begin{equation}
   DCG_{P} = \sum_{i=1}^{p} \frac{rel_{i}}{\log_{2}\left( i+1 \right)} = rel_{1} + \sum_{i=2}^{p} \frac{rel_{i}}{\log_{2}\left( i+1 \right)}
   \label{DCG}
\end{equation}
where $rel_{i}$ = the graded relevance of the result at position \textit{i}.

\subsection{CIDEr metric}
Consensus-based Image Description Evaluation (CIDEr) metric evaluates image descriptors based on human consensus \cite{vedantam2015cider}. The metric measures the differences between the generated caption and the given reference caption of the image. Firstly, the reference input captions are considered as ground truth. Secondly, the generated caption is compared with the ground truth using the BLEU (Bilingual Evaluation Understudy) score \cite{seljan2012bleu}. BLEU score mainly computes the n-gram overlap between the captions. Thirdly, IDF (Inverse Document Frequency) \cite{robertson2004understanding} is implemented to manipulate the BLEU score so that more weight is given to the words that were rare in the ground truth but present in the generated caption. Finally, all weighted BLEU scores of all generated captions are averaged to calculate the final CIDEr score. 

\subsection{ROUGE-L metric}
Recall-Oriented Understudy for Gisting Evaluation (ROUGE) is a set of metrics as well as a software package \cite{chin2004rouge}. It measures the quality of automatically generated text or summary in natural language processing. It compares the generated text with the reference text using different methods. ROUGE-L is an extended version where the L stands for Longest Common Subsequence (LCS) \cite{lin2004automatic}. ROUGE-L finds the longest common co-occurring in sequence n-grams using sentence-level structure similarity. The similarity between two texts or summaries, X of m length and Y of n length can be calculated using ROUGE-L and is measured using equation \ref{ROUGE-L}.
\begin{equation}
   R_{lcs} = \frac{LCS\left( X,Y \right)}{m}
   \label{ROUGE-L}
\end{equation}
where e LCS(X, Y) is the length of the longest common subsequence of X and Y. 

\subsection{METEOR metric}
Metric for Evaluation of Translation with Explicit ORdering (METEOR) metric is an evaluation metric for machine translation output \cite{banerjee2005meteor}. It is calculated by computing the harmonic mean of unigram precision and recall. The recall has a higher weight than precision in this metric. The harmonic mean and final score (M) of METEOR metric are calculated using equation \ref{mean} and equation \ref{meteor} respectively.
\begin{equation}
   F_{mean} = \frac{10PR}{R+9P}
   \label{mean}
\end{equation}
\begin{equation}
   M = F_{mean}(1-p)
   \label{meteor}
\end{equation}
where P = Unigram Precision, R = Unigram Recall and p = penalty.
Penalty (p) is calculated using equation \ref{penalty}.
\begin{equation}
   p = 0.5\left( \frac{c}{u_{m}} \right)
   \label{penalty}
\end{equation}
where c = The number of chunks that are adjacent in the hypothesis and in the reference, and  $u_{m}$ = The number of mapped unigrams. 
\subsection{CLIP R-Precision}
CLIP R-Precision \cite{park2021benchmark} is a benchmark for evaluating the performance of text-to-image synthesis systems. The metric uses the vision-and-language CLIP model to compute the R-Precision value. R-precision computes the top-R retrieval accuracy to retrieve the matching text. CLIP R-Precision can evaluate a system having both visual and textual inputs.

\subsection{Qualitative evaluation}
Qualitative evaluation in scene generation is a method that generates scenes or images instead of numerical data to compare with existing systems. The scenes or images generated by two or more systems are displayed and users try to find differences in images visually.

\subsection{User Study}
Sometimes, the users are asked to evaluate the automatically generated scene by the system. The main reason is to check if the system generated scene can fulfill the users' satisfaction or not. It is also called human evaluation.

\section{Use Cases}
\label{sec:use cases}
With the improvement of the advanced scene generation method, people are exploring different use cases of it. One side it is educating and improving skills for employees, on the other side it is keeping people entertained and engaged. Some use cases for scene generation are given below:

\subsection{Entertainment and Gaming}
Scene generation can be a great prominent tool for gaming and entertainment. If any gamer plays the same scenario again and again they might get bored. Interactive scene generation like \cite{kumaran2023scenecraft} or \cite{cheng2019interactive} can help a lot to generate new scenes every time where no two scenes are alike. This will improve dynamically in the game by changing the graphics of that game as well as the strategy and gameplay requirements. I will also help graphics designers and game designers to think of more complicated scenarios without compromising the performances. It also helps filmmakers add breathtaking visual effects and CGI-heavy sequences in their movies or series. So, it makes easier for those filmmakers to focus on narrating their story creatively to the viewers without additional cost or labor. Overall, scene generation helps entertainment and gaming because it pushes the creative envelope and makes it possible to create immersive experiences that draw viewers and players into vibrant, dynamic worlds with amazing experiences for the viewer.

\subsection{Architecture and Urban Planning}
Generated scenes can visualize the plan and design of architects and urban planners in a efficient way so they can decide what needs to be improved. It also makes it easier for them to convey their idea to their client, stakeholders, or other public. This helps the architects and planners to try out different ideas and options and see their impact on surrounding environments. Additionally, scene generation facilitates community engagement by providing immersive experiences that allow residents to experience and provide feedback on proposed projects. They can employ factors like traffic flow, historical preservation, pedestrian movement, or environmental impact in their assessment like a real environment. As a result, architects and urban planners can make informed decisions for creative unknown situations.

\subsection{Training and Simulation}
Scene generation can produce realistic scenarios for different fields like military exercise emergency response, and industry safety. It can also generate an interactive scene, through which students can interact with different objects, navigate through scenes, and practice in a controlled environment. It also offers the option to customize the scenario to match specific learning objectives, skill levels, and training requirements. As a result, the trainee can practice and improve their skill with real risk where they feel less confident. Finally, performance can be monitored, analyzed, and evaluated based on customized and updated scenes based on skillset.

\subsection{Education and Research}
Scene generation from text has a huge impact on education and research. Visual representation with text makes complex topics easier for learners. It also helps the second language learners. People with a learning disability or damaged brain might get huge help from the generated scene, when all the textual menu, operation, signs, and instructions are converted into a graphic representation. It also helps researchers by providing interactive environments for experimentation and exploration. It provides tools to the researcher real environment where they can test their hypothesis, simulate, execute the experiment, and analyze it. Additionally, researchers can understand human behavior and reactions in different circumstances.

\subsection{Safety and Emergency Preparedness}
To maintain safety on roads autonomous vehicle requires visuals from different sides. Research like \cite{devries2021unconstrained} or \cite{novotny2019perspectivenet} can ensure the consistency of those vehicles by providing lots of data from different directions. It is also able to generate a real environment for areas that helps rescuers make decisions based on the observation. Real-time data incorporated with generated scenes can help further management to make informed decisions based on the current situation. Additionally, it helps to analyze and identify potential hazards and risks based on critical infrastructure, high-risk areas, and vulnerabilities for proactive risk management.

\subsection{Communication and Advertisement}
Visualization is a strong tool for communicating with your users and stockholders. Scene generation can generate various characters, charts, layouts, and scenes that improve the quality of communication and also grab attention. Also helps users and stockholders to understand concepts better and provide their feedback based on understanding. For big industries, catchy scenes can get attention easily and create a long-lasting impression on the potential investors, and clients. Visual graphics are proven effective ways to raise awareness and gather community input from a large population. All this situation-generated scene makes communication easier and faster for everyone.

\section{Challenges of scene generation}
\label{sec:Challenges of scene generation}
Automatic creation of scenes is very popular nowadays. Non-domain experts can develop different complex scenes using the systems developed by different researchers. However different challenges occur during the development of scene generation scenes. Some challenges faced by different researchers are given below:

\textbf{Availability of Dataset:} Finding the appropriate dataset for training and evaluating the system is a big challenge. It is impossible to have a dataset that contains all types of objects or 3D models existing in our world. So, every system is created using a dataset having a fixed number of objects or a fixed category object. Also, there are many well-known datasets for different indoor scenes, but not for different outdoor scenes.

\textbf{Diversity:} A scene can be designed by using different types of objects. A scene has different types of attributes like color, number of objects, size of different objects, etc. A simple bedroom can have different colors, different objects, and different placements of the objects. The researcher of the scene generation system must consider these factors while developing the system.

\textbf{No Mention of Implicit Relations in the Input Text:} In the text-to-scene generation system, researchers face difficulty with implicit relations between objects that are not mentioned in the input. Sometimes, more objects are also needed in the scene that are not added to the input text. "A cake on the table" is an example where there is no mention that a cake must be placed on a plate and the plate is placed on the table. The input text also sometimes does not mention the proper location of objects ("A table in the room": There is no location of the table mentioned here like the middle, left, right, etc).

\textbf{Complexity and User Satisfaction:} Systems should be built to fulfill the user's satisfaction. Some people prefer to have simple scenes whereas others prefer having complex scenes with more objects. It is not always possible to satisfy a user with a system that generates scenes automatically. Users should be able to modify their scenes based on their requirements. So, an interactive scene generation system is more acceptable.

\textbf{Generalized System:} Different researchers used different types of input for creating the system like text, image, bounding box, layouts, graphs of objects, etc. There is no generalized system that can take any type of input chosen by the user. More hybrid systems combining all types of input are needed. However, the hybrid system can be slower as well as costly. So, every researcher must work on developing a system that is faster as well as cost-effective.

\textbf{Domain Knowledge:}
For developing a scene generation system, technical knowledge of graphics and other related fields is a must. Non-domain researchers of the scene generation system have to study a lot about different technical things related to computer graphics and other fields before developing a system.

\textbf{Selection of Model:} It is always a challenge to find out which model will work best for a system. There are different types of models proposed at different times. Some models work best for one type of dataset, whereas the same model can give very poor performance for other types of dataset. So selecting the model for a system is dependent on different factors and every researcher takes a much longer time to solve this challenge than other challenges. 

\textbf{Missing Class/Label of Objects of the Scene:} Most scene generation systems consider the class or category of the objects for developing the scene. This is called supervised learning. However, missing enough properly labeled data is one of the key challenges for any kind of supervised machine learning. Also creating a class/label for every object is not possible. So, a semi-supervised learning method is an effective solution for resolving this challenge. In semi-supervised learning, some objects have class labels that can guide the machine learning models to predict the class of other unknown objects to generate a scene.

\section{Discussion}
\label{sec:discussion}
Based on the comprehensive review and analysis conducted in this paper, several key insights and areas for future research have emerged in the field of automatic scene generation. The diversity of models, including VAEs, GANs, Transformers, and Diffusion Models, highlights the multifaceted approaches to scene generation. Each model type brings unique strengths and weaknesses, suggesting that hybrid approaches may offer significant advantages. Specialization within model categories, such as the use of Conditional VAEs for guided image generation and LayoutGANs for structured scene layouts, underscores the importance of context-specific model selection.

The extensive use of datasets like COCO-Stuff, Visual Genome, and MS-COCO demonstrates their critical role in advancing scene generation techniques. However, limitations such as noisy annotations and domain-specific biases in these datasets indicate the need for more refined and diverse datasets. Synthetic datasets, like SUNCG and RPLAN, offer controlled environments for training but may lack the complexity of real-world scenes, pointing to the need for better synthetic-real hybrid datasets.

Common evaluation metrics, including FID, KL Divergence, and IS, are essential for benchmarking models. However, these metrics often fall short in capturing the nuanced realism and contextual appropriateness of generated scenes. Developing new evaluation metrics that consider human perceptual qualities and context-specific realism is crucial for advancing the field.

Handling complex scenes with multiple objects and maintaining realistic spatial relationships remains a significant challenge. Techniques like object-aware cross-attention and scene graphs are promising but require further refinement. The integration of more sophisticated spatial reasoning and relational models can improve the handling of complex scenes.

Future research should explore hybrid models that combine the strengths of VAEs, GANs, Transformers, and Diffusion Models. For example, integrating the generative capabilities of GANs with the contextual understanding of Transformers could yield more realistic and coherent scene generation. Investigating how these hybrid models can be efficiently trained and deployed in real-time applications will be critical.

Developing new datasets that provide richer annotations and cover a broader range of scenes and contexts is essential. These datasets should include diverse environmental settings, object types, and interactions to better train and evaluate models. Advanced data augmentation techniques, such as domain adaptation and transfer learning, can help bridge the gap between synthetic and real-world data.

Designing new evaluation metrics that incorporate human perceptual feedback and contextual accuracy will provide a more comprehensive assessment of scene generation models. Crowdsourcing evaluations and leveraging psychophysical studies can contribute to this goal. Metrics that assess the consistency of object relationships and the plausibility of interactions within scenes are particularly needed.

Research should focus on enhancing models' abilities to understand and generate complex scenes with multiple interacting objects. This involves improving spatial reasoning and relational understanding within models. Techniques such as graph neural networks (GNNs) and relational reasoning modules can be integrated into existing architectures to better handle complex scene layouts and object interactions.

Achieving real-time scene generation is crucial for applications in robotics, gaming, and simulation. Research should prioritize optimizing model architectures and training processes to meet real-time performance requirements without compromising on quality. Exploring the use of hardware accelerators and efficient inference techniques can contribute to this goal.

Looking ahead, the field of automatic scene generation has the potential to achieve significant breakthroughs by addressing these key areas, ultimately leading to more intelligent and interactive systems.

\section{Conclusion}
\label{sec:conclusions}
Our paper has provided a comprehensive review of recent advancements in scene generation systems, encompassing various architectures, loss functions, optimizers, and datasets. These systems utilize diverse input types, such as text, images, graphs, bounding boxes, layouts, and 3D models, and are primarily developed using different versions of GANs, VAEs, Transformers, and Diffusion models. We have detailed the fundamental principles of these models and summarized the datasets and loss functions employed in scene generation. Additionally, we have outlined the evaluation metrics used to assess the performance of these systems.

Despite the significant progress made in recent years, scene generation systems still face several limitations and challenges, including maintaining realism, handling complex scenes, and ensuring consistency in object relationships and spatial arrangements. However, the advancements achieved so far indicate a promising future for this field.

By continuing to focus on hybrid models, enhanced datasets, improved evaluation metrics, better handling of complex scenes, and real-time capabilities, the field can push the boundaries of what is possible in scene generation. These improvements will have far-reaching implications across various applications, such as robotics, virtual reality, and visual content creation, ultimately leading to more intelligent and interactive systems.


\section*{Conflicts of interest}
The authors have no conflicts of interest regarding this research.






\bibliographystyle{IEEEtran}
\bibliography{ijcai19.bib}

\end{document}